\documentclass[lettersize,journal]{IEEEtran}

\usepackage{amsmath,amsfonts}
\usepackage{algorithmic}
\usepackage{algorithm}
\usepackage{array}
\usepackage[caption=false,font=normalsize,labelfont=sf,textfont=sf]{subfig}
\usepackage{textcomp}
\usepackage{stfloats}
\usepackage{url}
\usepackage{verbatim}
\usepackage{graphicx}
\usepackage{cite}
\usepackage{booktabs}
\usepackage{threeparttable}
\usepackage{makecell}

\usepackage{listings}
\usepackage{xcolor}

\usepackage{ragged2e} 
\usepackage{booktabs,makecell, multirow, tabularx}

\usepackage{subfig}
\usepackage{hyperref}
\usepackage{diagbox}

\lstdefinestyle{yaml}{
	language=YAML, % 设置语言为YAML
	basicstyle=\ttfamily\small, % 设置基本样式
	backgroundcolor=\color{gray!10}, % 设置背景色
	breaklines=true, % 自动换行
	captionpos=b, % 标题位置
	commentstyle=\color{gray}, % 注释样式
	keywordstyle=\color{blue}, % 关键词样式
	stringstyle=\color{orange}, % 字符串样式
	showstringspaces=false, % 不显示空格
	tabsize=2, % Tab大小
	frame=single, % 添加边框
	frameround=tttt, % 边框角落样式
	framexleftmargin=-5pt, % 左边框距离
	keepspaces=true % 保留空格
}

\lstdefinelanguage{yaml}{
	sensitive=true,
	morecomment=[l]{\#},
	morecomment=[s]{/*}{*/},
	morestring=[b]',
	morestring=[b]",
	stringstyle=\color{orange},
	keywordstyle=\color{blue},
	keywords={true,false},
	identifierstyle=\color{red},
	basicstyle=\small\ttfamily,
	commentstyle=\color{gray}\ttfamily,
	tabsize=2,
	showstringspaces=false
}

\usepackage{orcidlink}
\usepackage{etoolbox}

\begin{document}
	
	\title{Benchmarking SLAM Algorithms in the Cloud: 
		
		The SLAM Hive Benchmarking Suite}
	
	\author{Xinzhe Liu~\orcidlink{0009-0000-1489-3493}
		, Yuanyuan Yang~\orcidlink{0000-0001-9624-8585}
		, Bowen Xu~\orcidlink{0009-0004-4325-4876}
		, Delin Feng~\orcidlink{0009-0005-2092-0159}
		, S\"oren Schwertfeger
		~\orcidlink{0000-0003-2879-1636}
		~\IEEEmembership{Senior Member,~IEEE}
		\thanks{The authors are with ShanghaiTech University, Key Laboratory of Intelligent Perception and Human-Machine Collaboration – ShanghaiTech University, Ministry of Education, China
			{\tt\small \{liuxzh12023, yangyy2, xubw, fengdl, soerensch\}@shanghaitech.edu.cn}}
	}
	%\author{IEEE Publication Technology,~\IEEEmembership{Staff,~IEEE,}
		%
		%        % <-this % stops a space
		%\thanks{This paper was produced by the IEEE Publication Technology Group. They are in Piscataway, NJ.}% <-this % stops a space
		%\thanks{Manuscript received April 19, 2021; revised August 16, 2021.}}
	
	% TODO The paper headers
	%%%%% %%%%% 要不要？ \markboth{Journal of \LaTeX\ Class Files,~Vol.~14, No.~8, August~2021}%
	%%%%%{Shell \MakeLowercase{\textit{et al.}}: A Sample Article Using IEEEtran.cls for IEEE Journals}
	
	%%%%% 要不要？\IEEEpubid{0000--0000/00\$00.00~\copyright~2021 IEEE}
	\IEEEpubidadjcol
	\maketitle	
	\begin{abstract}
		
		Evaluating the performance of Simultaneous Localization and
		Mapping (SLAM) algorithms is essential for scientists and
		users of robotic systems alike. But there are a multitude of different permutations of possible options of hardware setups and algorithm configurations, as well as different datasets and algorithms, such that it was previously infeasible to thoroughly compare SLAM systems against the full state of the art. To solve that we present the SLAM Hive Benchmarking
		Suite, which is able to analyze SLAM algorithms in 1000’s of mapping runs, through its utilization of container technology and
		deployment in the cloud. This paper presents the architecture
		and open source implementation of SLAM Hive and compares
		it to existing efforts on SLAM evaluation. We perform mapping runs with popular visual, RGBD and LiDAR based SLAM algorithms against commonly used datasets and show how SLAM Hive can be used to conveniently analyze the results against various aspects. Through this we envision that SLAM Hive can become
		an essential tool for proper comparisons and evaluations of
		SLAM algorithms and thus drive the scientific development in
		the research on SLAM. The open source software as well as a demo showing the live analysis of more than 1700 mapping runs can be found on \url{https://slam-hive.net}.
	\end{abstract}
	
	
	\section{Introduction}
	\IEEEPARstart{S}{imultaneous} Localization and Mapping (SLAM) is an essential capability for many mobile robotic systems. Consequently, there is lots of research work on SLAM, utilizing various types of sensors and algorithms and employing these to various application scenarios \cite{cadena2016past}. In order to scientifically evaluate the performance of SLAM systems, experiments with them must be reproducible, repeatable and properly compared to other solutions.
	% (key word: evaluation; ) 
	
	Running SLAM algorithms with pre-recorded datasets is essential for this, as it allows to repeatedly perform mapping runs with the exact same data. Using ground-truth path information for said datasets it is possible to quantitatively evaluate the localization error of the mapping run, which is generally considered a sufficient measure for the performance of SLAM algorithms \cite{fornasier2021vinseval}. 
	%(key word:exact datasets; groundtruth;)
	
	Furthermore, methods should not only be evaluated against the Absolute Trajectory Error (ATE) and Relative Pose Error (RPE) w.r.t. the ground truth path, since memory consumption, processing time or map quality can also be important factors to consider. 
	%(key word: evaluation metrics)

	SLAM is a complex topic and the performance of an approach depends on various factors, including the type of environment, the path the robot took in that environment, sensor types, sensor placement on the robot, settings for the sensor data like frame rate, resolution or maximum range, as well as various configuration parameters for the algorithm like number of particles, number of features or various other thresholds and settings. 
	
	A robotics engineer who wants to deploy a SLAM system for a specific application should be able to select the best SLAM algorithm, given the number and quality of sensors and computing resources his scenario allows for. Likewise, scientists developing SLAM software should be able to evaluate the performance of their solution under various aspects and configurations. But the number of mapping runs needed to  exhaustively test an algorithm under many permutations of configurations is very big - e.g. comparing 10 algorithms with 5 different configuration each, 10 sensor combinations (e.g. front camera or back camera or both, etc.), 4 different resolutions, 4 different frame rates and 10 different datasets with different environments and amounts of loops, requires 80,000 mapping runs. But currently papers on SLAM only test with and against very few mapping runs. 
	%(keyword: many influence to slam application ; big parameter space; parameter of algorithm, dataset and configuration)
	
	\begin{figure}[t]
		\centering
		\includegraphics[width=3.5in]{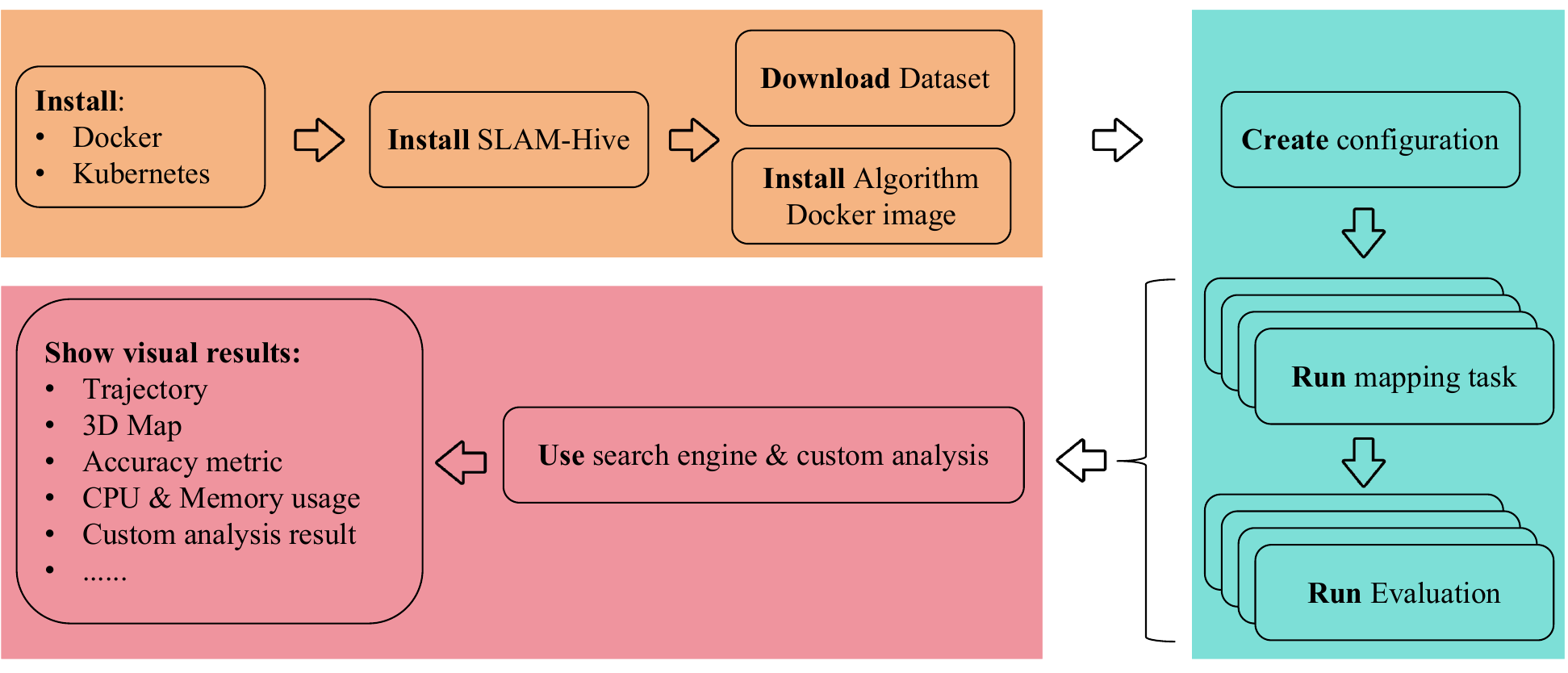}
		\caption{SLAM Hive Benchmarking Suite Workflow}
		\label{fig: newworkflow}
			\vspace{-0.6cm}
	\end{figure}

	Performing that many mapping experiments and analyzing them by hand is obviously impossible. In this paper we provide the SLAM Hive Benchmarking Suite, whose workflow is outlined in Fig. \ref{fig: newworkflow}. It provides solutions to the following challenges:  1) Such a software must be able to work with various SLAM algorithms, that may use different system environments or even operating systems. 2) The system also must provide some kind of solution to feed various datasets to the algorithms in a standard way. 3) Also, there must be some convenient way to configure algorithms and datasets systematically. 4) Then we must run the mapping tasks, ideally with support to run on a workstation but also in a cloud environment. 5) The next problem is, that the results of the mapping runs need to be evaluated, typically against a ground truth path, but potentially also w.r.t. other aspects, such at computation resources or map quality. 6) Finally, we need to be able to analyze all the results, being able to compare various aspects of the mapping runs with each other.  To the best of our knowledge, SLAM Hive is the only system available solving all these problems.

	\begin{figure*}[t]
		\centering
		\includegraphics[width=6in]{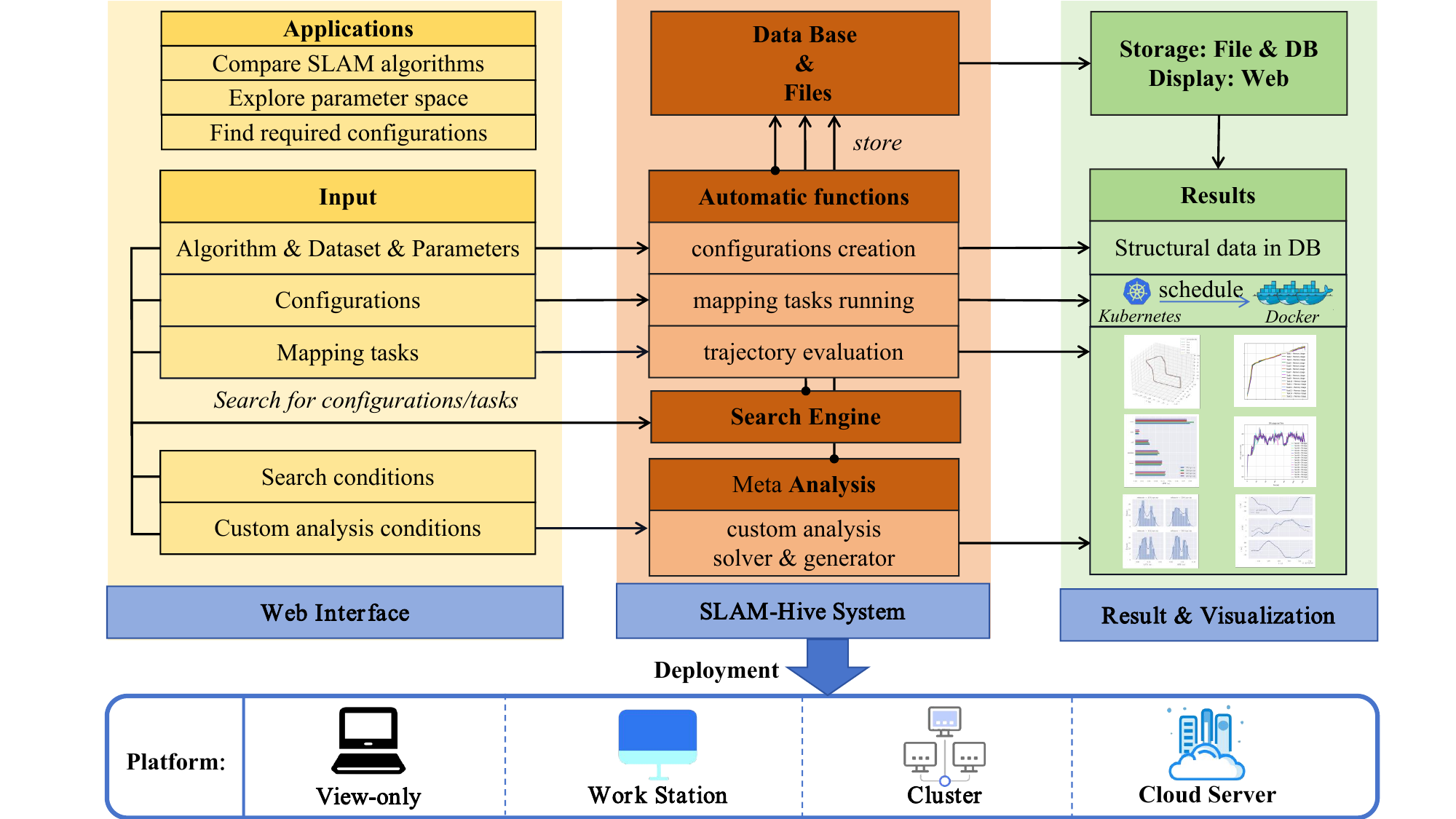}
		\caption{SLAM Hive Benchmarking Suite Overview}
		\label{fig: new overview}
			\vspace{-0.3cm}
	\end{figure*}

	Fig. \ref{fig: new overview} shows an overview of the SLAM Hive Benchmarking Suite, which was first introduced with basic features in \cite{yang2023slam}. 
    We provide a cross-platform SLAM algorithm execution and evaluation software system, deployed as a Docker container. It can be configured to be deployed  on a workstation, on a cluster or the cloud, and as a view-only version for disseminating the results on the web without the ability to start new mapping runs. It provides a user-friendly web interface for configuration of the mapping tasks, inspecting the individual mapping results and comprehensively analyzing and comparing all the data.

	We invite interested researchers to use and contribute to SLAM Hive, which is available on GitHub \footnote{\url{https://github.com/SLAM-Hive}} and has a live demo on \url{https://slam-hive.net/}. 
	
\begin{figure}[t]
		\centering
		
		\captionsetup[subfloat]{font=scriptsize}
		
		\subfloat[Configuration search result]{
			\label{fig:subfig:screenshot1} 
			\includegraphics[width=0.48\linewidth]{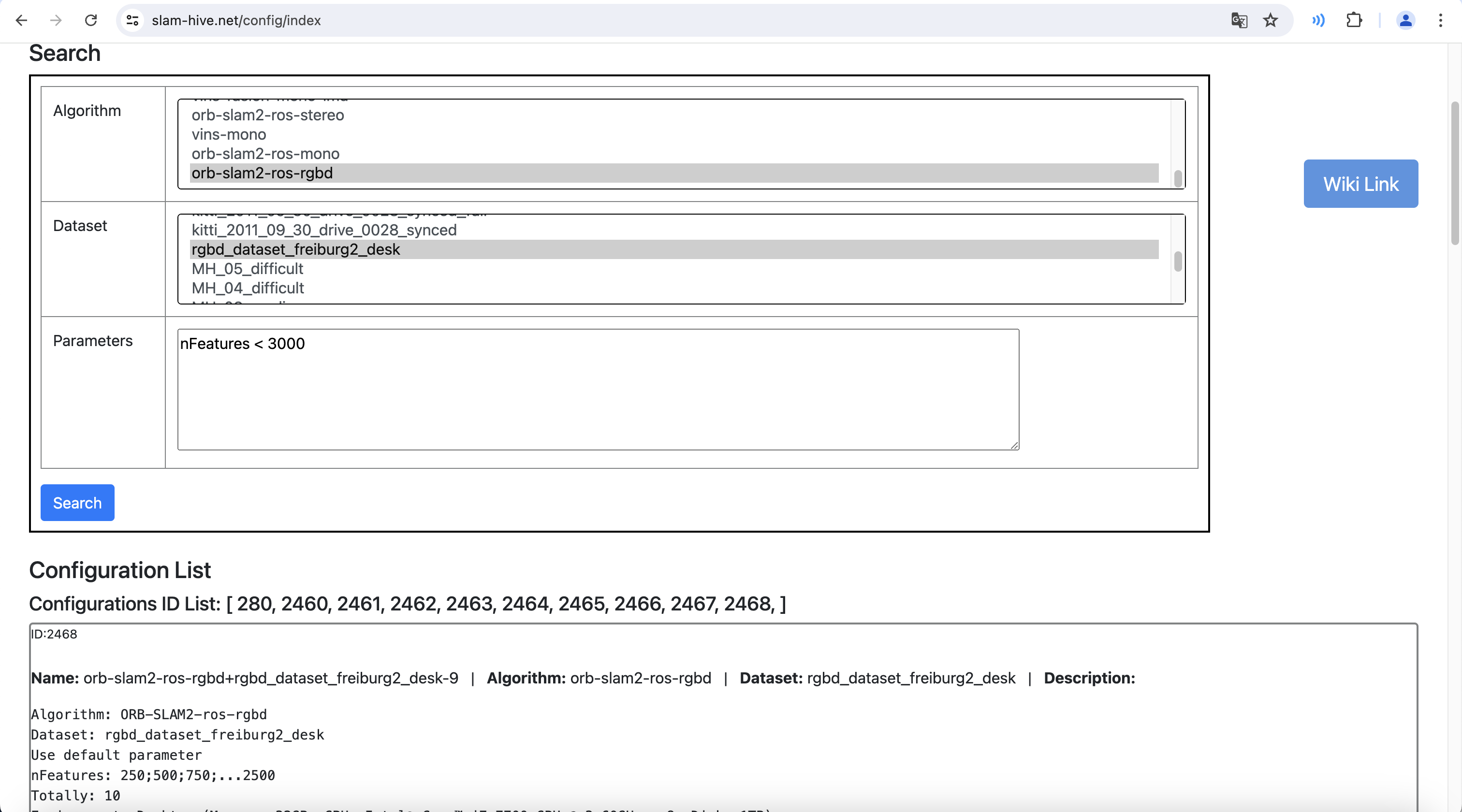}}
				\subfloat[Mapping result display]{
			\label{fig:subfig:screenshot2} 
			\includegraphics[width=0.48\linewidth]{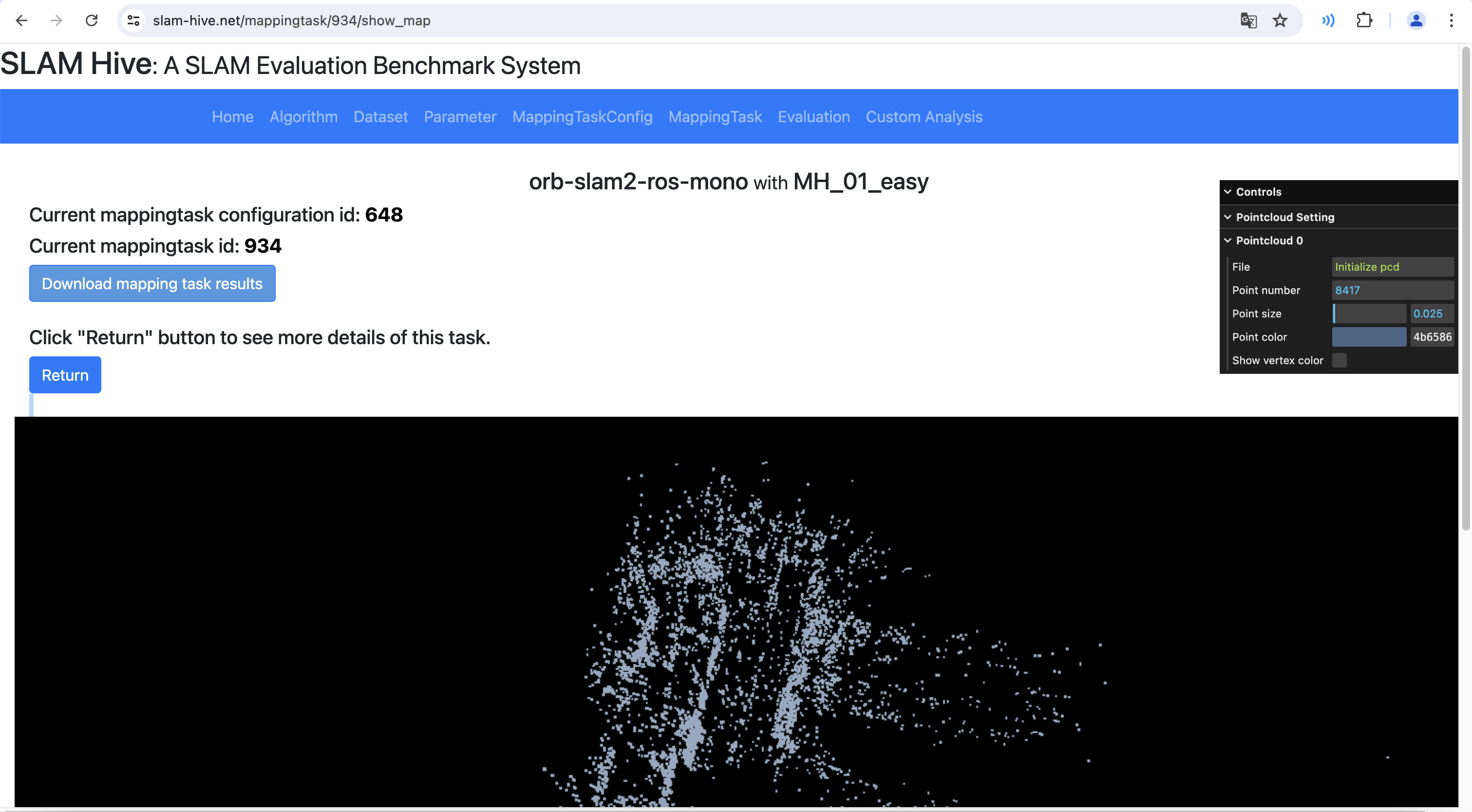}}
		
		\subfloat[3D scatter: Image Resolution vs Image Rate vs ATE Mean]{
			\label{fig:subfig:screenshot3} 
			\includegraphics[width=0.48\linewidth]{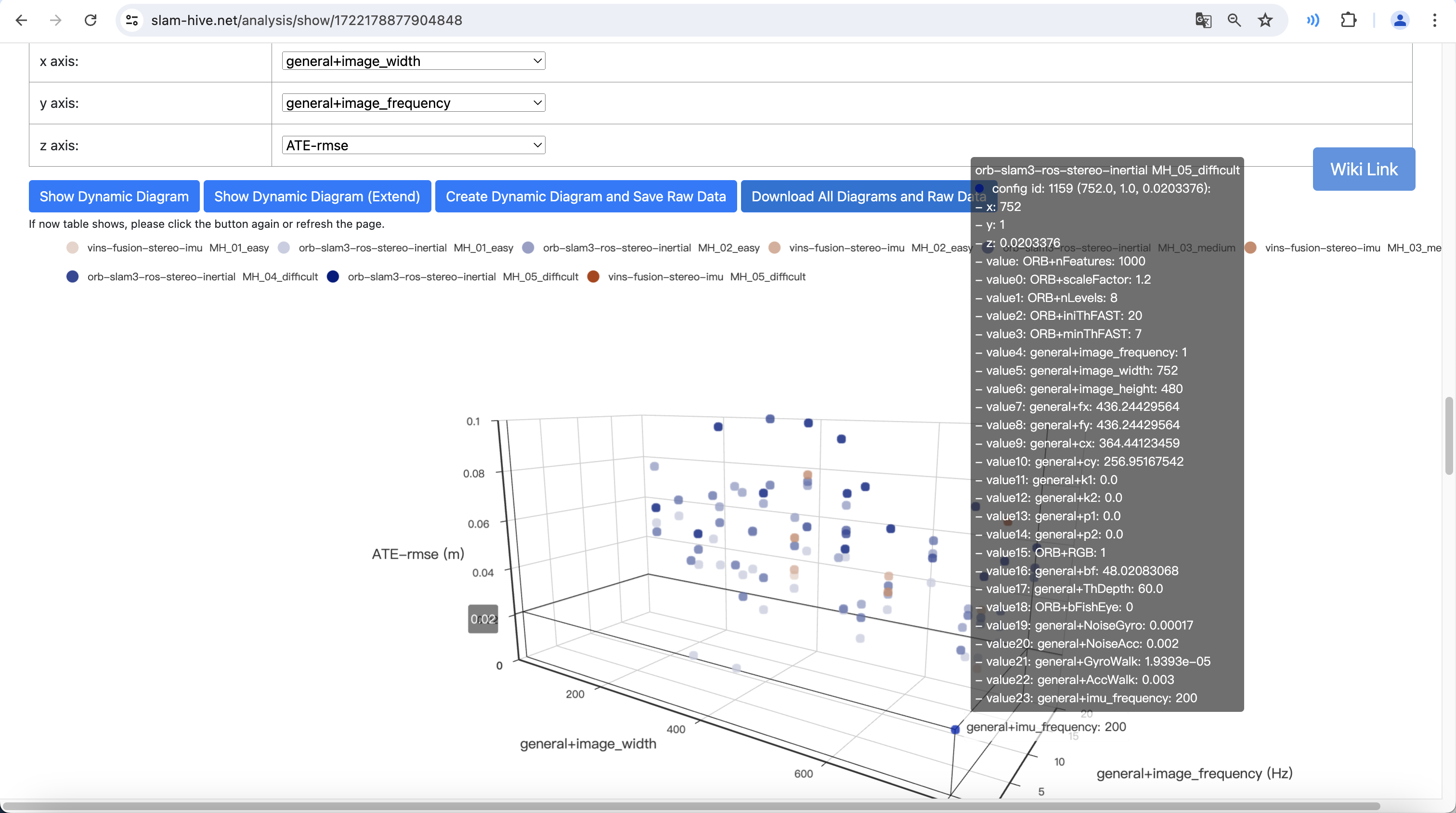}}
		\subfloat[Trajectory results comparison]{
			\label{fig:subfig:screenshot4} 
			\includegraphics[width=0.48\linewidth]{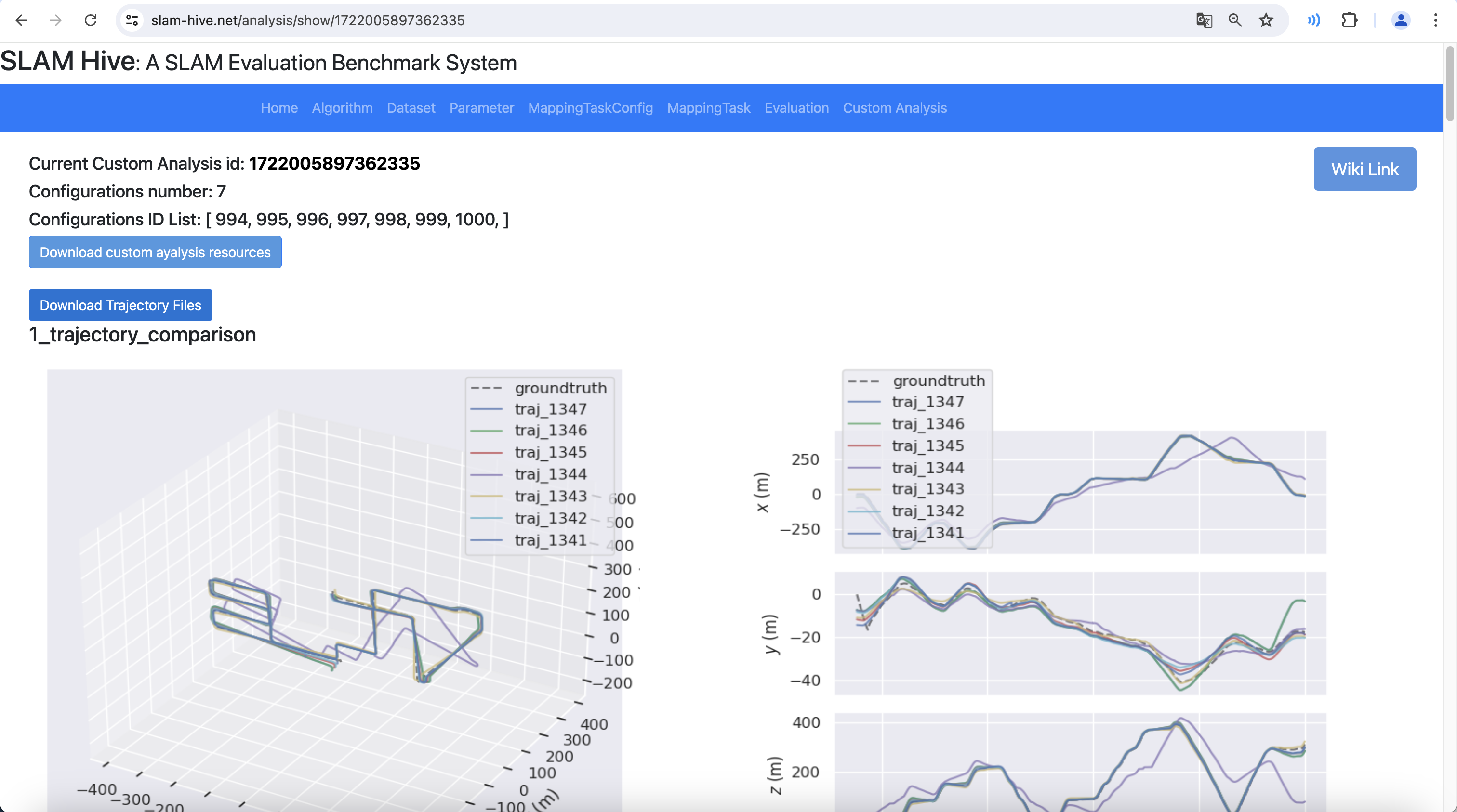}}

		\caption{The SLAM Hive system web interface.}
		\label{fig: screenshot} 
		
	\end{figure}

	The key contributions of the SLAM Hive Benchmarking Suite and this paper are:
	\begin{itemize}
		\item Providing a user friendly web interface that enables users to configure and run hundreds of mapping runs in one go. 	Fig \ref{fig: screenshot} shows some screenshots of SLAM Hive system.
		\item Defining and implementing a Docker-based cross-platform framework performing mapping runs of various SLAM Algorithms using various configurations and datasets.
		\item Next to a view-only and stand-alone workstation mode, providing a Kubernetes-based cluster mode that can distribute mapping runs over many compute nodes in a cluster. The cloud mode then extends this further to dynamically allocate resources in the cloud, here implemented with Alibaba's Aliyun cloud. 
		\item Implementing a Docker-based SLAM evaluation process that is analyzing the results of the mapping runs by comparing the paths against a ground-truth path. 
		\item Providing a custom analysis page, where users can, even in view-only mode, perform and export in-depths analysis of the evaluation results w.r.t. to the configuration parameters and algorithm and dataset attributes. 
		% in experiments, include 1900+ tasks
		\item We performed 1716 mapping runs with SLAM Hive, on 9 different SLAM algorithms and 3 different datasets with 7 different sequences. We provide the results for download, so users can analyze the them with their workstation SLAM Hive. On  \url{https://archive.slam-hive.net/} we show the results and analysis of those 1716 mapping runs.
	\end{itemize}
	
	A basic version of SLAM Hive was presented at IEEE ICRA 2023 \cite{yang2023slam}. Following the evolutionary publication model, this journal paper extends the previous publication significantly:
	\begin{itemize}
		\item In \cite{yang2023slam} we provided the workstation mode, where all mapping runs are performed sequentially on the same machine. Now, we have implemented a view-only mode for deployment on a web server, such that users can view mapping evaluations and perform analysis without being able to start new mapping runs.
		\item Furthermore, we implemented the Kubernetes-based cluster mode of SLAM Hive, capable of orchestrating a multitude of nodes to perform the mapping runs, managing the deployment of datasets and algorithm containers and centrally collecting the results. 
		\item Also, we provide code to perform cloud management - on Aliyun - dynamically purchasing and configuring nodes as needed. 
		\item The novel analysis system allows to systematically analyze mapping runs under various aspects, in order to be able to draw quantitatively substantiated scientific conclusions from the experiments. 
		\item We perform many more experiments with SLAM Hive compared to the conference paper, already drawing interesting conclusions, and also provide these results for analysis and download on \url{https://archive.slam-hive.net/}. 
	\end{itemize}

	\section{Related Work}
	\subsection{SLAM Datasets}
	
	Public datasets play a fundamental role in the evaluation of Simultaneous Localization and Mapping (SLAM) algorithms. This is because by using datasets, experiments with SLAM algorithms are easily repeatable and reproducible at a very low cost and with exactly the same data. The motivation behind SLAM Hive is to enable to compare most, if not all, SLAM algorithms with each other. 
	For a valid comparison, ideally, the same dataset should be used for all algorithms. However, different algorithms have unique requirements for sensor types and configurations. 
	While datasets like \cite{chen2020advanced} and \cite{Yin2022M2DGR} provide large-scale data with diverse sensor types and an extensive range of sensors, they still fall short of meeting all criteria. 
	An ideal dataset would offer data from a full range of sensors across varied configurations and viewing directions, include ground truth paths, and cover diverse environments and conditions. 
	Although such a comprehensive dataset does not yet exist, we are actively working on collecting one using the ShanghaiTech Mapping Robot \cite{xu2024shanghaitechmappingrobotneed, lin2024collecting}.
	
	% 一些常用的数据集
	Nevertheless, SLAM Hive can also utilize less comprehensive datasets. 
	At present, numerous widely recognized public datasets are available to researchers, such as the TUM RGB-D dataset \cite{sturm2012benchmark} which includes depth values, the KITTI dataset \cite{Geiger2013IJRR} often employed in autonomous driving, and the EuRoC MAV dataset \cite{burri2016euroc} acquired using unmanned aerial vehicles (UAVs).
	The TUM VI benchmark \cite{schubert2018tum} introduced a distinctive dataset showcasing diverse sequences in various environments designed to assess Visual Inertial Odometry (VIO) algorithms. 
	% To validate its benchmarking potential, this study features the outcomes produced by three state-of-the-art open-source methods- ROVIO, OKVIS, and VINS-Mono. The obtained outcomes indicate the presence of significant drift, even under long-distance and visually demanding sequences, thus providing researchers with a suitable dataset for evaluating high-performing algorithms \cite{schubert2018tum}. It is clear that the development of SLAM necessitates more rigorously demanding datasets to effectively evaluate SLAM systems.
	
	% 最近；为某些特定任务服务的传感器；各种专门场景下的数据集
	% 最近有很多数据集，为了专门的任务或场景而设计
	
	Among the recently released datasets, some are designed specifically for certain tasks or scenarios. 
	\cite{shi2020we} published the Openloris-Scene, a dataset designed specifically for service robots.  % 服务机器人 在各种复杂的场景下收集了极具挑战性的数据集
	\cite{wang2020tartanair} presented TartanAir, a challenging dataset collected in realistic simulated environments with various dynamic factors.  % 模拟场景
	\cite{Benseddik2020PanoraMIS} developed PanoraMIS, a dataset utilizing a panoramic camera to capture a 360-degree view of the environment.  % 全景相机
	\cite{Cheng2021Are} proposed the USVInland Multisensor Dataset, specifically designed for underwater missions.  % 水下
	\cite{Giubilato2022} collected a challenging dataset in a volcanic environment using a handheld sensor. % 手持设备在火山环境
	\cite{Ramezani2020College}, \cite{Zhang2021Multi}  introduced a large dataset, which used handheld devices to collected data in school scene (Oxford University New College). % 学校场景
		% 具有挑战性
	\cite{Helmberger2022Hilti}, \cite{Zhang2023Hilti} introduced dataset containing sequences from varying complex and challenging environments.
		
	% 自动驾驶；汽车收集
	Numerous autonomous driving datasets have surfaced alongside the well-known KITTI dataset \cite{Geiger2013IJRR}. Among them are UrbanLoco \cite{Wen2019UrbanLoco}, DSEC \cite{Gehrig2021DSEC}, Oxford Radar \cite{Barnes2020Oxford}, Audi Autonomous Driving Dataset A2D2 \cite{Jakob2020A2D2}, Waymo Open Dataset \cite{Sun2020Waymo}, NuScenes \cite{Caesar2020nuScenes} and Brno Urban Dataset  \cite{Ligocki2020Brno}.

	\begin{table*}[t]
		\caption{SLAM Benchmarking systems comparison}
		\label{table: benchmark list}
		\begin{center}
			\begin{threeparttable}
				\begin{tabular}{l|l|l|c|c|c|c|c|c}
					\hline
					System & Dataset Support & Data Types &  \makecell{Run \\ any alg. \\ in Docker} & \makecell{Monitor\\CPU \\ \& \\Memory} &  \makecell{Automated \\Evaluation} & \makecell{Meta\\Analysis}  & \makecell{Deploy in \\Cluster} & Year \\ 
					\hline
					RAWSEEDS \cite{Bonarini_2006_IROS} & RAWSEEDS & mono, stereo, IMU & & & && &2006\\
					KITTI Vision \cite{Geiger2012CVPR} & KITTI dataset\cite{Geiger2013IJRR} & mono, stereo, IMU, LiDAR  & &&   & && 2012\\
					TUM RGB-D \cite{sturm2012benchmark} & TUM RGB-D &  RGB-D, IMU & && && & 2012 \\
					ICL-NUIM RGB-D \cite{burri2016euroc}  & Simulation data & mono, stereo, RGB-D, IMU &  & & \checkmark & \checkmark & & 2016\\
					EVO \cite{grupp2017evo} & \makecell[l]{Trajectories meet\\ specific formats\tnote{*}} & \makecell[l]{mono, stereo, RGB-D, IMU,\\ LiDAR, etc. } & & &&& & 2017\\
					TUM VI \cite{schubert2018tum}  & Simulation data & mono, stereo, RGB-D, IMU &  & & \checkmark & \checkmark & & 2018\\
					SLAMBench \cite{nardi2015introducing,bodin2018slambench2,bujanca2019slambench}& 9 public datasets & mono, stereo, RGB-D, IMU & &  \checkmark & \checkmark &&& 2019\\
					GSLAM \cite{gslamICCV2019} & 9 public datasets & mono, stereo,  RGB-D, IMU & & \checkmark & \checkmark  & &&2019\\
					VINSEval \cite{fornasier2021vinseval}  & Simulation data & mono, stereo, RGB-D, IMU &  & & \checkmark & \checkmark & & 2021\\ 
					%论文里提到了 但是没有写到表格里（yuanyuan姐的硕士毕业论文里没有这个 是后加的）
					
					%gslam中提到的 可以加进去（to do 也可以考虑加入dataset）
					\hline
					\textbf{SLAM Hive} & any ROS bag & \makecell[l]{mono, stereo, RGB-D, IMU,\\ LiDAR, etc. }& \checkmark & \checkmark & \checkmark & \checkmark & \checkmark & 2024 \\
					\hline
				\end{tabular}
				\begin{tablenotes}
					\scriptsize
					\item[*] EVO is to evaluate the trajectory output by the algorithm. The supported trajectory formats are: 'TUM' trajectory files, 'KITTI' pose files, 'EuRoC MAV', ROS and ROS2 bagfile with geometry\_msgs/PoseStamped, geometry\_msgs/TransformStamped, geometry\_msgs/PoseWithCovarianceStamped or nav\_msgs/Odometry topics or TF messages.
				\end{tablenotes}
			\end{threeparttable}
		\end{center}
	\end{table*}
	
	\subsection{SLAM Evaluation Systems}
	
	% 不同系统的总结： different system overall
	% RAWSEEDS
	%  - early的toolkit
	%  - 文章提出了evaluate的必要性
	%  - 优点：提供了高质量的 多传感器的数据集；描述了基准测试的基本问题
	%  - 缺点：缺少可重用性和自动化
	% KITTI Vision Benchmark Suite
	%  - 适用算法：Vision 
	%  - 评估精度：双目 光流 里程计 物体识别
	%  - 只能使用kitti数据集 
	%  - 用户提供运行轨迹（kitti tools本身不运行算法）
	% TUM RGB-D 2018
	%  - 提供了RGBD数据集
	%  - 提出了ate rpe
	%  - evo
	% slambench系列
	%  - 提出将精度 与 时间；能耗结合
	%  - 2 ？
	%  - 3 ？
	% tum 4 (没看过) 2018
	% ICL-NUIM（论文里没写，但是也是一个evaluation+dataset的system 2014）
	% vins eval
	%  - 生成数据集
	%  - vi系统的评估
	% gslam
	%  - 帮助用户开发 （和slamhive一样 都是提供了运行环境）
	%  - 评估
	%  - 缺点：需要用户按照gslam提供的规范开发slam框架，可能会降低灵活性；同时评估已有的算法可能会有问题
	% - 侧重点在于slam算法的开发 
	%   在本文中，介绍了一个新型的SLAM平台，不仅提供评估功能，还为研究人员提供了有用的工具，以快速开发他们自己的SLAM系统。通过这个平台，常用功能以插件形式提供，因此用户可以直接使用它们或创建自己的功能以实现更好的性能。
	%   为了更容易实现一个SLAM系统，GSLAM提供了一个实用类。
	
	% 这玩意好像也可以运行在docker上，但是 any algo（表示不行 需要修改）

	Until now, evaluation systems for a number of SLAM algorithms have emerged. 
	Proposed by \cite{Rawseeds2014Fontana,Ceriani_2009_AutonomousRobots,Bonarini_2006_IROS}, the RAWSEEDS project is an early benchmarking toolkit focused on sensor fusion, localization, mapping, and SLAM evaluation for autonomous mobile robots. 
%	The paper emphasizes that the lack of benchmarks hampers scientific recognition and innovation, making robotics less accessible due to high entry costs. RAWSEEDS provides quality multi-sensor datasets and customizable benchmarking solutions. 
%	Despite the absence of complete automation capabilities and modern sensors like 3D LiDARS or event or RGBD cameras, it still remains as an initial reference for assessing the scientific achievements in SLAM.
	
	The famous  KITTI Vision Benchmark Suite \cite{Geiger2012CVPR} provides a set of tools that can evaluate the accuracy of the stereo, optical flow, odometry and object recognition, but it can only use the KITTI dataset of the autonomous driving scenarios they provide and only evaluates the path provided by the user, so it doesn't run the SLAM algorithm itself.
	
	\cite{sturm2012benchmark} provides a RGB-D dataset and automatic evaluation tools both for the evaluation of drift of visual odometry systems and the global pose error of SLAM systems.  
	The two well known metrics: RPE and ATE estimating trajectory instead of the resulting map were proposed in this paper. 
	% For both evaluation metrics, they provided easy-to-use evaluation scripts for the users. 
	On this basis, EVO provides executables and a small library for handling, evaluating and comparing the trajectory output of odometry and SLAM algorithms \cite{grupp2017evo}. 
	
	% SLAMBench系列论文提出了一系列SLAM evaluation systems，指出需要在评估算法精度的同时兼顾 execution time and energy consumption，并且提出要针对不同的算法进行相应的全面评估。
	SLAMBench \cite{nardi2015introducing,bodin2018slambench2,bujanca2019slambench} proposes a series of SLAM evaluation systems, emphasizing the need to consider execution time and energy consumption in addition to algorithm accuracy. It suggests that a comprehensive evaluation should be conducted for different SLAM algorithms. SLAMBench needs algorithms to be linked against the software, restricting the compatibility. Also, it delivers the next image when the algorithm is ready, more like a 3D reconstruction approach, while SLAM Hive plays the data in real-time, thus being closer to the real application scenario. 
	
	% Although SLAMBench, SLAMBench2 and SLAMBench3 are as comprehensive as possible and support plug and play, it is not as general SLAM Hive w.r.t. adding algorithms, as those need to be linked against SLAMBench, while SLAM Hive runs them in their environment in a container - SLAM Hive could theoretically even support algorithms coded in MatLab and/ or other operating systems.  Moreover, SLAMBench only supports camera (and IMU) data, no LiDAR or other data. 
	
	% The TUM VI benchmark published a novel dataset with a diverse set of sequences in different scenes for evaluating visual-inertial odometry (VIO). 
	% To verify that the dataset is suitable for benchmarking, it provides the results of three open-source state-of-the-art methods: ROVIO, OKVIS and VINS-Mono,  which shows that even the best performing algorithms have significant drift in long (magistrale, outdoors) and visually challenging (slides) sequences \cite{schubert2018tum}. 
	The TUM VI benchmark published a novel dataset with a diverse set of sequences in different scenes for evaluating visual-inertial odometry (VIO). 
	The results of several state-of-the-art methods on this dataset show that even the best-performing algorithms experience significant drift in long (e.g., magistrale, outdoors) and visually challenging (e.g., slides) sequences \cite{schubert2018tum}, highlighting the urgent need for more challenging datasets for SLAM system evaluation.
	%It can be seen that the development of SLAM urgently needs more challenging datasets for SLAM systems evaluation.
	
	GSLAM \cite{gslamICCV2019} introduce a novel SLAM platform which provides not only evaluation functionality, but also useful toolkit for researchers to quickly develop their own SLAM systems. 
	% GSLAM implements a set of interfaces which makes it easy to interact with different datasets, SLAM algorithms, common optimized modules and applications with plugin forms in an unified frameworks.
	
	VINSEval  \cite{fornasier2021vinseval} presented a unified framework for statistical relevant evaluation of consistency and robustness of Visual-Inertial Navigation System (VINS) algorithms in simulation environment, with fully automated scoreboard generation over a set of selectable attributes. 
	% VINSEval uses FlightGoggles, a development environment envisioned to allow the design, implementation, testing and validation of autonomous super vehicles, to generate simulation data with different IMU noisy, amount features, illumination and time delay instead of the real dataset.
	
	In \cite{jorge2024impact} the impact of 3D LiDAR Resolution in SLAM is investigated and they found, that in fact the resolution does influence the accuracy and computation time, showing the need for such evaluations in SLAM Hive. 
	
	% TODO topo map如何修改删减
	All the above mentioned methods evaluate SLAM systems based on a ground truth path. In contrast to that, there are also methods that aim to compare the created maps against ground truth maps. In \cite{schwertfeger2016map} we matched the TopologyGraphs, a variant of the Voronoi Diagram, of a robot generated map with a ground truth map to estimate the relative and absolute accuracy as well as the coverage. The work in \cite{hou2022matching} we extended the matching to utilize a topometric representation called Area Graph. In general, map matching methods like \cite{shahbandi20192d} and \cite{saeedi2014map} can also be used to evaluate map quality. Specially shaped objects, fiducials, can be placed in the environment to be recognized in the maps and then used for map evaluation in 2D \cite{schwertfeger2010evaluation} and 3D \cite{schwertfeger2015using}. The advantage of such approaches is, that no ground truth path is required, just a ground truth map, so the performance of different mapping runs/ different datasets can still be evaluated in the same system. Also, it is often easier to aquire a ground truth map (e.g. with a static 3D scanner) than a ground truth path. Furthermore, many algorithms rely on maps, so directly measuring their quality instead of indirectly just measuring the localization accuracy and hoping that the localization accuracy correlates with the map quality, is advantageous. But map evaluation is more complicated to perform. This is why initially SLAM Hive will provide only SLAM evaluation via the comparison of the estimated paths with the ground truth path. But SLAM Hive is already designed to support multiple mapping run evaluation methods and running the evaluations in Dockers. So we plan also integrate map-based evaluation into SLAM Hive in the future. 
	
	% \textbf{to do}: Add some other latest SLAM evaluation system
	% \begin{enumerate} 
		%     \item G-SLAM(重新读一下论文，根据上面的形式总结) （数据集 算法 用法；通用/特殊 优点 缺点 应用场景）加上了
		%      ICL-NUIM RGB-D Benchmark Dataset（也是一个rgbd格式的数据集，可以考虑加进来）
		% \end{enumerate}
	
	Table \ref{table: benchmark list} shows an overview the existing SLAM evaluation systems. According to the studies mentioned above, it can be found that KITTI Vision Bnechmark Suite \cite{Geiger2012CVPR}, RGB-D SLAM systems benchmark \cite{sturm2012benchmark} and the TUM VI benchmark \cite{schubert2018tum} all publish a dataset and provide tools or easy-to-use scripts for evaluating the output of existing SLAM methods. All of them only support using the dataset they provide, which limits the breadth of evaluation of different SLAM algorithms. The SLAMBench series \cite{nardi2015introducing,bodin2018slambench2,bujanca2019slambench} provided more evaluation metrics, supported many algorithms and datasets, but is limited to SLAM algorithms linked against it. Also, it does not support Meta Analysis of many mapping runs, nor performing or creating them automatically. GSLAM\cite{gslamICCV2019} provides a unified platform for both SLAM system evaluation and development, but SLAM algorithms have to follow its API, so its scope is very limited. VINSEval \cite{fornasier2021vinseval} provides a robust evaluation method for the visual SLAM community, however it is difficult to ensure the correlation of simulated data with real data and they don't support other sensor data like LiDARs. 
% TODO	
%	\subsection{SLAM Methods}
%	This section may not be necessary
%	\subsection{Evaluation Metrics}
%	This section may not be necessary

	\section{SLAM Hive}

	We implement a cross-platform automatic SLAM algorithm evaluation software named ``SLAM Hive Benchmarking Suite'', which is, 
	to the best of our knowledge, the first cross-platform SLAM benchmarking framework that can be deployed across different types of hardware environments. A system overview is provided in Fig. \ref{fig: new overview}.
		
	The system is fully deployed in a Docker container. The software is written in Python with the Flask web framework. Interacting with the web interface, users configure mapping runs, w.r.t. dataset and algorithm configuration parameters (e.g. select what data to use, its frame rate and resolution, and algorithm parameters such as number of particles or number of features). SLAM algorithms also run in Docker containers, orchestrated by Kubernetes when running in a cluster or the cloud. Standard interfaces (mainly yaml configuration file format and storage conventions) ensure that the scripts for dataset playback and running the algorithms use the correct configurations and output the results in a format usable by SLAM Hive. Results are stored as files and later in a database. The trajectories and other results of mapping runs are then evaluated against ground truth data with separate  Docker-based evaluation software, again following standard interfaces and storing results in the database and as files. The Meta Analysis module finally allows to analyze and visualize the results of a set of mapping runs (obtained via the search functionality) w.r.t. to various parameters, e.g. plotting 3D diagrams of accuracy vs. max. CPU usage vs. frame resolution for visual SLAM algorithms. The details of the various components for the basic workstation mode of operation will be introduced in the following sections. Afterwards, more details on the different modes of operation (view-only, workstation, cluster, and cloud) are described.

	\begin{table*}[t]
		\caption{Supported SLAM methods and datasets of SLAM Hive}
		\label{tab: support_list}
		\centering
		%	\scriptsize
		\setlength{\tabcolsep}{4pt}% column separation
		\renewcommand{\arraystretch}{1.2}%row space 
		\begin{center}
			\begin{tabular}{l|c|c|c|c|c|c|c}
				\hline
				Method & Monocular & Monocular Inertial & Stereo & Stereo Inertial & RGBD & Lidar & Lidar Inertial\\ 
				\hline 
				%			OKVIS \cite{article} &  & BRISK feature &  &  &  & \checkmark &  & & 2015\\
				ORB-SLAM2 \cite{mur2017ORB2}  & \checkmark &  & \checkmark &  & \checkmark & & \\
				VINS-Mono \cite{qin2018vins-mono}   &  & \checkmark &  & & & &\\
				% DSO \cite{Engel2018direct}  & \checkmark  & & &  &  &  &\\
				VINS-Fusion \cite{qin2019vins-fusion} &   & \checkmark & \checkmark & \checkmark &  & & \\
				ORB-SLAM3 \cite{campos2021orb3} & \checkmark & \checkmark & \checkmark & \checkmark & \checkmark & &\\ \hline
				%			\hline
				LIO-SAM \cite{liosam2020shan} &  &  &  & &  & & \checkmark\\
				%			Cortograhpher &  & x & x & x & x & x  & x & \\
				LOAM \cite{zhang2014loam} &  &  &  & &  & \checkmark & \\
				A-LOAM \cite{zhang2014loam} &  &  &  & &  & \checkmark & \\
				LeGO-LOAM \cite{legoloamIROS2018} &  &  &  & &  & \checkmark & \checkmark\\
				NDT-LOAM \cite{ndtloamSJ2022} &  &  &  & &  & \checkmark& \\
				% to do: other slam algorithms 
				%  - solid-state lidar 固态雷达的相关算法 2个（数据集没有 no groudtruth？？？）
				%  - dso
				%  - svo
				\hline
			\end{tabular}
		\end{center}
		\begin{center}
			\begin{tabular}{l|c|c|c|c|c|c|l}
				\hline 
				Dataset & Grayscale camera & RGB camera & RGB-D camera & IMU & GPS & Lidar & Groundtruth\\ 
				\hline 
				TUM RGBD \cite{sturm2012benchmark} & & & 1 & \checkmark &  & & Tracking System \\
				KITTI \cite{Geiger2013IJRR} & 2 & 2 & & \checkmark & \checkmark & \checkmark & Pose via GPS \\
				EuRoC \cite{burri2016euroc} & & 2 & & \checkmark & & & Tracking System \\
				% TODO
				% ICL-NUIM RGB-D(to do) \cite{burri2016euroc} & & 2 & & \checkmark & & & Tracking System \\
				%livox (solid-state lidar)(to do) \cite{burri2016euroc} & & 2 & & \checkmark & & & Tracking System \\
				%			MARS MAP &RGB-D & \checkmark & x\\
				\hline
			\end{tabular}
		\end{center}
	\end{table*}

	\subsection{SLAM Hive Web}
	
	The graphical user interface is provided by the web controller, which is built using the lightweight web application \textit{Flask}. 
	It handles user interactions like the creation of mapping and evaluation tasks, facilitates the addition of new algorithms, datasets, parameters, and configurations as well as displaying the results of mapping runs\footnote{\url{https://archive.slam-hive.net/mappingtask/934/show_map}}, mapping evaluations and the analysis. Every page has a link to the SLAM Hive Wiki with detailed help regarding that page.

	\subsection{Standard Interface}
	
	Standard interface is an important guarantee that the system has the ability to expand and support many different algorithms.
	% Our interface consists of standard containers, unified configuration format, algorithm execution scripts, dataset component and standard trajectory format, which are documented on github.
	% 标准容器：保证了算法都运行在相同的环境中
	% 统一的配置文件格式
	%% 基于该配置格式，衍生出用于批量创造config的方式
	% 算法运行脚本（slamhive 文件）
	% 数据集 + 轨迹格式
	
	\subsubsection{Standard Container}
	It's difficult to create an unified environment for all algorithms, since different SLAM software require different system environments. Docker is an open source application container engine that can package applications and dependencies into lightweight containers. 
	We provide many SLAM methods and an evaluation Docker image and all mapping runs and evaluation processes on SLAM Hive are executed in Docker containers. 
	In order to ensure the validity of already performed mapping runs, containers may not be modified after being added to SLAM Hive - if changes are needed one will have to add a new algorithm container. 
	Currently, all our algorithms support ROS 1 as sensor data input system, either out of the box or through slight modifications by us. 
	But if needed, the algorithm script (which is included with the algorithm Docker) might also extract the data from the ROS data stream to feed it to the algorithm in the format it requires.

	\subsubsection{Unified Configuration Format}

	Most of the SLAM algorithms utilize configuration files in some standard format (such as yaml format), for users to setup the parameters of algorithm, intrinsic and extrinsic parameters of datasets, etc. 
	When we use ROS to play the dataset, the ROS topic remapping configuration is needed in order to stream the correct data to the algorithm, which also facilitates selecting which sensor(s) from the dataset to use. 
	At the same time, researchers and engineers may have big scale datasets with high framerates and high resolution data, and they may wish to reuse it to test the performance and robustness of the algorithms by reducing the framerate and/ or resolution to simulate lower quality sensors.
	We thus implemented a \emph{key-value} format, for users to create all of above parameters.
	
	%大多数的算法都会为自己的算法提供一个配置文件（yaml file），用户通过该文件来设置算法内部的参数的数值，以及数据集的内参和外餐
	%同时，当我们有了一个具有高分辨率和高频的大规模数据集（引用），可能希望通过逐步降低数据的频率或者分辨率，来测试算法 所能的接受的数据精度和鲁棒性
	%上述内容都可以表达为key-value的形式，该形式对用户来说方便输入，对于程序来说方便解析。所以我们的配置文件时这种形式，用户只需要在网页中输入关于所有参数的参数名字和数值，系统会自动解析
	%另外，当我们需要运行大规模的tasks时，他们一般时有一定规律的（same algo same dataset different config），这时候，逐个创建config会非常的繁琐。所以我们提供了combine方式，批量生产（排列组合）（举个例子）；自动归类，方便后续查找
	
	When a mapping task is created, a unified configuration is automatically generated in the form of a yaml file, including algorithm selection, dataset selection, algorithm parameters, dataset parameters and ROS remap configuration. 
	
	The algorithm and dataset execution scripts that will extract the required information from the configuration for automated mapping runs are explained in the following sections.

	\subsubsection{Algorithm Execution Scripts}
	The Docker containers for the algorithm come each with algorithm execution scripts, consisting of a parameters template script and an execution script. Before starting the mapping run, the script gets the algorithm parameter values, dataset parameter values and remap information from the yaml configuration file, then generates  custom mapping configuration files according to parameters template and the requirements of the algorithm software, which will be read at the beginning of the mapping run. 

	\subsubsection{Dataset Component}
	The dataset component, specific for each supported dataset, consists of the dataset in the form of ROS bags, groundtruth, and dataset play script. In order to realize custom remap topic, the dataset play script can extract the dataset remap topic from configuration file and generate corresponding dataset play and remap commands.
	
	\subsubsection{Dataset Pre-processing Script}
	We want to benchmark SLAM algorithms with different image resolutions and/ or with different frame rates, in order to test their robustness against lower-quality data. In the future we will also add down-sampling or maximum range cut-off for point clouds and RGB-D data as options. For this SLAM Hive provides a dataset pre-processing script. This will take the original bagfile and, using the python rosbag api, store a modified dataset bagfile in the algorithms volume. SLAM Hive will then mount the modified folder instead of the original dataset folder as dataset volume to the algorithm container.

	It is worth noting that since cameras and their lenses have different models, so when re-scaling the image resolution  attention should be paid to modifying the corresponding intrinsic and extrinsic parameters, as well as the corresponding correction and undistortion parameters. This has to be done by the user by specifying the new camera configuration parameters for the scaled-down versions when creating new mapping tasks.

	\subsubsection{Standard Trajectory Format}
	Most SLAM algorithms nowadays use ROS. So all algorithms containers are built based on ROS in order to replay the dataset and record the topic of the estimated pose. For later evaluation, we convert estimated trajectories and groundtruth to the TUM format \cite{sturm2012benchmark}: $ tx\ ty\ tz\ qx\ qy\ qz\ qw $.

	\subsection{Multiple Configurations Mechanism}
	\label{sec:multipleconfigs}
	%Our system has a unified configuration format.
	Users configure mapping runs through the web interface. Configurations can be easily generated by selecting algorithms, datasets and inputting algorithm parameter values, dataset parameter values and remap topic from scratch or just copying an existing configuration and modifying some of its parameters. These configurations are recorded in the database for further performance analysis. The same algorithm and dataset with different algorithm parameters, or the same algorithm and parameters with different datasets may yield interesting results. Nondeterministic algorithms may be executed several times with the exact same parameters to evaluate their stability.
	
	\subsubsection{Combination Configuration Creation}
	Often users may wish to test a large number of permutations of configuration parameters. For example experiments with 5 different resolutions, 5 different frame rates on 5 different algorithms and datasets will lead to 625 different mapping configurations. SLAM Hive supports the easy generation of those mapping run configurations in batch by specifying parameter ranges in the web interface. 
	The operation flow is similar to creating a single configuration, except that multiple values are needed in the combined parameters separated by $|$: ``$ value1\ |\ value2\ |\ value3\ ...$''.

%	\subsubsection{Dataset Framerate and Resolution}
	%与普通的算法参数不同，当组合的参数涉及到数据集的帧率，图像的分辨率等时，不仅需要修改与之相关的参数的值（例如相机的内参），还需要对数据集进行相应的修改。
	%我们在组合创建configurations时，为涉及到数据集修改的参数提供了额外的输入，因为这种参数不能简单地对其进行组合（一组参数值对应某一个特定的频率或者分辨率）。
	%并且通过额外的输入，使得系统可以兼容不同的相机模型：用户仅仅需要提供新的内参数值即可。
	Different from ordinary algorithm parameters, when the combined parameters involve the dataset framerate, the resolution of images, etc, not only the values of related parameters (such as the intrinsics of the camera) need to be modified, but also the dataset needs to be modified accordingly.
	So when creating such configurations, we provide additional input for parameters that involve dataset modification, as such parameters which cannot be simply combined (a set of values corresponding to a specific framerate or resolution).
	Moreover, the system can be compatible with different camera models through additional input: the user only needs to provide a new parameter value.

		\begin{figure}[t]
		\centering
		\hspace{-1cm}
		\includegraphics[width=2.8in]{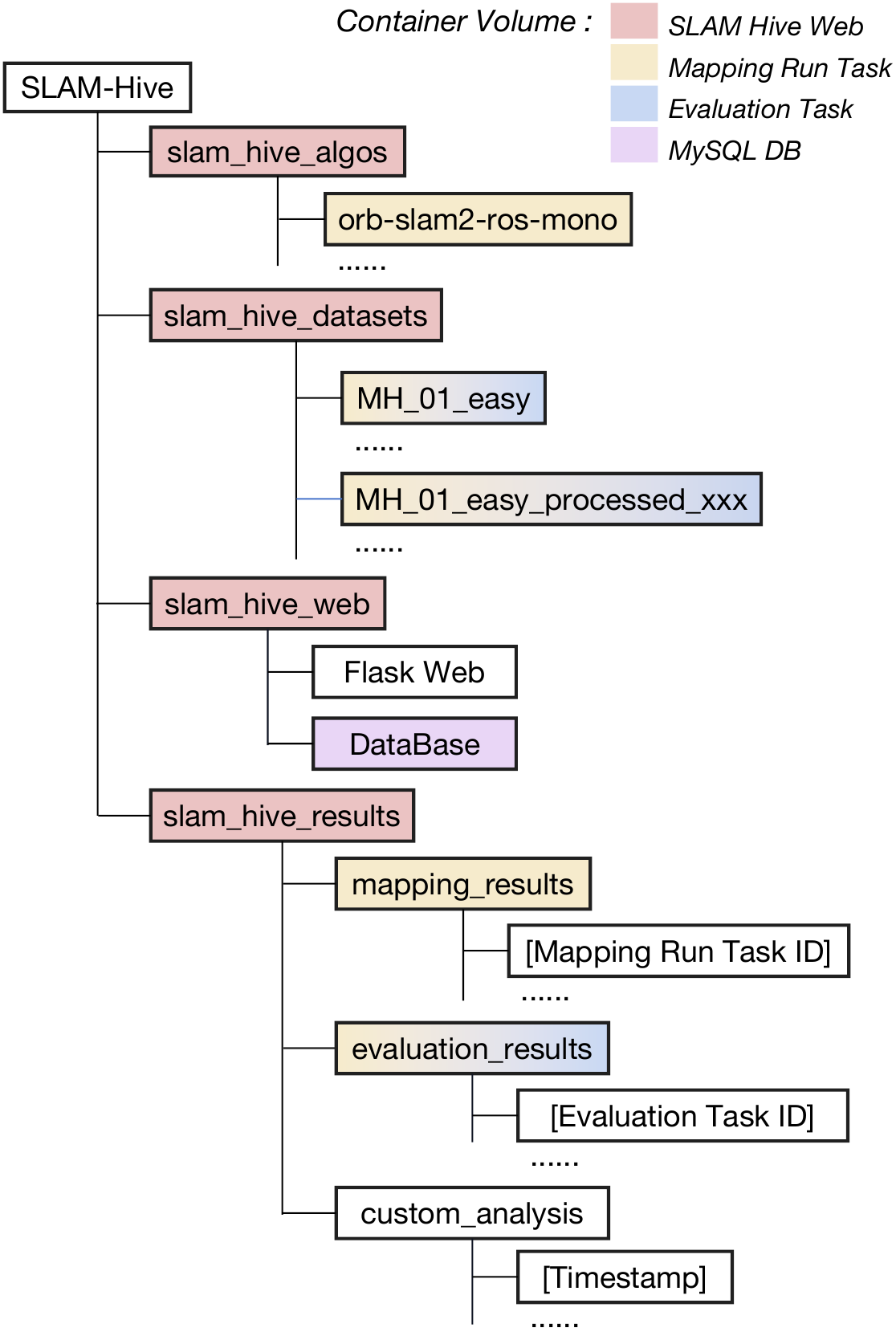}
		\caption{File storage structure in master node. Colors indicate which docker containers are mounting the folders as volumes. }
		\label{fig: file_structure}
		%	\vspace{-0.6cm}
	\end{figure}
	
	\subsection{Storage System}
	% 系统主要使用database和文件系统来存储数据。我们使用了关系型数据库-mysql来存储算法，数据集，配置文件，建图和评估任务等的参数数据，以及对结果的评估指标
	The SLAM Hive system uses a MySQL database, running in a separate Docker container, and the file system to store data. 
	MySQL stores information about algorithms, datasets, configuration files, mapping and evaluation tasks, as well as evaluation metrics for the results and is used heavily by the Flask web server to display and process this data.
	As detailed in Fig. \ref{fig: file_structure}, we use several folders that are mounted in the various Docker containers for algorithms, datasets and results. 
	The workflow of the file system around the mapping task is as follows: the system first creates the mapping run container based on the corresponding algorithm image, and then mounts the (potentially pre-processed) dataset as a read-only volume and the mapping results folder as read-write volume. 
	
	  Mapping tasks run consecutively on the workstation and store their results as files in the mapping\_results volume. These include the path estimated by the SLAM algorithm in TUM format as "traj.txt", CPU and memory profiling data as "profiling.csv" and two diagrams for CPU and memory as png images.  If algorithms support it, mapping runs can save the generated map as .pcd file for point cloud maps or .png for 2D grid maps. Those maps may use quite a lot of storage space, so this feature needs to be activated via setting the configuration parameter "save\_map" to true.
	  
	  SLAM Hive web is monitoring the result directory for a file called "finished", indicating that the run is finished. At that point the program copies CPU and memory information from the files to the database. Additionally, the length of the estimated trajectory, as a factor of the ground truth trajectory, is calculated and stored in the DB as well, having a value between 1. and 0. Mapping runs only returning a partial trajectory (e.g. due to tracking loss or frame loss due to limited computation resources) will get a traj\_length value $< 1$.

  \subsection{Mapping Run Evaluation}
	Mapping run evaluations are triggered manually, by clicking on the according button in the mapping run evaluation view, or by providing a list of mapping run ids or by clicking the button to evaluate all un-evaluated runs. 
	
	Currently, one evaluation container is utilized, that is using EVO \cite{grupp2017evo} to evaluate the estimated path against the ground truth path. In the future we plan to add more evaluation methods, e.g. w.r.t. the generated maps. The evaluation container mounts the dataset and slam\_hive\_results volumes and then runs evo, which creates 18 files of results, including diagram images and two .zip files containing more details about ape and rpe, respectively. 
	
	After  the evaluation container is finished, SLAM Hive will parse the result files and put the following values into the database for easy access: ATE and RPE: RMSE, mean, median, STD, min, max, SSE; as well as CPU max and mean and RAM max.
	
	\subsection{Search Engine}
	%数据库的一个重要的优势在于，相对于传统的文件存储方式，它使用结构化的形式来存储数据，有利于后续对数据进行各种统计学上的分析。
	%并且，考虑一个在实际应用中非常重要的场景：当研究者或者industry希望在某些特定的环境下，搭建一套满足需求的SLAM系统。由于各种类型的slam算法以及庞大的参数空间的存在，从零开始寻找一套适配的算法和配置将会是十分复杂和耗时的。
	%为了充分利用数据库的优势，以及解决上面提到的问题，我们基于我们的系统中的数据库，开发了一套搜索引擎，可以用于给出满足符合一定条件的configuration集合，从而实现后续的丰富的定制化分析。
	
	SLAM Hive may contain results of more than 100,000 mapping runs. It is thus essential to be able to search for mapping task configurations and mapping run evaluations, given various criteria. These search results will also facilitate the custom analysis described in the next section by filtering the results just for the mapping runs relevant for this analysis. 
	
	For the search, users choose the algorithms to include and datasets to include from two multi-selection lists. Furthermore, they can specify filters for specific configuration parameters as key value pairs, e.g. $nFeatures \ge 2000$ as "nFeatures $=>$ 2000". Additionally, when searching mapping run evaluations, users can specify minimum and maximum values of various evaluation results as filters, e.g. for ATE RMSE. \\\\
	
	% no need to many details to describe search engine
		
	% no need to describe the generator and resolver; no detail of YAML file
	% TODO remain this list or not
	\begin{lstlisting}[style=yaml, caption={Part of the Custom Analysis Input. Showing the basic components of the Input Text Content.}, label={lst:list2}]
		group_name: "name"
		group_description: "description"
		
		evaluation_form:
		algorithm_dataset_type: 0 
		1_trajectory_comparison:
		choose: 1
		...... other analysis form
		
		configuration_choose: 
		configuration_id: [1,2,3,4,5] # provide id
		comb_configuration_id: [5]
		limitation_rules:   # by a rule
		algorithm_id: [12]
		dataset_id: [15]
		parameters_value: ["nFeatures < 1200"]
		evaluation_value:
		ate_rmse_nolimitation: 1              # 0 or 1
		ate_rmse_minimum:                   
		# if just minimum and no maximum, can just fill the minimum
		ate_rmse_maximun:
		...... other evaluation metrics
		combination_rule:                    
		# how to combine two ways: Union set; intersection set; complement set; (1U2) - (3); U: Union; I: in
		first_one: [2]
		first_rule: ["I"]
		second_one: [0,1]
		second_rule: ["U"]
	\end{lstlisting}

	\subsection{Meta Analysis}
	% 参考vinseval对元分析的描述
	% yaml file format; resolver; generator;
	% 如何解析（规则
	% 生成（集中图标；后续便于添加）
	%可以加上几个副标题
	
	Meta analysis is one core function of our system, allowing users to compare and analyze sets of mapping runs w.r.t various parameters. This function is configured in YAML format. It selects the mapping runs to include, which can be further filtered down by specifying certain parameter values (e.g. "nFeatures $<= 1200$" or "traj\_length $> 0.75$"). 
	Furthermore the YAML file selects one or more types of the seven analysis modes we provide. See  Listing \ref{lst:list2} as an example. 
	The mapping runs are specified by configuration ids. If multiple mapping tasks are performed for the same configuration id (for non-deterministic algorithms) the analysis module for mode "3\_accuracy\_metrics\_comparison" will automatically calculate and use the mean and standard deviation of results, while only the first mapping run will be used in the other modes. 
	
	The trajectory\_comparison mode runs the EVO container to use that program to compare multiple trajectories, that must come from the same dataset. All other modes just run in the SLAM Hive instance. The other modes generate accuracy metric comparisons as diagrams or histograms, CPU/ RAM usage metrics comparisons, 2D and 3D interactive scatter diagrams and a mode for repeatability analysis for non-deterministic algorithms. Results are visualized as diagrams but their raw data is also available for download and further analysis. In experiments Section \ref{sec:experiment} we make extensive use of the Meta Analysis.

			\begin{figure*}[t]
		\centering
		\hspace{-1cm}
		\includegraphics[width=6.8in]{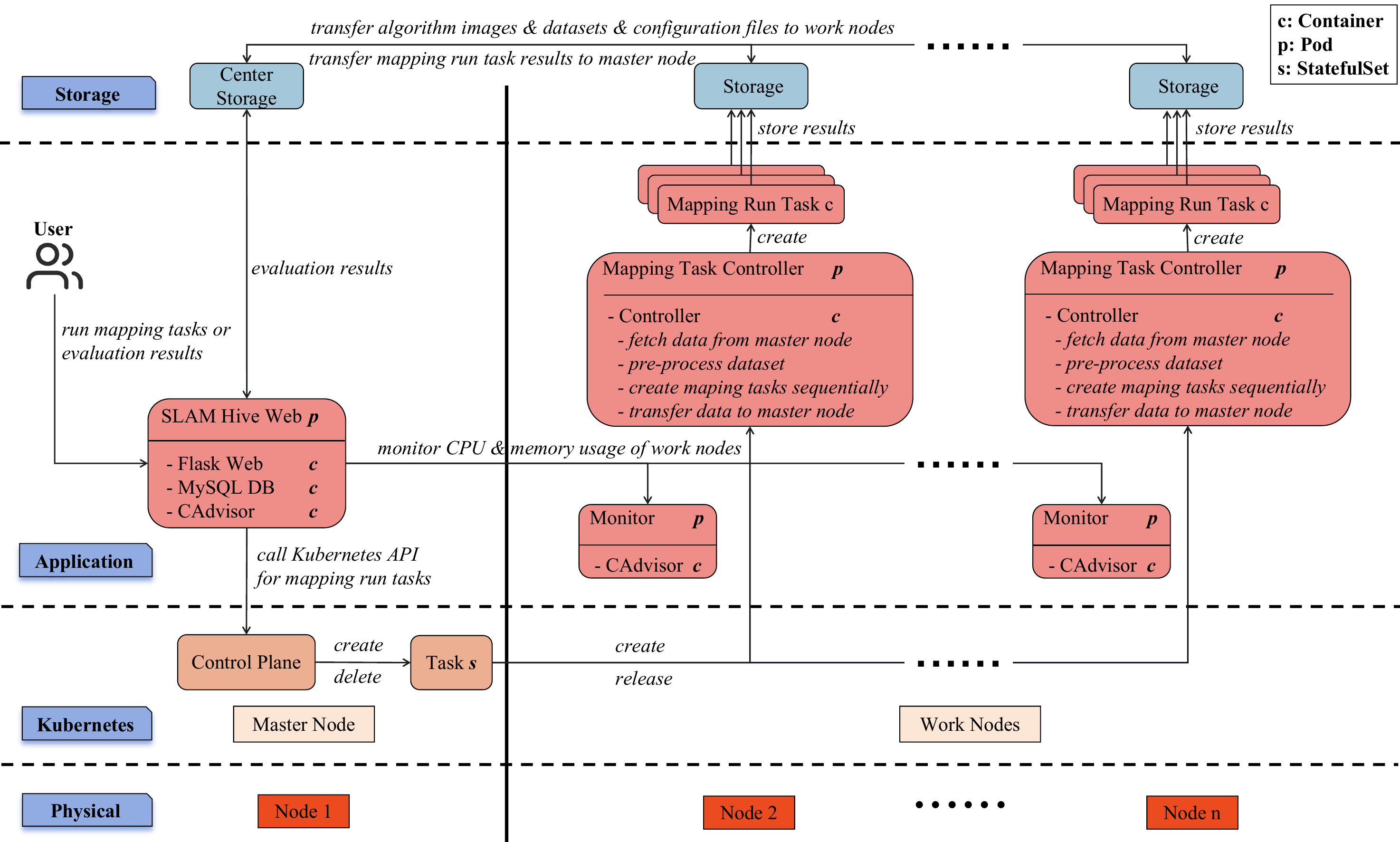}
		\caption{Cluster Workflow.} %有点不清晰
		\label{fig: clustercontroller}
	    \vspace{-0.5cm}
	\end{figure*}

	\section{Different Modes of SLAM Hive} %现在有4中模式 先读一下（尤其修改一下aliyun）
	SLAM Hive supports four different modes of deployment: The workstation mode is described above. Additionally, we have the view-only version web version for disseminating results and analysis, the cluster version for deployment in a local cluster and the could version with aliyun as example. 
	
	The mode, along with various other settings, needs to be configured in the configuration.py file of SLAM Hive before starting its container.

	\subsection{View-only Mode}
    For view-only mode Flask is checking the mode when generating web-pages and receiving posts to prevent any changes in the database and avoid starting mapping runs. This is demoed on \url{https://slam-hive.net/}. But the creation of new analysis is allowed even in view-only mode. Those will not show up in the list of available analysis when created in view-only mode, but still be available at the generated URL. A further configuration value "NO\_NEW\_ANALYSIS" can be set to true to also prevent the creation of new analysis runs, which is used for the archive version of SLAM Hive. 

    \subsection{Cluster Mode}
	Using the Multiple Configurations Mechanism (Section \ref{sec:multipleconfigs}) users can create 10's of thousands of mapping tasks. Running those on the workstation would take too much time. Therefore, we want to distribute these tasks to a cluster of computers or even the cloud (see next section).

	Cluster deployment works similarly on local and cloud clusters:  you only need to include information about the nodes available (such as the node host name and inner IP) in the configuration.py file.

\begin{figure*}[b]
	\vspace{-0.7cm}
	\centering
	\captionsetup[subfloat]{font=scriptsize}
	
	\subfloat[CPU usage over time]{
		\label{fig:subfig:onefunction}
		\includegraphics[width=0.45\textwidth]{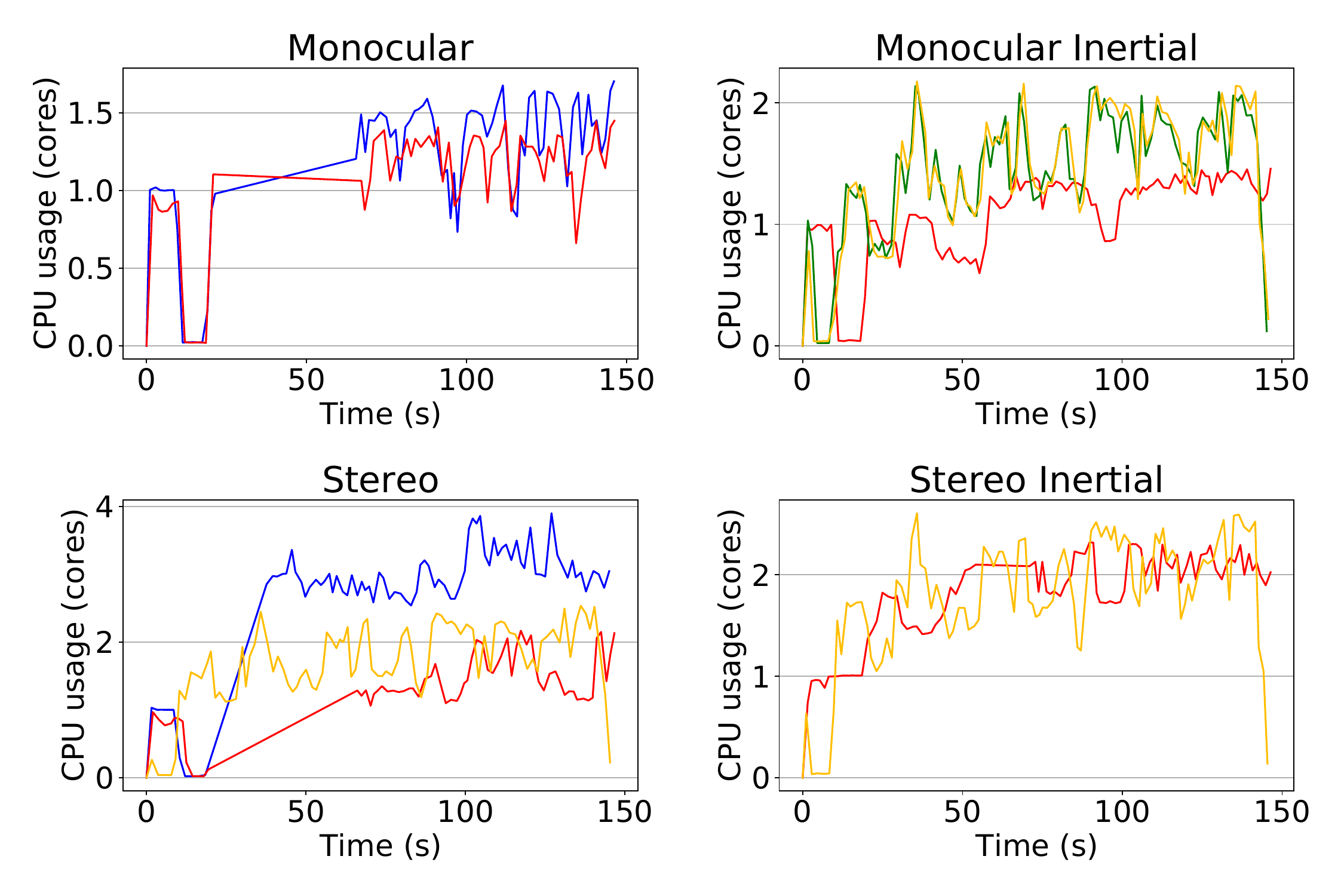}}
	\hspace{0.05\textwidth}
	\subfloat[RAM usage over time]{
		\label{fig:subfig:twofunction}
		\includegraphics[width=0.45\textwidth]{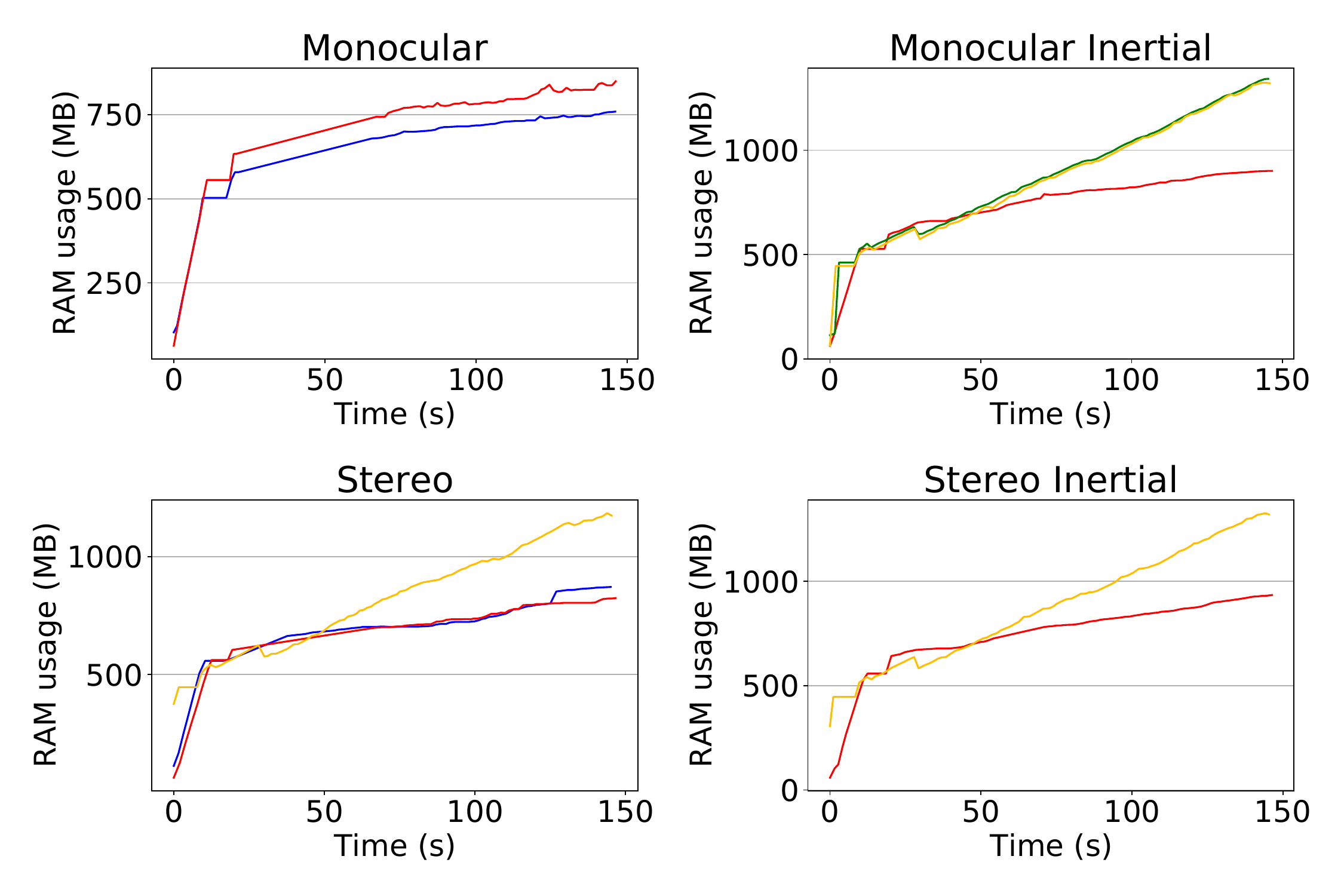}}
	
	\caption{(a) and (b) are the CPU and memory usage change over time. Both of them are different methods under the same sensors combination on one of sequence on EuRoC dataset: \textit{MH\_03\_medium}. Blue is ORB-SLAM2, red is ORB-SLAM3, green is VINS-Mono, yellow is VINS-Fusion.}
	\label{fig:usage}
\end{figure*}

	\subsubsection{Mapping Task Controller}
	\label{subsection: cluster containers controller}
	Our system uses the open source container orchestration engine Kubernetes to create, deploy and manage a big number of containers on a multi-node computational cluster environment. 
	We implement a Mapping Task Controller based on Kubernetes, which is responsible for the entire life cycle of the task containers. Its overview is given in Fig. \ref{fig: clustercontroller}. 

	% TODO figure out the correct workflow (no need to change the text)
	When users create $n$ tasks on a cluster with $m$ available work nodes, the SLAM Hive will parse the request and use Kubernetes to start $min(m, n)$ Mapping Task Controllers, maximum one on each work node. Mapping Task Controllers are responsible for managing the algorithm containers and the transfer of the data the algorithm containers need and create (i.e. the mapping results). 	The static resources (algorithm images (obtained via docker save as .tar files on the SLAM Hive node), datasets and configuration files) need to be fetched onto the node by the controller. 
	
	 To minimize the network transfer of the big algorithm images and dataset files, SLAM Hive first classifies tasks according to their algorithm and dataset, called ``class''. So each pair of algorithm and dataset will be assigned to one ``class'', where the ``class'' then is the set of different configuration permutations for that algorithm and/ or dataset. We then assign every ``class'' to one specific mapping task controller randomly. One configuration file "subTask.txt" with the assignments of mapping task ids per node/ mapping task controller is then generated and given to the mapping task controllers upon startup. 
	
	For each mapping task to be run, the controller first checks if the algorithm image and dataset folder are present in the "Node Storage" volume that is mounted from the nodes local file system. If not, it downloads these via scp from Node 1, which is running SLAM Hive. Because of the classification done above, for subsequent mapping runs on the same node it is highly likely that no additional transfer is needed, because that mapping task controller will mainly be responsible for mapping tasks with the same algorithm and dataset but varying other configuration settings. 
	
	After that, the controller runs the potentially needed dataset pre-processing and then starts the algorithm container and executes the algorithm script, which in turn runs the dataset playback script, to run the mapping task. Finally, the mapping run results, including trajectory file and performance monitor data, are transferred to the master node's slam\_hive\_results folder via scp. Then the controller may start the process again for the next mapping task.
	
		Actually, minimizing data transfers while keeping the workload of all nodes balanced is a complicated problem, since not only the sizes of individual algorithm images and datasets are important, but also the playtime of the datasets as well as time and storage needed for dataset pre-processing are significant factors. So our solution is not optimal and could be improved upon in the future.

	\subsection{Cloud Mode}
	Cloud Mode enable the deployment of SLAM Hive in the cloud, e.g. on Aliyun, Alibaba's cloud service. This enables users to, on the fly, rent many compute nodes and thus run a multitude of mapping runs in parallel. When doing so, care has to be taken to run all tasks on nodes with the same type of CPU, such that the core-utilization count is comparable. In fact, SLAM Hive saves CPU type and number of available cores in the DB for each run. 
	
	There are three main differences to the cluster mode: 1) the cloud providers API is used to purchase and manage compute nodes on-line. 2) transfer costs of algorithm images and datasets can be reduced by creating snapshots of nodes and initializing new nodes based on these snapshots. 3) The dynamic addition of new nodes via the cloud provider's API necessitates the dynamic configuration of Kubernetes. Problems 1) and 3) are solved in SLAM Hive by using the according API's, but problem 2) is more complicated and discussed in detail below.

	\begin{table*}[h]
		\scriptsize
		\tiny
		\centering
		\tabcolsep=0.14cm
		\renewcommand\arraystretch{1.3}
		\caption{RMSE ATE, CPU and Memory usage of different algorithms under the same sensors combination on EuRoC dataset}
		\label{table: comparison}
		\begin{threeparttable}
			\begin{tabular}{c|l|cccc|cccc|cccc|cccc|cccc}
				\hline
				\multirow{2}*{\makecell[c]{Sensor\\Combinations}} & \multirow{2}*{Algorithms}
				&\multicolumn{4}{c|}{MH\_01\_easy\tnote{1}} & \multicolumn{4}{c|}{MH\_02\_easy\tnote{1}} & \multicolumn{4}{c|}{MH\_03\_medium\tnote{1}} & \multicolumn{4}{c|}{MH\_04\_difficult\tnote{1}} & \multicolumn{4}{c}{MH\_05\_difficult\tnote{1}}\\
				\cline{3-22} && MEAN\tnote{2} & STD\tnote{2} & CPU\tnote{3} & RAM\tnote{4} & MEAN & STD & CPU & RAM & MEAN & STD & CPU & RAM & MEAN & STD & CPU & RAM & MEAN & STD & CPU & RAM \\
				
				\hline \hline 
				\multirow{3}*{Monocular}
				&ORB-SLAM2& {0.071} & 0.0647 & 1.19/1.84 & {1884} & 0.036 & \textbf{0.0005} & 1.22/1.83 & \textbf{858} & 0.038 & \textbf{0.0019} & 1.15/1.71 & \textbf{805} & 0.066 & \textbf{0.0038} & \textbf{1.11/1.69} & \textbf{847} & 0.052 & 0.0042 & 1.18/2.36 & \textbf{872}\\
				%			&DSO      &   &  &  &  & &  &  &  &  & & & &&&\\
				&ORB-SLAM3& \textbf{0.046} & \textbf{0.0044} & \textbf{1.12/1.58} & \textbf{947} & \textbf{0.034} & 0.0017 & \textbf{1.07/1.54} & 1470 & \textbf{0.037} & 0.0020 & \textbf{1.02/1.46} & 881 & \textbf{0.061} & 0.0192 & 1.13/1.76 & 873 & \textbf{0.049} & \textbf{0.0016} & \textbf{1.06}/\textbf{1.89} & 1206\\
				
				\hline \hline 
				\multirow{3}*{\makecell[c]{Monocular\\ Inertial}}
				&VINS-Mono& 0.113 & 0.0113& 1.45/2.36 & 1278 & 0.157 & 0.0030 & 1.41/2.18 & 1443 & 0.110 & 0.0065 & 1.42/2.17 & 1350 & 0.114 & \textbf{0.0011} & 1.24/2.05 & 1085 & 0.180 & \textbf{0.0009} & 1.24/2.05 & 1085\\
				&VINS-Fusion& 0.065 & \textbf{0.0026} & 1.44/2.29 & 1610 & \textbf{0.038} & \textbf{0.0008} & 1.42/2.15 & 1453 & 0.065 & \textbf{0.0026} & 1.42/2.24 & 1788 & 0.099 & 0.0090 & \textbf{1.20}/2.07 & 1082 & 0.106 & 0.0016 & 1.19/2.06 & 1125\\
				&ORB-SLAM3& \textbf{0.047} & 0.0044 & \textbf{1.18}/\textbf{1.68} & \textbf{988} & 0.054 & 0.0074 & \textbf{1.13}/\textbf{1.64} & \textbf{1028} & \textbf{0.054} & 0.0046 & \textbf{1.06}/\textbf{1.48} & \textbf{1014} & \textbf{0.052} & 0.0039 & 1.22/\textbf{1.85} & \textbf{865} & \textbf{0.063} & 0.0114 & \textbf{1.01}/\textbf{1.45} & \textbf{966}\\
				
				\hline \hline 
				\multirow{3}*{Stereo}
				&ORB-SLAM2& \textbf{0.035} & 0.0026 & 2.56/3.76 & 1059 & 0.038 & 0.0036 & 2.36/3.67 & {888} & 0.105 & 0.0821 & 2.54/4.09 & 996 & {0.109} & 0.0377 & 2.71/4.12 & 1000 & {0.082} & 0.0509 & 2.85/4.50 & 1162\\
				&VINS-Fusion & {0.052} & \textbf{0.0009} & 1.64/2.42 & 1301 & 0.054 & 0.0287 & 1.64/2.49 & 1236 & \textbf{0.047} & \textbf{0.0006} &1.68/2.55 & 1205 & 0.112 & \textbf{0.0010} & 1.48/2.39 & 1070 & 0.083 & \textbf{0.0010} & 1.46/2.39 & 1067 \\
				&ORB-SLAM3& 0.042 & 0.0022 & \textbf{1.51}/\textbf{2.23} & \textbf{1020} & \textbf{0.034} & \textbf{0.0014} & \textbf{1.09}/\textbf{1.85} & \textbf{884} & {0.064} & 0.0416 & \textbf{1.24}/\textbf{2.22} & \textbf{874} & \textbf{0.099} & 0.0348 & \textbf{1.37}/\textbf{2.32} & \textbf{923} & \textbf{0.058} & 0.0063 & \textbf{1.32}/\textbf{2.37} & \textbf{882} \\
				
				\hline \hline 
				\multirow{2}*{\makecell[c]{Stereo\\ Inertial}}
				&VINS-Fusion& 0.042 & 0.0020 & \textbf{1.75}/2.64 & 1568 & \textbf{0.020} & \textbf{0.0012} & 1.76/2.61 & 1625 & \textbf{0.044} & 0.0040 & 1.79/2.61 & 1349 & 0.111 & 0.0088 & \textbf{1.56}/2.54 & 1090 & 0.084 & \textbf{0.0021} & \textbf{1.55}/2.54 & 1090\\
				&ORB-SLAM3& \textbf{0.040} & \textbf{0.0014} & 1.82/\textbf{2.37} & \textbf{1501} & 0.049 & {0.0047} & \textbf{1.67}/\textbf{2.30} & \textbf{917} & {0.045} & \textbf{0.0012} & \textbf{1.76}/\textbf{2.36} & \textbf{980} & \textbf{0.048} & \textbf{0.0017} & 1.96/\textbf{2.50} & \textbf{932} & \textbf{0.060} & \textbf{0.0021} & 1.79/\textbf{2.40} & \textbf{894} \\
				\hline 
				%		\hline \hline 
				%		\multirow{1}*{Lidar Inertial}
				%		&LIO-SAM&  &  &  &  & &  &  &  &  & & & \\
				%		\hline
			\end{tabular}

			\begin{tablenotes}
				%			\footnotesize
				\scriptsize
				%			\tiny
				\item[1] They are five sequences of varying difficulty on the EuRoC dataset. \textit{MH\_01\_easy} and \textit{MH\_02\_easy} are with good texture and bright scene, \textit{MH\_03\_medium} is with fast motion and bright scene, \textit{MH\_04\_difficult} and \textit{MH\_05\_difficult} are with fast motion and dark scene.
				\item[2] These represent the mean and standard deviation of RMSE respectively. RMSE is the root mean square error of the ATE, the unit is meter.
				\item[3] The calculation of CPU usage is the sum of all core usage. CPU in the above table is average/maximum CPU usage during mapping, its unit is core. 
				\item[4] RAM is the maximum memory usage during mapping, its unit is MB.
				%			\item[5] Avg is the average RMSE, CPU and RAM of 5 sequences.
				
			\end{tablenotes}
		\end{threeparttable}
		\vspace{-0.4cm}
	\end{table*}

	\subsubsection{Data Sharing via Snapshots}
	% 正如之前（todo第几章第几节）提到的，如何平衡节点之间的传输速率和每个节点所运行的容器数量是一个复杂的问题。
	% TODO (in Chapter 4 of todo)
	As mentioned before, balancing the number of containers running on each node is a complex issue which is further complicated in the cloud by adding the costs incurred as another optimization objective. 
	% 简单来说，这取决于主节点传输任务所需要的算法镜像和数据集资源所耗费的时间与每个任务运行所需要的时间之间的长短关系，类似于计算机系统中的任务并行调度以及流水化执行。而且这只是理想化的情况，事实上，如果仅由主节点负责数据传输，将会给其带来巨量的负载，影响主节点上运行的其他服务；而如果将传输任务均摊给其他工作节点，使整个传输过程并行化，这看起来可以大幅度缩减传输时间，但是受限于整个局域网的带宽，实际上优化效果不是很明显，并且这种复杂的机制将会变得难以管理，增加任务出错的风险。
	It depends on the taken time between the master node to transfer the algorithm image and dataset resources and each task to run, similar to the parallel scheduling and pipelining of tasks in computer systems. In fact, if the master is solely responsible for data delivery, it will put a huge amount of load on it, especially with 1000s of work nodes, affecting other services running on the master. If the transmission task is distributed to other work nodes and the whole transmission process is parallelized, the transmission time can be greatly reduced. But cloud providers typically cap the bandwidth available to a virtual LAN to a low value (e.g. 5Gbps for Aliyun), such that even the distributed sharing of images and datasets also does not scale well.

	Our solution to this problem is to employ the snapshot feature cloud services provide. A snapshot is the template image from which the node is booting, containing the operating system but also the other data. Users can create snapshots containing their application and data, also on the fly as a snapshot of an existing node. Utilizing this to "copy" algorithm images and datasets to new nodes is much cheaper than the cost of the time spend for transferring those via the network, and also faster by an order of magnitude, according to our tests.

	%据此，我们设计了一套“1 + 1 + （n-1）”机制：第一个“1”代表一台主服务器节点；第二个“1”代表用于生成模版镜像的工作节点；“n”代表本次任务所需要的工作节点的总数，因此“n-1”代表除去模版节点外的其他工作节点。
	Accordingly, we design a ``$1 + 1 + (n-1)$'' mechanism: the first ``1'' represents the master node; The second ``1'' represents the work node used to generate the snapshot; ``n'' is the total number of work nodes required for this task, so ``n-1'' is the number of worker nodes excluding the template work node.
	%在任务执行前，运行在主服务器上的控制器将会受限创建一个工作节点，并将本次任务所需要的静态资源全部传输到工作节点中；然后，控制器将会以该节点为模版创建镜像；最后，其他“n-1”个节点同时以此镜像为基础创建。
	Before the task is executed, the SLAM Hive running on the master will create a work node and transfer all the static resources needed for this task to the work node. Then it will create a snapshot of that node. Finally, ``n-1'' other nodes are simultaneously created from this image.
	
	% no need to show the detail of math formulation of aliyun transfer data

	\section{Benchmarking Experiments}
	\label{sec:experiment}

	To highlight the utility of SLAM Hive and to already explore the performance of SLAM algorithms under various parameters, we first conduct three small experiments for visual SLAM (vSLAM) algorithms, LiDAR-based algorithms and show how SLAM Hive can be used to explore the parameter space, on the example of RGBD ORB-SLAM2. The main experiment then is analyzing four vSLAM algorithms in great details, from a total of 1,500 mapping runs. The results and analysis of all experiments of this paper are archived and can be viewed and downloaded on \url{https://archive.slam-hive.net}, while \url{https://slam-hive.net} may be updated with newer results in the future. The archived SLAM Hive with all the results can be downloaded for further analysis from github.

	All tests are done in workstation mode on the same computer, with the following configuration: 32 GB memory, Intel Core i7-7700 CPU @ 3.4=60GHz $\times$ 8.

	\begin{figure}[t]
		\centering
		\includegraphics[width=3.in]{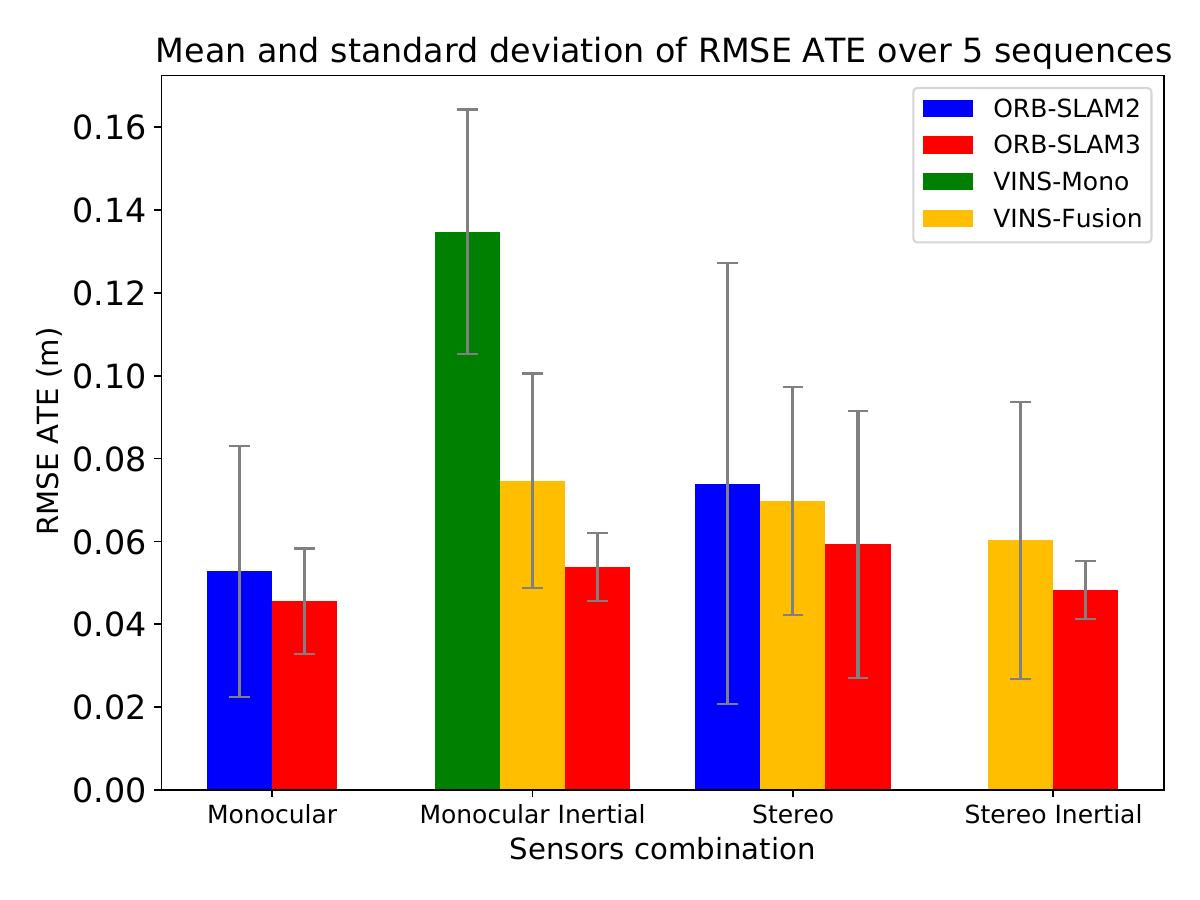}
		\caption{Mean and standard deviation of RMSE ATE over 5 sequences on EuRoC dataset.}
		\label{fig: ate}
		\vspace{-0.4cm}
	\end{figure}

	\subsection{Performance Comparison of SLAM Methods}
	
	We first explore four vSLAM methods with 250 mapping runs and five LiDAR algorithms with six mapping runs. We analyze those runs w.r.t. CPU and memory consumption as well as the RMSE (root mean squared error) ATE (absolute trajectory error). Finally, we explore the ORB-SLAM2 RGBD algorithm with the varying parameter "number of ORB features" with 10 different values.

	%There are currently some SLAM solutions for different sensor fusions, such as monolular, monolular intertial, stereo, stereo intertial, lidar inertial etc.
	
	%总体说明一下：我们在不同的数据集上，对不同不同类型的SLAM算法进行了比较。
	%（加上bowen的数据集，说明如果有了一个全面的数据集，将可以。。。）
	% Due to the lack of public datasets rich in various sensors, it is somewhat difficult to compare visual SLAM methods with LiDAR SLAM methods. In the future we will use the ShanghaiTech Mapping Robot II to create such datasets \cite{yang2022cluster}.
	
	\subsubsection{Performance Comparison of Multiple Visual SLAM Methods}
	\label{subsec:vSLAM_experiments}
	
	%ToDo: add citations for the algorithms
	We compare ORB-SLAM2 \cite{murORB2} and 3 \cite{9440682} with VINS-Mono \cite{qin2018vins-mono} and VINS-Fusion \cite{qin2019vins-fusion} with monocular and stereo as well as with and without IMU data on five, increasingly difficult sequences from the EuRoC dataset \cite{burri2016euroc}. We repeat each mapping run five times to also explore the influence of the randomized algorithm components. Fig. \ref{fig: ate} shows the RMSE ATE of the experiments in a more condensed version that the SLAM Hive result 
		\footnote{\url{https://archive.slam-hive.net/analysis/show/1721823767561678}}, while Table \ref{table: comparison} shows the details of the 10 different algorithm-sensor setup combinations, with additional data downloaded from 
		 \footnote{\url{https://archive.slam-hive.net/analysis/show/1721799577566257}}
		\footnote{\url{https://archive.slam-hive.net/analysis/show/1721799873765728}}
		\footnote{\url{https://archive.slam-hive.net/analysis/show/1721799706550709}}
		Those 10 combinations over 5 sequences repeated each 5 times result in 250 executed mapping runs.

%	 shows the comparison of different methods under the same sensor combination on 5 different sequences from the EuRoC dataset. They are increasingly difficult to process in terms of flight dynamics and lighting conditions. For each algorithm, we used the default parameters. The results of each method in each sensor mode are summarized in Table \ref{table: comparison}, with additional data downloaded from 

	As shown in Fig. \ref{fig:usage}, in the 4 modes, the CPU usage of ORB-SLAM3 is smaller, while the memory usage is larger. In both monocular and stereo modes, the memory usage of ORB-SLAM3 is almost twice that of ORB-SLAM2. Fig. \ref{fig: ate} indicates that ORB-SLAM3 performs more stable and has the highest accuracy. The CPU usage of ORB-SLAM2 stereo mode is almost twice that of monocular mode. There is no obvious difference in the memory usage of ORB-SLAM3 in the 4 modes, and there is no significant difference in CPU consumption with or without the fusion of the IMU when the number of cameras is the same.
	
   Because we ran each algorithm 5 times we can also analyze the variance of each algorithm. The error bars in Fig. \ref{fig: ate} show this variance mixed with the variance from the 5 different sequences. We can see that ORB-SLAM2 in stereo mode has a very high standard divination, while the the variance of ORB-SLAM3 is quite small. Overall, ORB-SLAM3 has the best accuracy.

	\subsubsection{Performance Comparison of Multiple  LiDAR SLAM Methods}
	
\begin{figure*}[t]
		\centering
		
		\captionsetup[subfloat]{font=scriptsize}
		\subfloat[CPU usage over time]{
			\label{fig:subfig:onefunction} 
			\includegraphics[width=2.6in]{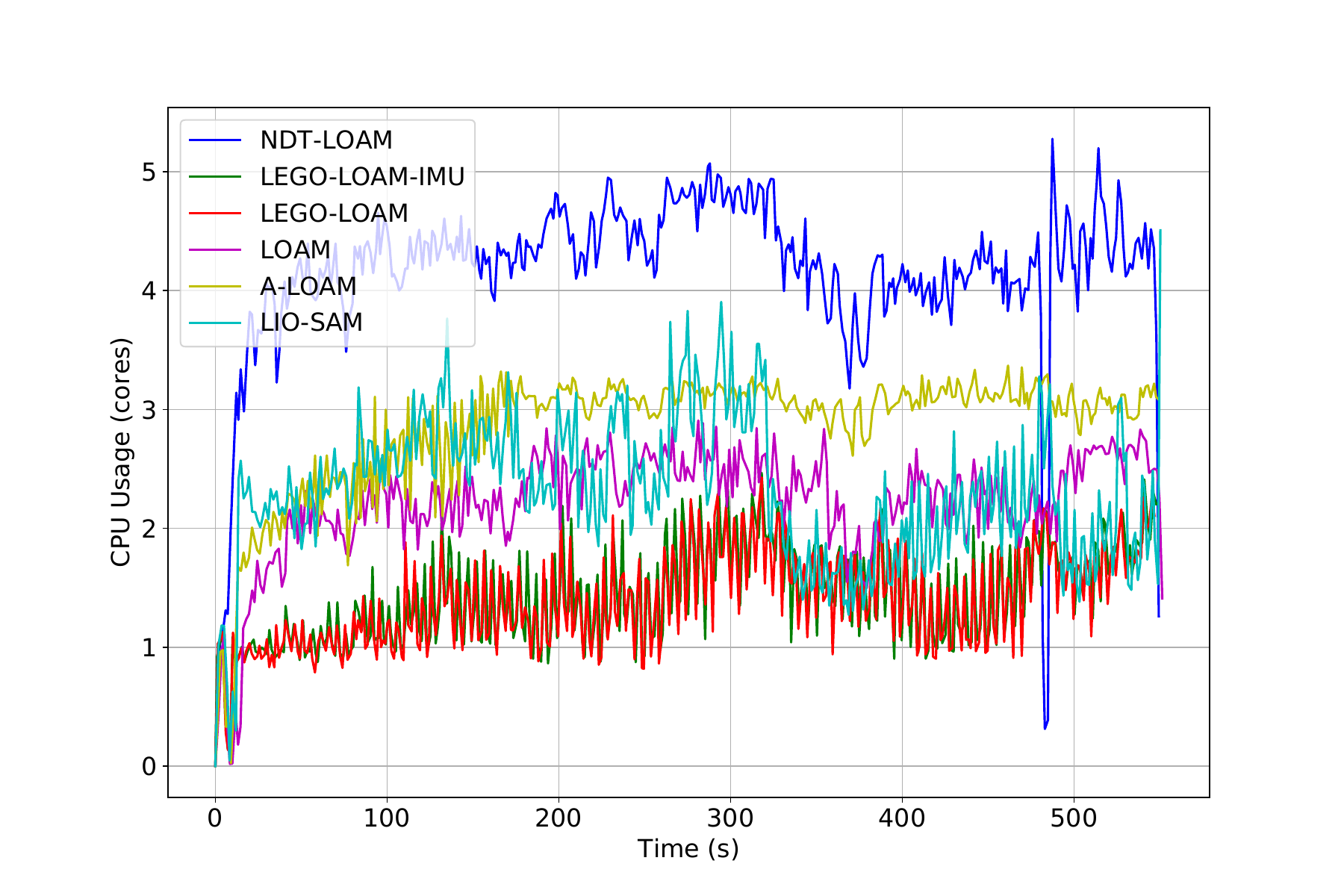}}
		\hspace{0.5in} 
		\subfloat[RAM usage over time]{
			\label{fig:subfig:twofunction} 
			\includegraphics[width=2.7in]{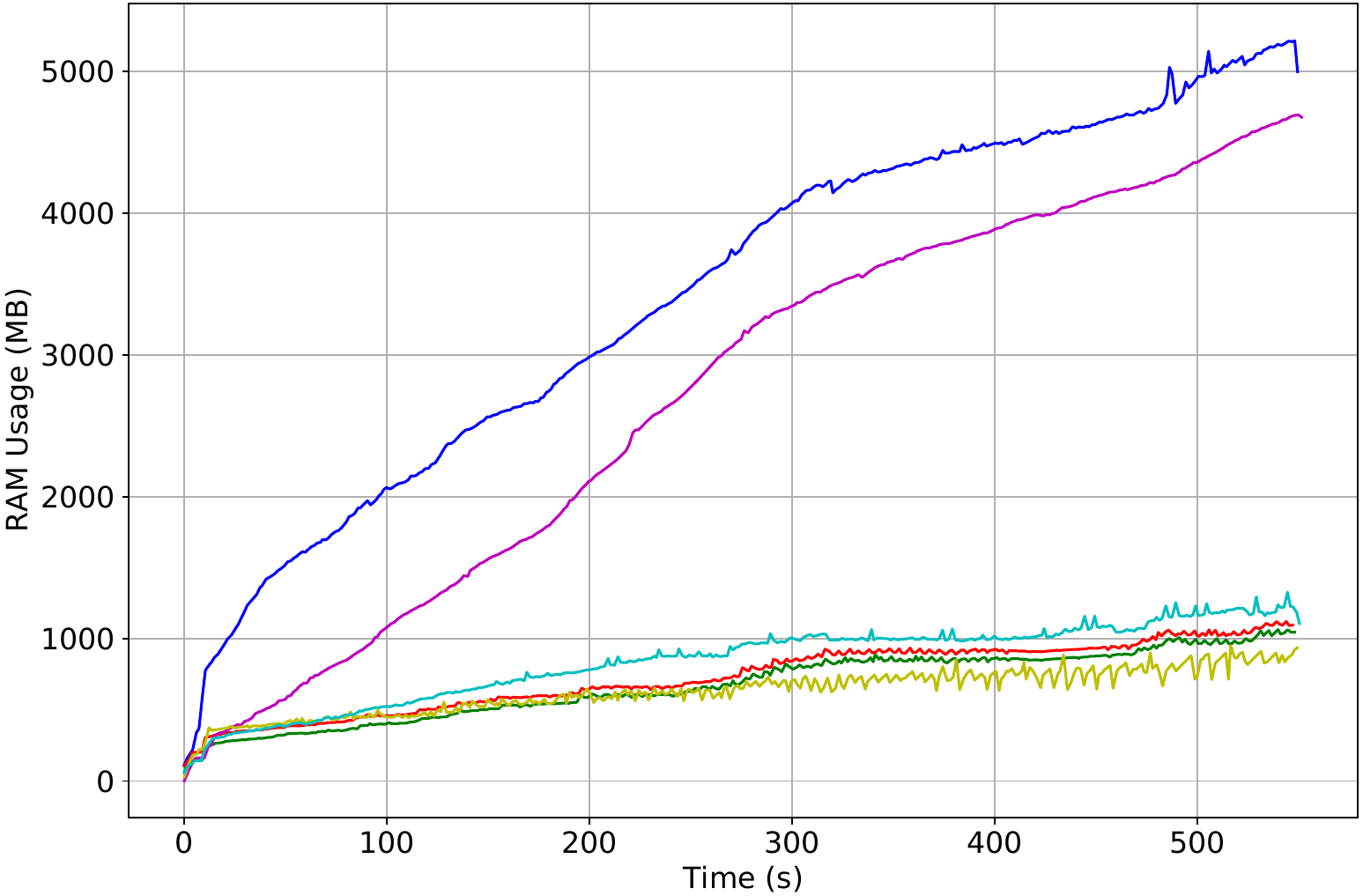}}
		\caption{(a) and (b) are the CPU and memory usage change over time. Both of them are Lidar-based methods on one of sequence on KITTI dataset: \textit{kitti\_2011\_09\_30\_drive\_0028}. }
		\label{fig: lidar_usage} 
		\vspace{-0.4cm}
	\end{figure*}

\begin{figure*}[t]
		\centering
		
		\captionsetup[subfloat]{font=scriptsize}
		\subfloat[Trajectory comparison]{
			\label{fig:subfig:onefunction} 
			\includegraphics[width=2.5in]{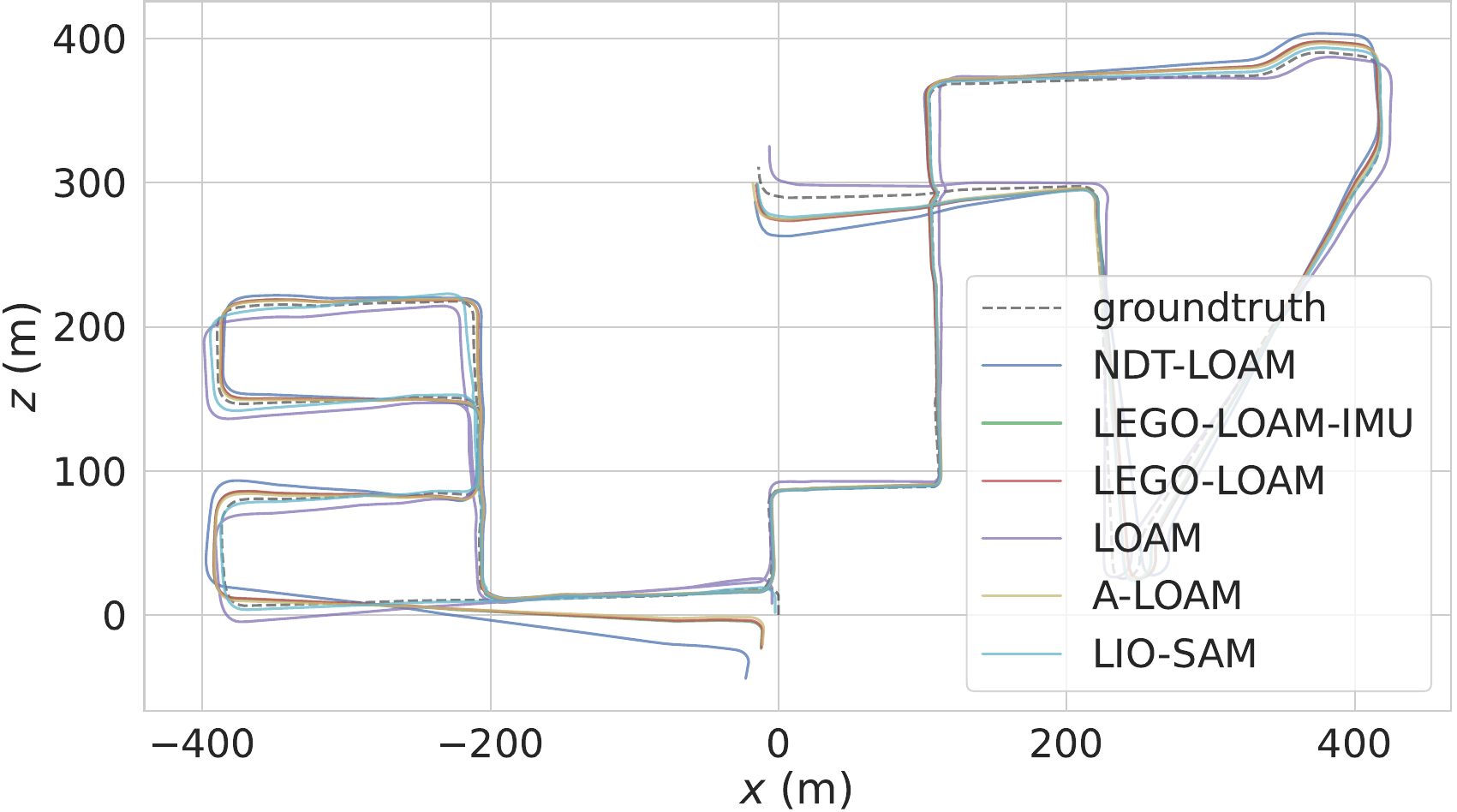}}
		\hspace{0.5in} 
		\subfloat[XYZ value comparison]{
			\label{fig:subfig:twofunction} 
			\includegraphics[width=2.8in]{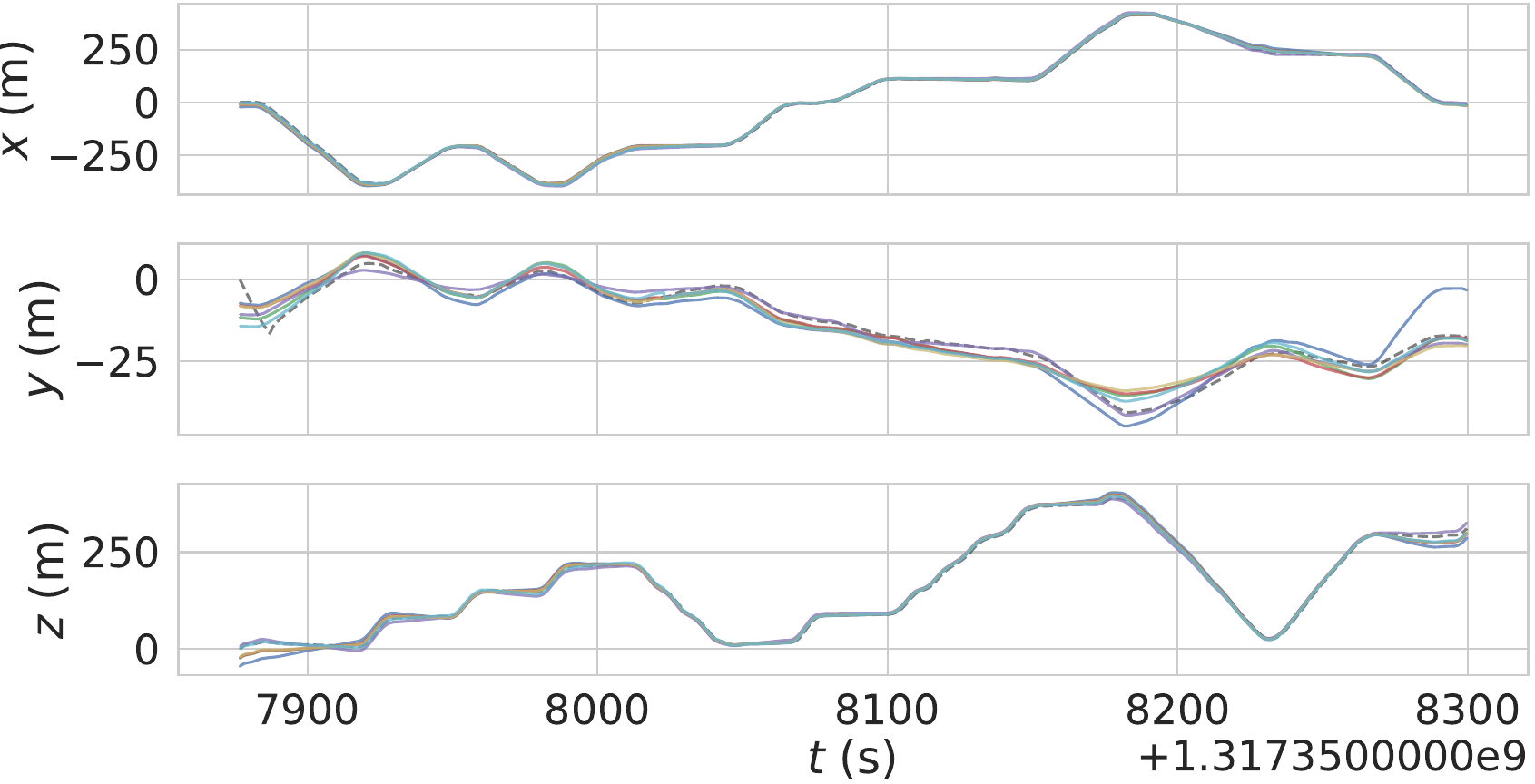}}
		\caption{(a) and (b) are the comparison graphs between the estimated and groundtruth of the trajectories on one of sequence on KITTI dataset: \textit{kitti\_2011\_09\_30\_drive\_0028} output by different Lidar-based methods. }
		\label{fig: lidar_traj} 
		\vspace{-0.4cm}
	\end{figure*}

\begin{figure*}[t]
		\centering
		
		\captionsetup[subfloat]{font=scriptsize}
		\subfloat[ATE basic comparison]{
			\label{fig:subfig:onefunction} 
			\includegraphics[width=2.in]{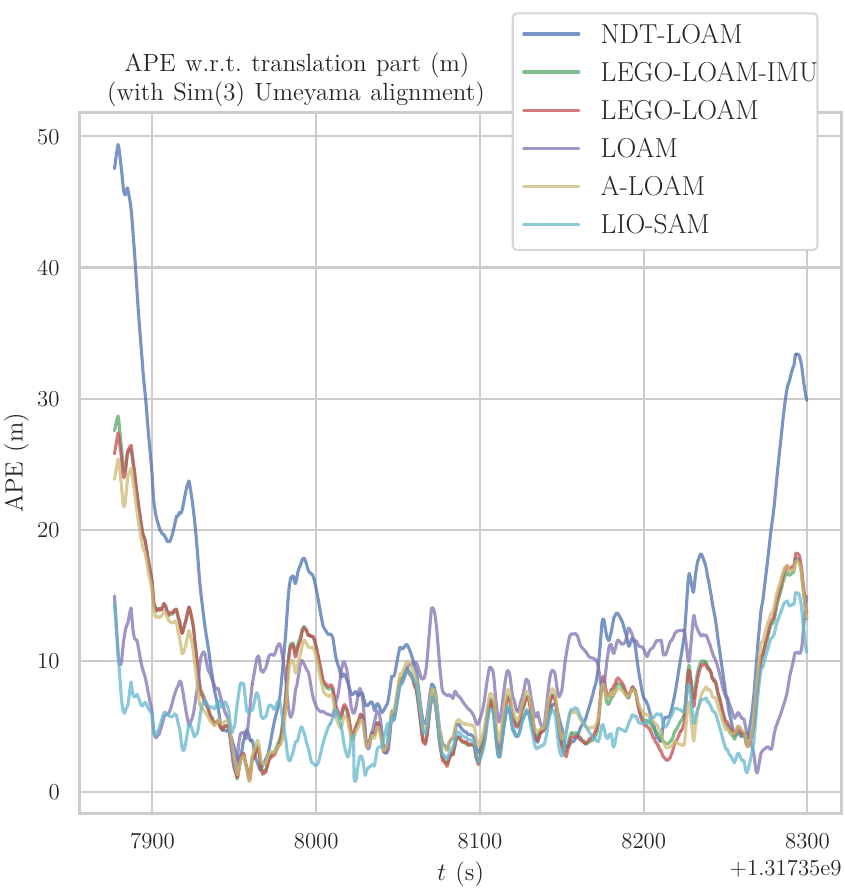}}
		\hspace{0.8in} 
		%\subfloat[RPE basic comparison]{
			%	\label{fig:subfig:twofunction} 
			%	\includegraphics[width=2.2in]{imgs/lidar_base_final/new_rpe_raw.pdf}}
		\captionsetup[subfloat]{font=scriptsize}
		%\subfloat[ATE statistic value comparison]{
			%	\label{fig:subfig:onefunction} 
			%	\includegraphics[width=2.5in]{imgs/lidar_base_final/new_ape_stat.pdf}}
		%\hspace{0.5in} 
		\subfloat[RPE statistic value comparison of the trajectory over time]{
			\label{fig:subfig:twofunction} 
			\includegraphics[width=2.in]{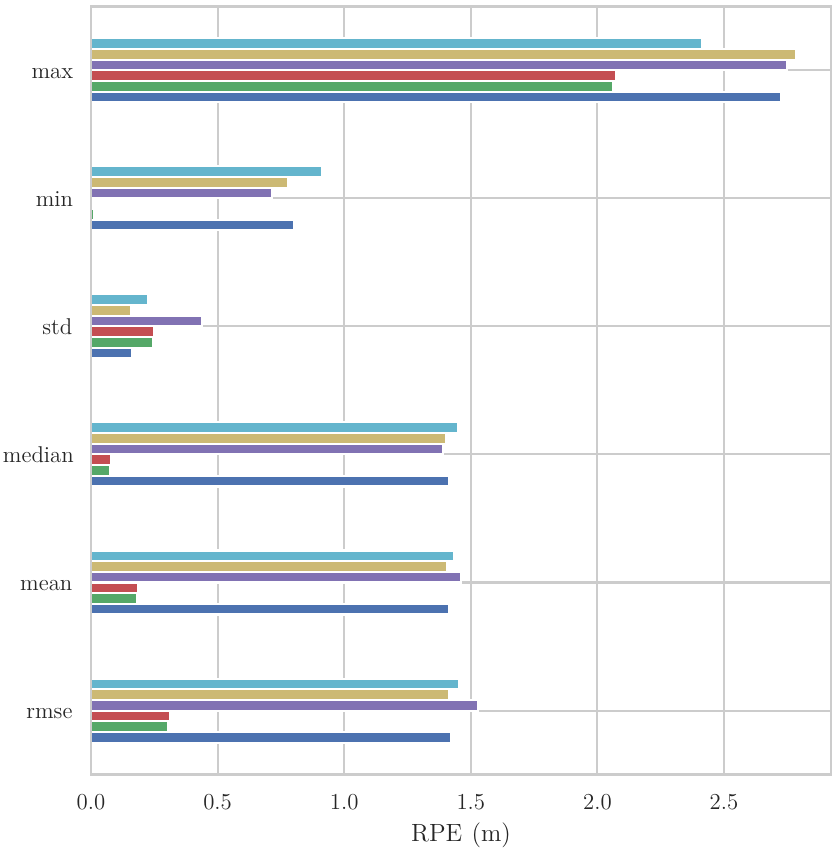}}
		\caption{Detailed comparison of APE and RPE of different Lidar-based methods. }
		\label{fig: lidar_ate_rpe} 
		
		\vspace{-0.4cm}

	\end{figure*}

	We compare LOAM \cite{zhang2014loam}, A-LOAM, LEGO-LOAM \cite{legoloamIROS2018}, LEGO-LOAM-IMU, NDT-LOAM  \cite{ndtloamSJ2022} and LIO-SAM \cite{liosam2020shan}  on the \textit{kitti\_2011\_09\_30\_drive\_0028} \cite{Geiger2013IJRR}  dataset 	\footnote{\url{https://archive.slam-hive.net/analysis/show/1722005897362335}}. 
	Fig. \ref{fig: lidar_traj} shows the trajectories of different Lidar-based methods. 
	Fig. \ref{fig: lidar_ate_rpe} shows their ATE and RPE statistics comparison results.
	LIO-SAM wins the ATE comparison while LEGO-LOAM is way ahead in the RPE. We find that fusing IMU data does not improve the accuracy, we think because the framerate of IMU data in KITTI dataset is low.
	Fig. \ref{fig: lidar_usage} shows the CPU and memory usage. Lightweight LEGO-LOAM and A-LOAM (optimized version of LOAM) have the least overhead, while NDT-LOAM and LOAM perform bad in resource usage. We guess that the former is because the algorithm is not optimized,  and the latter is because a costly point cloud registration algorithm is used.
	In summary, LIO-SAM fusing multiple sensors performs the best overall in terms of accuracy, but still slightly worse than LEGO-LOAM in resource usage.

	\begin{table}[!ht]
		\vspace{0.3cm}
		\centering
		\caption{ORB-SLAM2 RGB-D comparison under different parameters on freiburg2/desk sequence of TUM-RGBD dataset}
		\label{table: test2}
		\begin{threeparttable}
			\begin{tabular}{c|c|c|c|c}
				\hline
				\makecell[c]{The number of \\ ORB features }& \makecell[c]{RMSE ATE\tnote{1} \\(cm)} & \makecell[c]{RMSE RPE\tnote{1} \\ (cm)} & \makecell[c]{CPU\tnote{1} \\(core)} & \makecell[c]{RAM\tnote{1} \\(MB)}\\ \hline
				250\tnote{2}  & - & - & - & -\\ 
				500  & 0.881 & 1.27 & \textbf{1.23}/\textbf{1.96} & \textbf{733} \\ 
				750  & 0.692 & 1.23 & 1.28/2.24 & 767 \\ 
				1000 & 0.765 & 1.17& 1.56/3.33 & 868 \\ 
				1250 & 0.710 & 1.15 & 1.76/3.37 & 944 \\ 
				1500 & 0.647 & 1.15 & 1.90/3.52 & 989 \\ 
				1750 & 0.678 & \textbf{1.11} & 2.02/3.72 & 1045 \\ 
				2000 & 0.800& 1.16 & 2.07/3.76 & 1076 \\ 
				2250 & 0.675 & 1.16 & 2.16/3.74 & 1113 \\ 
				2500 & \textbf{0.564} & 1.13 & 2.31/3.85 & 1131 \\ \hline
			\end{tabular}% \textbf{\textcolor{red}{1.94}}
			\begin{tablenotes}
				\scriptsize
				\item[1] Our values are the average of 5 executions. RMSE ATE and RMSE RPE are obtained by aligning with the ground truth trajectory, where RPE is the error of translation per meter.
				\item[2] ORB-SLAM2 RGB-D fails when extracting 250 ORB feature points.
			\end{tablenotes}
		\end{threeparttable}
		\vspace{-0.4cm}
	\end{table}

\subsubsection{Correlation of Parameters to Performance}
Each SLAM method has some configuration parameters of its own and it is difficult to determine how much the change of the parameter value will affect the system. This experiment explores the impact of the number of ORB feature points extracted per image on accuracy, CPU and memory usage for the ORB-SLAM2 RGBD algorithm. All tests were performed on the \textit{freiburg2/desk} sequence of the TUM-RGBD dataset \cite{schubert2018tum}, a 99.36s, 18.88m, sequence with loop closure \footnote{\url{https://archive.slam-hive.net/analysis/show/1722262903200936}}.

As shown in Table \ref{table: test2}, as the number of feature points increases, the ATE and RPE tends to decrease, while the CPU and memory usage tends to increase. When only 250 feature points are extracted, the algorithm fails, indirectly indicating that if the scene has too few features, the performance of ORB-SLAM2 may no longer be robust.

	\subsection{Big vSLAM Experiment}

\begin{table*}[b]
	
	\scriptsize
	%	\tiny
	\centering
	\tabcolsep=0.14cm
	\renewcommand\arraystretch{1.3}
	\caption{Image Quality Parameter Space Exploration of Visual SLAM Algorithm on EuRoC Dataset}
	\label{table: exploration3}
	\begin{threeparttable}
		\begin{tabular}{c|cc|cc|cc|c}
			\hline
			\multirow{2}{*}{Mode} 
			& \multicolumn{2}{c|}{Algorithm} 
			& \multicolumn{2}{c|}{Image Resolution} 
			& \multicolumn{2}{c|}{Image Rate} 
			& \multirow{2}{*}{Configuration Number\tnote{1} } \\
			\cline{2-7}
			& number & value
			& number & value 
			& number & value (Hz) \\
			\hline \hline
			
			\multirow{1}{*}{Monocular} 
			&2 & ORB-SLAM2/ORB-SLAM3
			&6 & 1/0.8/0.6/0.5/0.4/0.2 
			&5 & 20/10/5/2/1
			& 60 \\
			\hline
			
			\multirow{1}{*}{Monocular Inertial} 
			&3 & ORB-SLAM3/VINS-Mono/VINS-Fusion
			&6 & 1/0.8/0.6/0.5/0.4/0.2 
			&5 & 20/10/5/2/1
			& 90 \\
			\hline
			
			\multirow{1}{*}{Stereo} 
			&3 & ORB-SLAM2/ORB-SLAM3/VINS-Fusion
			&6 & 1/0.8/0.6/0.5/0.4/0.2 
			&5 & 20/10/5/2/1
			& 90 \\
			\hline
			
			\multirow{1}{*}{Stereo Inertial} 
			&2 & ORB-SLAM3/VINS-Fusion
			&6 & 1/0.8/0.6/0.5/0.4/0.2 
			&5 & 20/10/5/2/1
			& 60 \\
			\hline
			\hline
			
			\multirow{1}{*}{All} 
			& \multicolumn{1}{c}{} 
			& \multicolumn{1}{c}{} 
			& \multicolumn{1}{c}{} 
			& \multicolumn{1}{c}{} 
			&  &  
			& 300*5 = 1500 \\
			\hline
			
		\end{tabular}

	\end{threeparttable}

\end{table*}

\begin{table*}[b]
	\scriptsize
	%	\tiny
	\centering
	\tabcolsep=0.14cm
	\renewcommand\arraystretch{1.3}
	\caption{Best accuracy performances}
	\label{table: comparison2}
	\begin{threeparttable}
		\resizebox{\linewidth}{!} {
			\begin{tabular}{c|l|ccc|ccc|ccc|ccc|ccc}
				\hline
				\multirow{2}*{\makecell[c]{Sensor\\Combinations}} & \multirow{2}*{Algorithms}
				&\multicolumn{3}{c|}{MH\_01\_easy} & \multicolumn{3}{c|}{MH\_02\_easy} & \multicolumn{3}{c|}{MH\_03\_medium} & \multicolumn{3}{c|}{MH\_04\_difficult} & \multicolumn{3}{c}{MH\_05\_difficult}\\
				\cline{3-17} && RMSE\tnote{1} & CPU\tnote{2} / RAM\tnote{3} & FPS\tnote{4} / Res\tnote{5} & RMSE & CPU / RAM & FPS / Res & RMSE & CPU / RAM & FPS / Res & RMSE & CPU / RAM & FPS / Res & RMSE & CPU / RAM & FPS / Res \\
				
				\hline \hline 
				\multirow{2}*{Monocular}
				&ORB-SLAM2& \textbf{0.042} & 1.09/827 & 20/0.6 & \textbf{0.033} & 1.04/804 & 20/0.6 & \textbf{0.035} & 1.04/773 & 20/0.8 & \textbf{0.051} & 1.04/779 & 20/0.8 & 0.049 & 1.04/744 & 10/1\\
				%			&DSO      &   &  &  &  & &  &  &  &  & & & &&&\\
				&ORB-SLAM3& 0.045 & 0.89/865 & 10/0.5 & \textbf{0.033} & 1.08/1539 & 20/1 & 0.036 & 1.01/849 & 20/1 & {0.052} & 0.63/754 & 5/0.8 & \textbf{0.048} & {1.04}/981 &20/1\\
				
				\hline \hline 
				\multirow{3}*{\makecell[c]{Monocular\\ Inertial}}
				&VINS-Mono& 0.086 & 1.30/1500 & 10/1 & 0.101 & 1.25/1331 & 10/1 & 0.098 & 1.19/1156 & 20/0.8 & 0.095 & 1.08/1054 & 10/1 & 0.153 & 1.10/1053 & 10/1\\
				&VINS-Fusion& \textbf{0.040} & 0.90/1193 & 10/1 & \textbf{0.031} & 0.88/1097 & 10/1 & 0.054 & 0.85/1017 & 10/1 & 0.097 & \textbf{1.19}/1063 & 20/1 & 0.103 & 0.79/886 & 10/1\\
				&ORB-SLAM3& {0.040} & 1.51/919 & 20/1 & 0.033 & 1.07/878 & 20/1 & \textbf{0.040} & 0.95/705 & 20/0.8 & \textbf{0.084} & 1.30/836 & 20/1 & \textbf{0.059} & 1.27/870 & 20/1\\
				
				\hline \hline 
				\multirow{3}*{Stereo}
				&ORB-SLAM2& {0.036} & 2.47/1026 & 20/1 & 0.034 & 1.79/802 & 10/1 & 0.046 & 2.18/771 & 20/0.8 & \textbf{0.063} & 2.65/925 & 20/1 & \textbf{0.048} & 2.80/1042 & 20/1\\
				&VINS-Fusion & \textbf{0.025} & 1.67/1287 & 20/1 & 0.054 & 1.01/830 & 20/0.6 & 0.047 &1.68/1185 & 20/1 & 0.106 & 1.17/861 & 20/0.8 & 0.084 & 1.49/1047 & 20/1 \\
				&ORB-SLAM3& 0.044 & 1.51/979 & 20/1 & \textbf{0.033} & {1.07}/878 & 20/1 & \textbf{0.040} & {0.95}/705 & 20/0.8 & 0.084 & {1.30}/836 & 20/1 & 0.059 & 1.27/870 & 20/1 \\
				
				\hline \hline 
				\multirow{2}*{\makecell[c]{Stereo\\ Inertial}}
				&VINS-Fusion& \textbf{0.031} & {1.07}/1201 & 10/1 & \textbf{0.019} & 1.76/1102 & 20/1 & \textbf{0.038} & 1.01/1023 & 10/1 & 0.105 & 1.56/1065 & 20/1 & 0.073 & {0.95}/896 & 10/1\\
				&ORB-SLAM3& {0.034} & 1.34/811 & 5/0.8 & 0.040 & 1.50/849 & 10/1 & 0.041 & 1.40/811 & 5/0.8 & \textbf{0.049} & 1.65/993 & 20/0.8 & \textbf{0.058} & 1.79/898 & 20/1 \\
				\hline 
				%		\hline \hline 
				%		\multirow{1}*{Lidar Inertial}
				%		&LIO-SAM&  &  &  &  & &  &  &  &  & & & \\
				%		\hline
			\end{tabular}
		}
		
		\begin{tablenotes}
			%			\footnotesize
			\scriptsize
			%			\tiny
			% \item[1] They are five sequences of varying difficulty on the EuRoC dataset. \textit{MH\_01\_easy} and \textit{MH\_02\_easy} are with good texture and bright scene, \textit{MH\_03\_medium} is with fast motion and bright scene, \textit{MH\_04\_difficult} and \textit{MH\_05\_difficult} are with fast motion and dark scene.
			\item[1] RMSE is the root mean square error of ATE, the unit is meter.
			\item[2] The calculation of CPU usage is the sum of all core usage. CPU in the above table is average CPU usage during mapping, its unit is core. 
			\item[3] RAM is the maximum memory usage during mapping, its unit is MB.
			\item[4] FPS is the framerate of the image data, its unit is Hz.
			\item[5] Res is the scale relative to the resolution of the original image data, and the original resolution is: $752$ $\times$ $480$.
			\item[6] For each algorithm, we use its default parameters.

			%			\item[5] Avg is the average RMSE, CPU and RAM of 5 sequences.
			
		\end{tablenotes}
	\end{threeparttable}
	%\vspace{-0.6cm}
\end{table*}

\begin{figure*}[t]
	\centering
	% \hspace{-1cm}
	\includegraphics[width=7.0in]{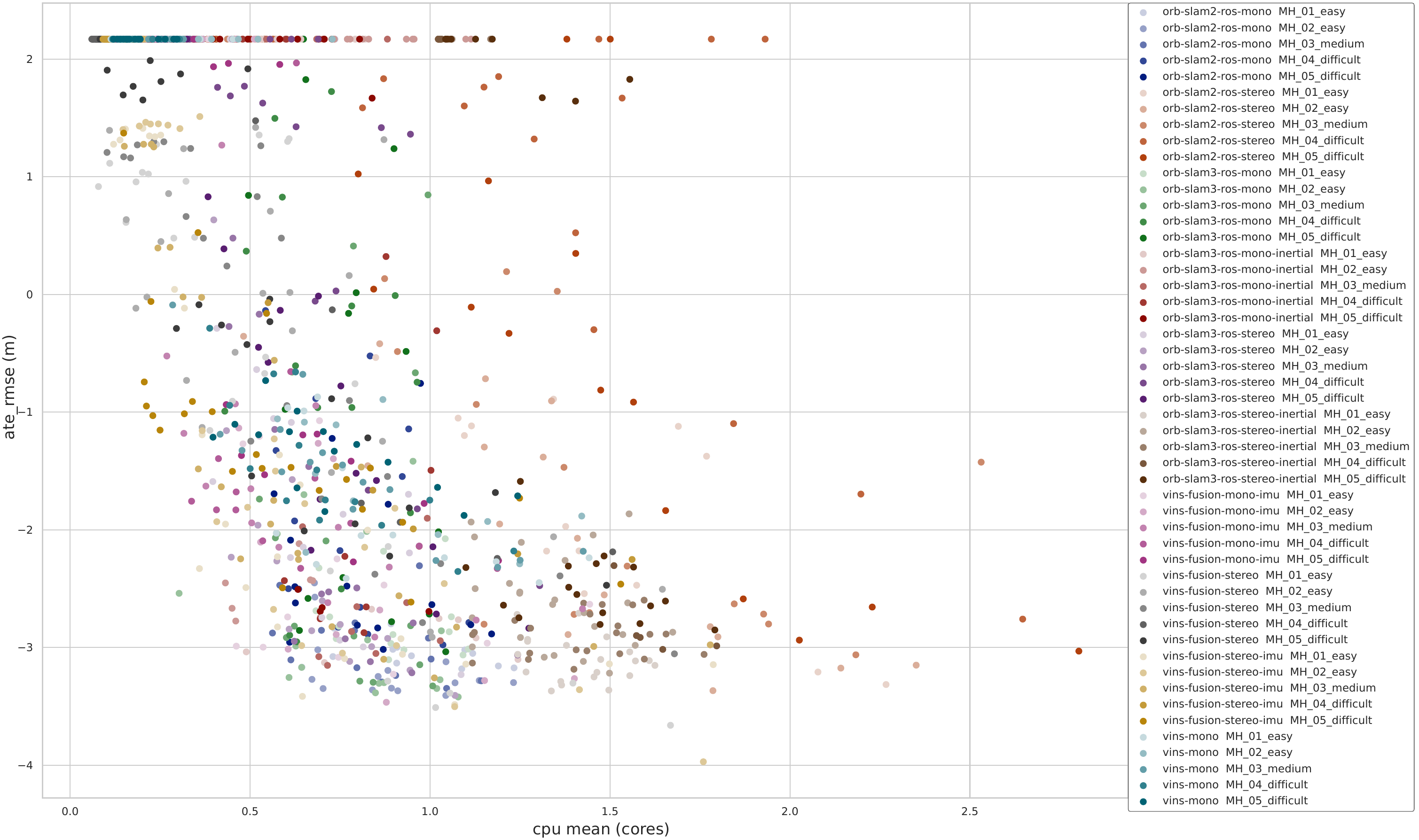}
	\caption{Comparison between accuracy and resource usage metrics of Visual-based SLAM algorithms.} %有点不清晰
	\label{fig: exploration1_1}
		\vspace{-0.4cm}

\end{figure*}

To highlight the capabilities of SLAM Hive and to investigate vSLAM algorithms in detail we finally perform one big experiment with 1,500 mapping runs. It investigates the four vSLAM algorithms (ORB-SLAM2 and 3 and VINS-Mono and VINS-Fusion)  in detail on the same five sequences of the   EuRoC dataset as above. The difference is, that we investigate the effect of image resolution (6 sizes) and frame rate (5 rates). The 10 different algorithm-sensor data pairs multiplied by 6 resolutions and 5 frame rates over 5 dataset gives 1,500 mapping runs that were performed, as shown in Table \ref{table: exploration3}. SLAM Hive is providing scaled down versions of the original image resolution of 752 x 480, multiplied with factors $1/ 0.8/ 0.6/ 0.5/ 0.4/ 0.2$ - so the smallest resolution is 150 x 96. The frame rate is just taking every n-th image, from the original 20Hz data over 10Hz, 5Hz, 2Hz down to 1Hz. 
%
%
	% 这个实验探究了图像数据集的质量（包括图像的频率和分辨率）对于Visual-based SLAM算法的精度以及资源使用率的影响。
%	This experiment explores how the quality of Image data (resolution and framerate) affects Visual-based SLAM algorithms.
%	As described above, if algorithms loose tracking, we utilize our Extended metric parameters to multiply the ATE and RPE for mapping runs with the following internal ranges $[1, 0.75), [0.75, 0.5), [0.5, 0.25), [0.25, 0]$ by the following factors $[1, 2, 2.5, 3]$.
	
	Often vSLAM algorithms loose tracking and thus return pose estimates for a subset of the input frames. We calculated the traj\_length factor after the mapping run, as a factor of estimated number of frames vs. ground truth frames. EVO can report good ATE-RMSE even for runs that have a very low traj\_length, e.g. 0.2. We would like to consider those runs as failed and thus mark of all runs with a traj\_length below 0.75 as failed. Out of the 1500 mapping runs, coincidentally exactly half, 750, are marked as failed this way.

	Fig. \ref{fig: exploration1_1} shows a diagram of CPU mean vs. ATE RMSE. All failed runs are displayed with a ATE value 20\% above the maximum ATE of the good runs. The color indicates the mapping run as combination of algorithm and dataset. Those colors are hard to distinguish, but the interactive web version will show the details of the mapping run when hovering over the entry \footnote{\url{https://archive.slam-hive.net/analysis/show/1731141820841522} Click on "Show Dynamic Diagram" to display the interactive diagram}.
    The diagram shows a general trend that better ATE values require more CPU resources, even though there are some approaches that achieve good results with fewer calculations (e.g. VINS Fusion with less than half a CPU and 0.5 ATE RMSE).

	Table \ref{table: comparison2} shows the best ATE-RMSE for each algorithm/ dataset combination, and the according CPU and memory usage as well as the framerate and image resolution used. Basically it shows that all tested algorithms can perform well on the 5 sequences. The difficult sequences have an ATE-RMSE of just about 0.01 up to 0.04 worse value. Future experiments should utilize datasets that are longer and more challenging for state-of-the-art algorithms. The expectation was, that full resolution (1.) and full framerate (20Hz) would give the best results. This is not always the case, but this is due to the fact that the successful results are very close to each other, such that it was a random choice of which ones had the actual highest number. 
	
\begin{figure*}[!h]
		\centering
		
		\captionsetup[subfloat]{font=scriptsize}
		
		\subfloat[Mono: FPS vs CPU vs ATE]{
			\label{fig:subfig:onefunction311} 
			\includegraphics[width=1.6in]{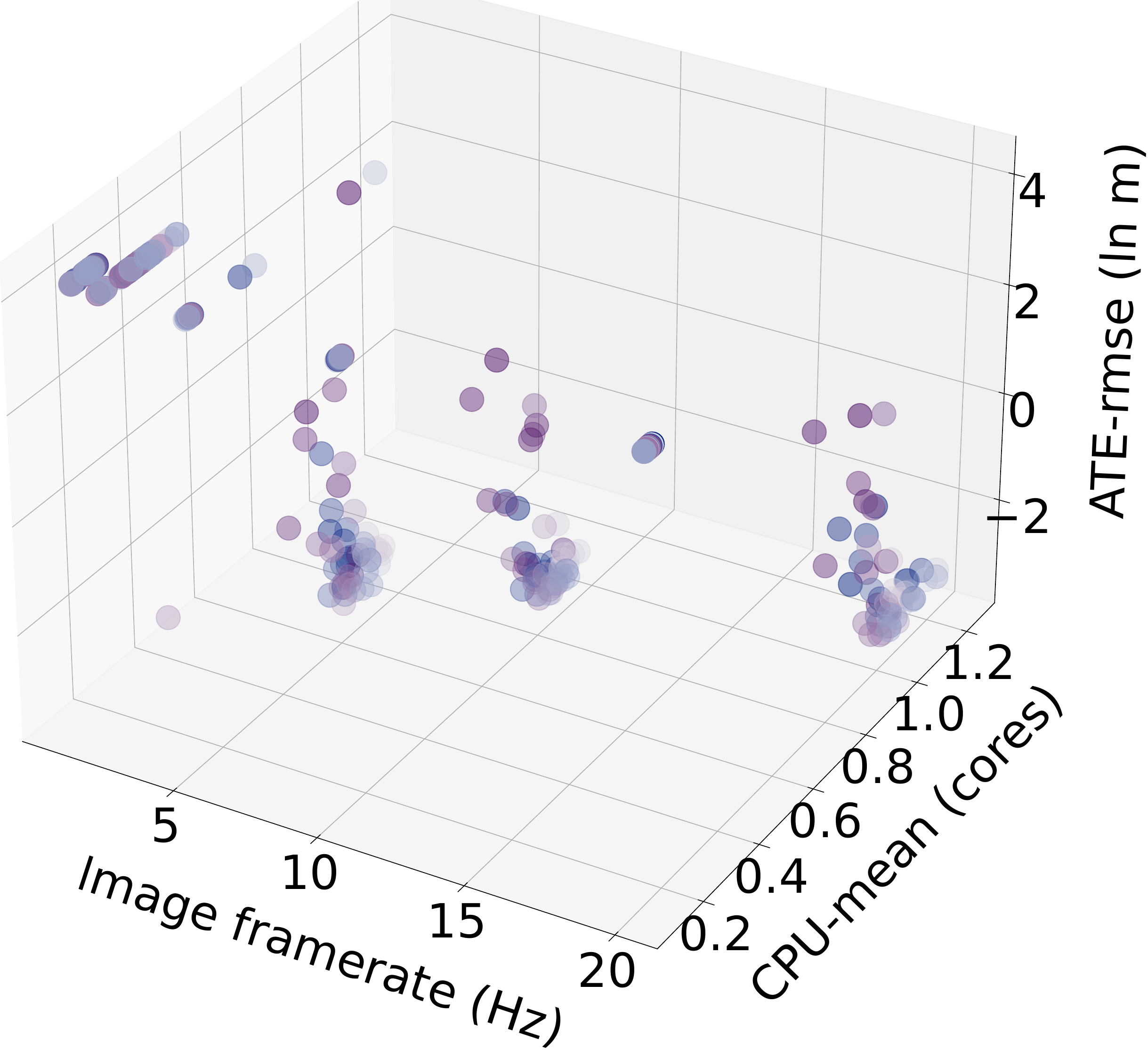}}
		\subfloat[Mono-IMU: FPS vs CPU vs ATE]{
			\label{fig:subfig:onefunction312} 
			\includegraphics[width=1.6in]{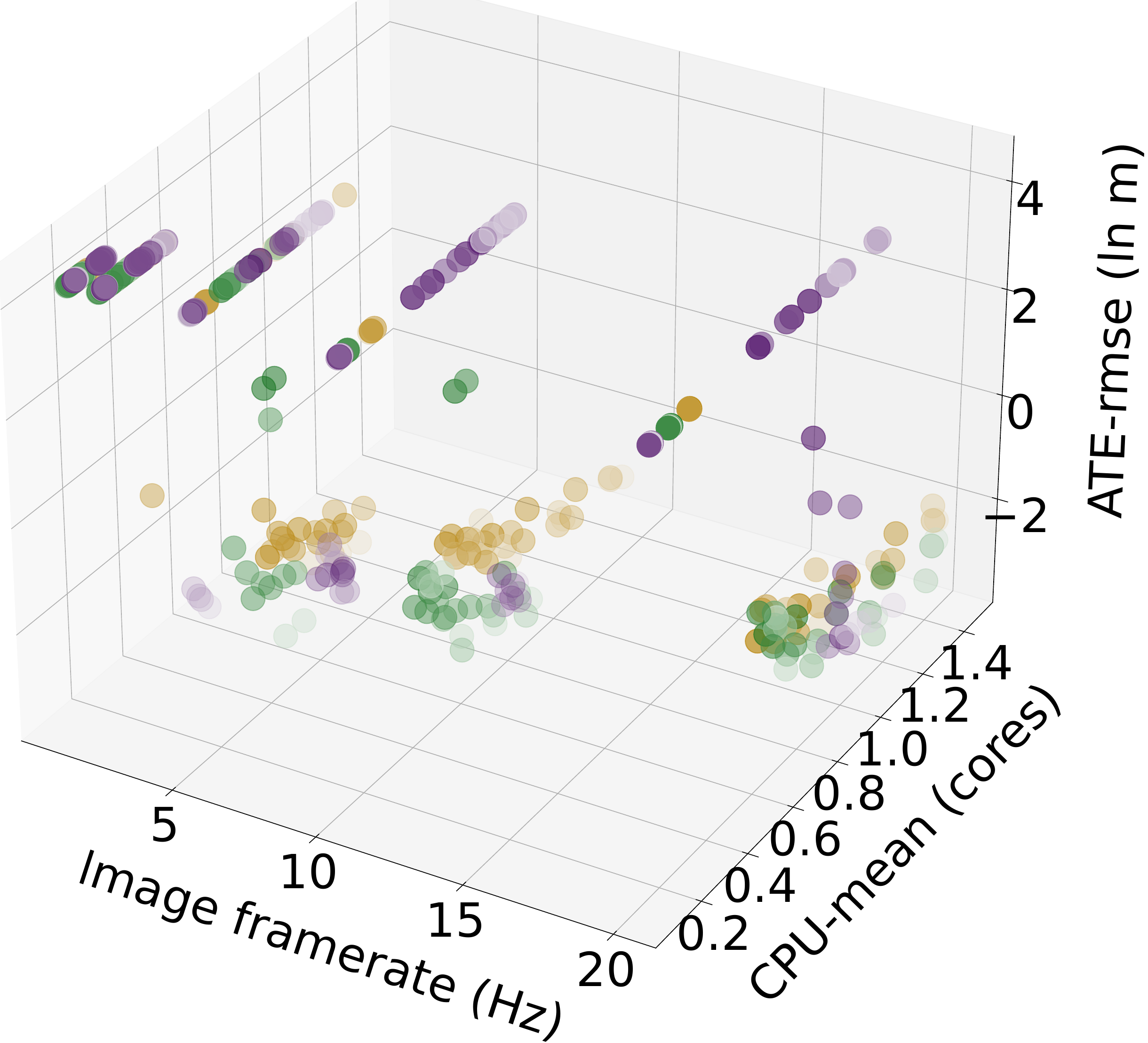}}
		\subfloat[Stereo: FPS vs CPU vs ATE]{
			\label{fig:subfig:onefunction313} 
			\includegraphics[width=1.6in]{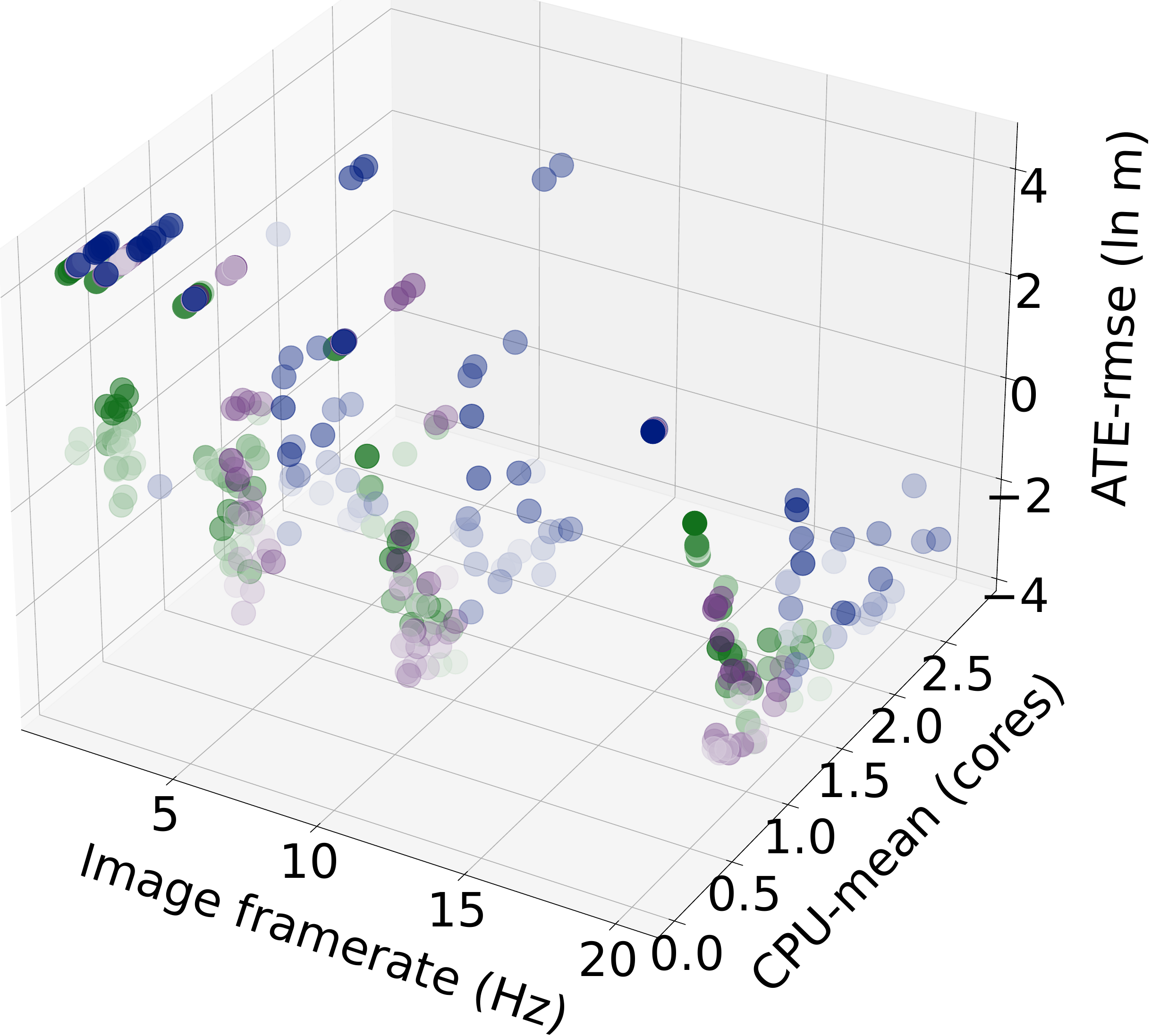}}
		\subfloat[Stereo-IMU: FPS vs CPU vs ATE]{
			\label{fig:subfig:onefunction314} 
			\includegraphics[width=1.6in]{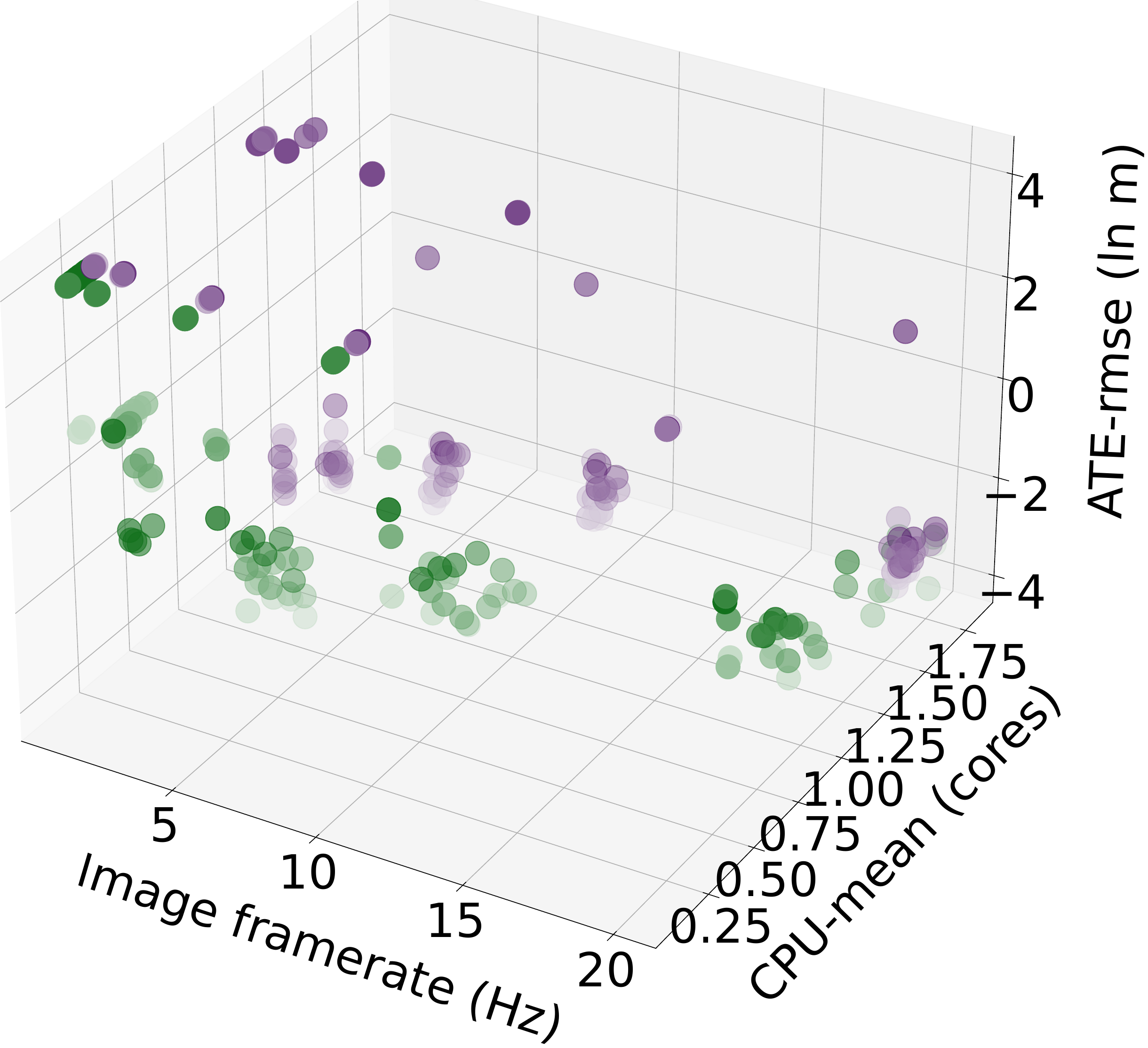}}
		\newline
		
		\subfloat[Mono: FPS vs Mem vs ATE]{
			\label{fig:subfig:twofunction321} 
			\includegraphics[width=1.6in]{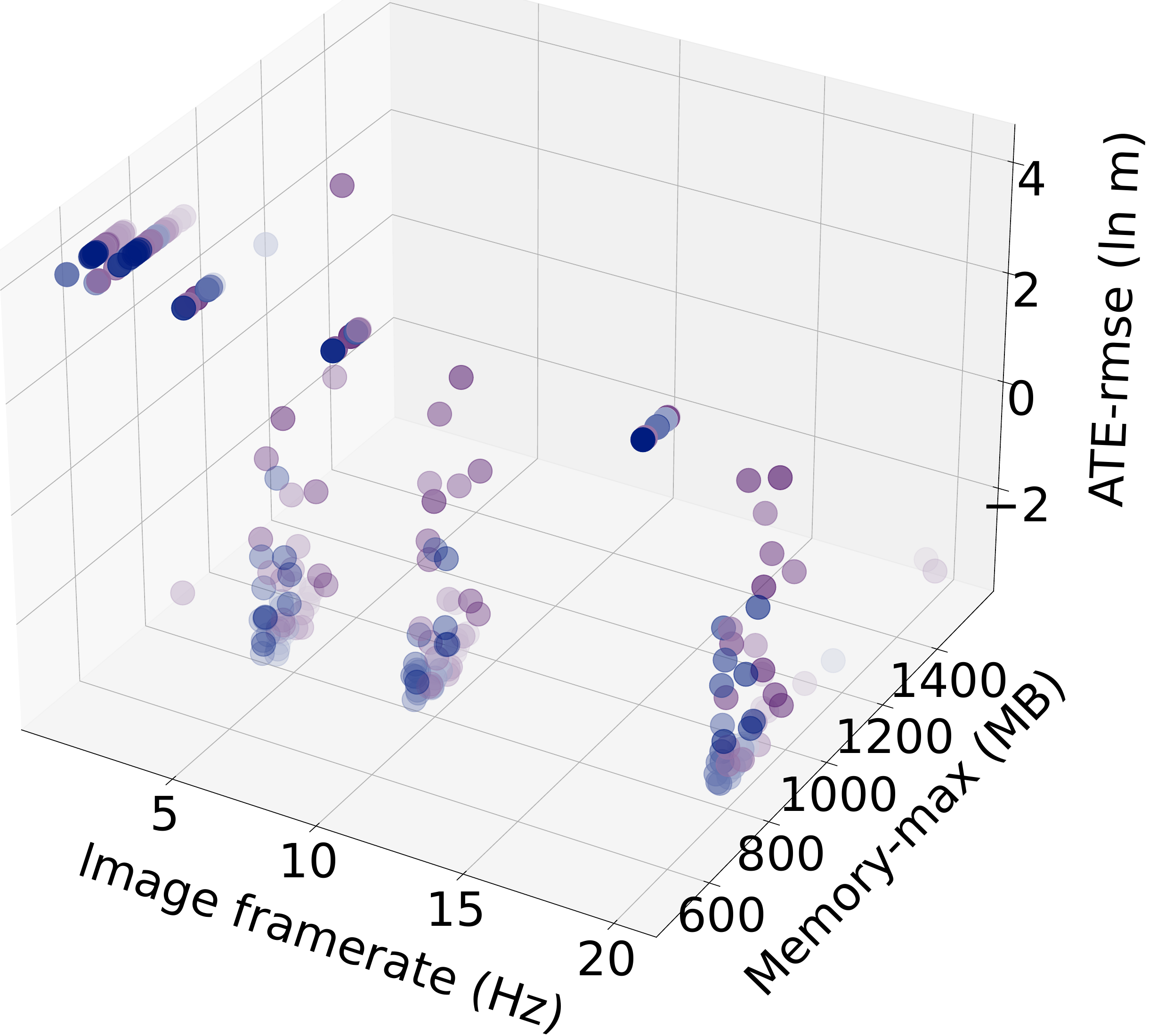}}
		\subfloat[Mono-IMU: FPS vs Mem vs ATE]{
			\label{fig:subfig:twofunction322} 
			\includegraphics[width=1.6in]{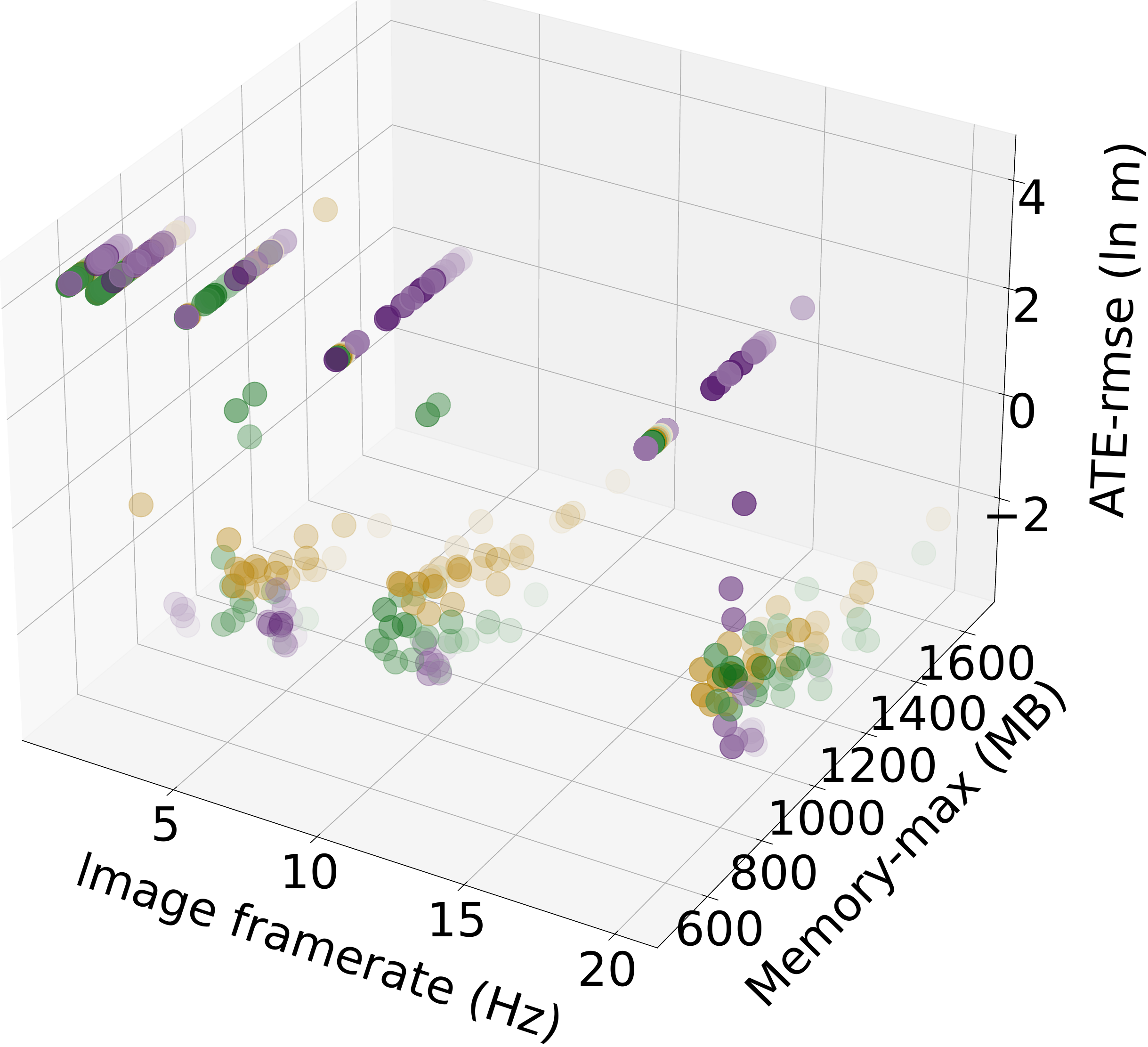}}
		\subfloat[Stereo: FPS vs Mem vs ATE]{
			\label{fig:subfig:twofunction323} 
			\includegraphics[width=1.6in]{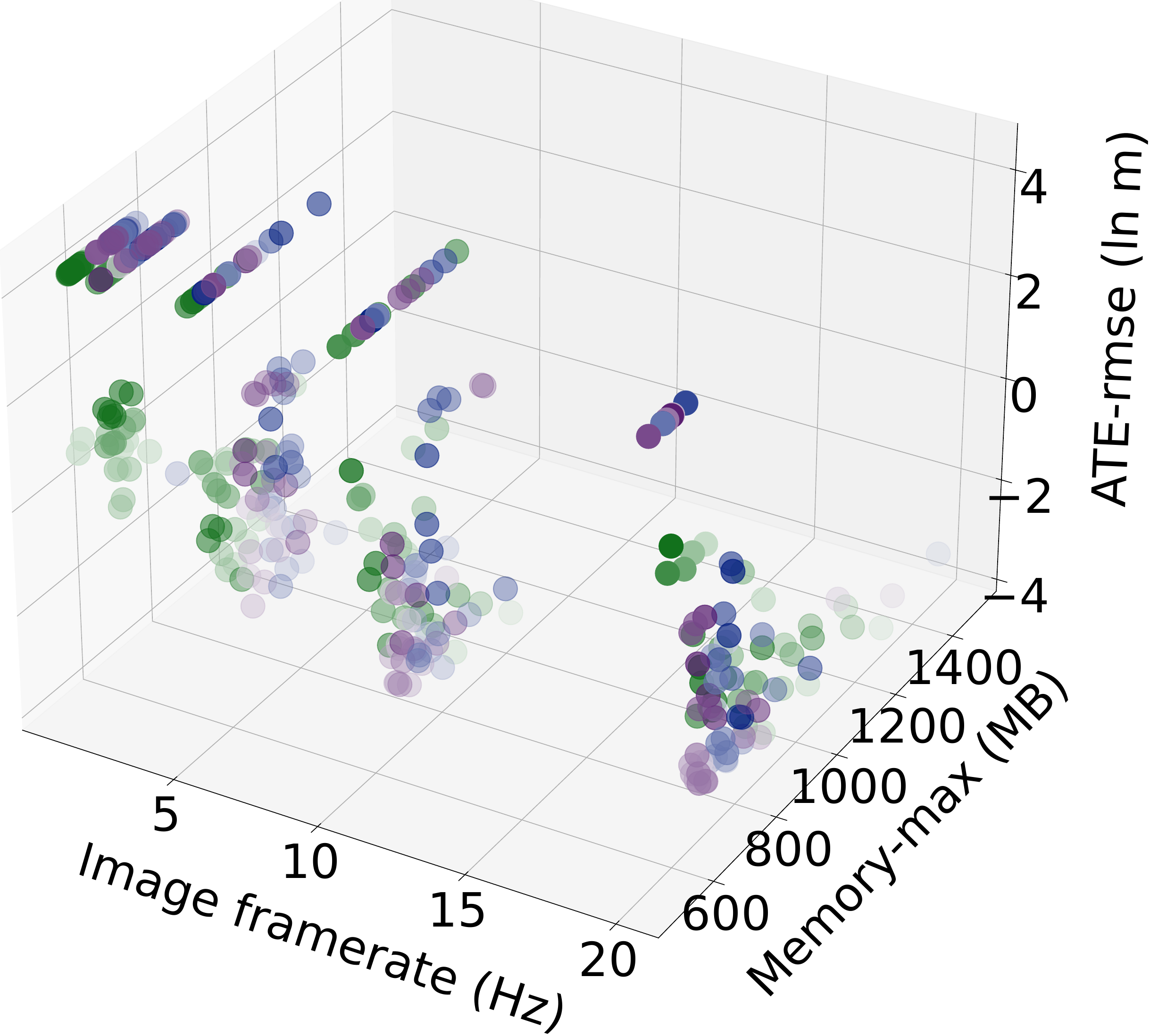}}
		\subfloat[Stereo-IMU: FPS vs Mem vs ATE]{
			\label{fig:subfig:twofunction324} 
			\includegraphics[width=1.6in]{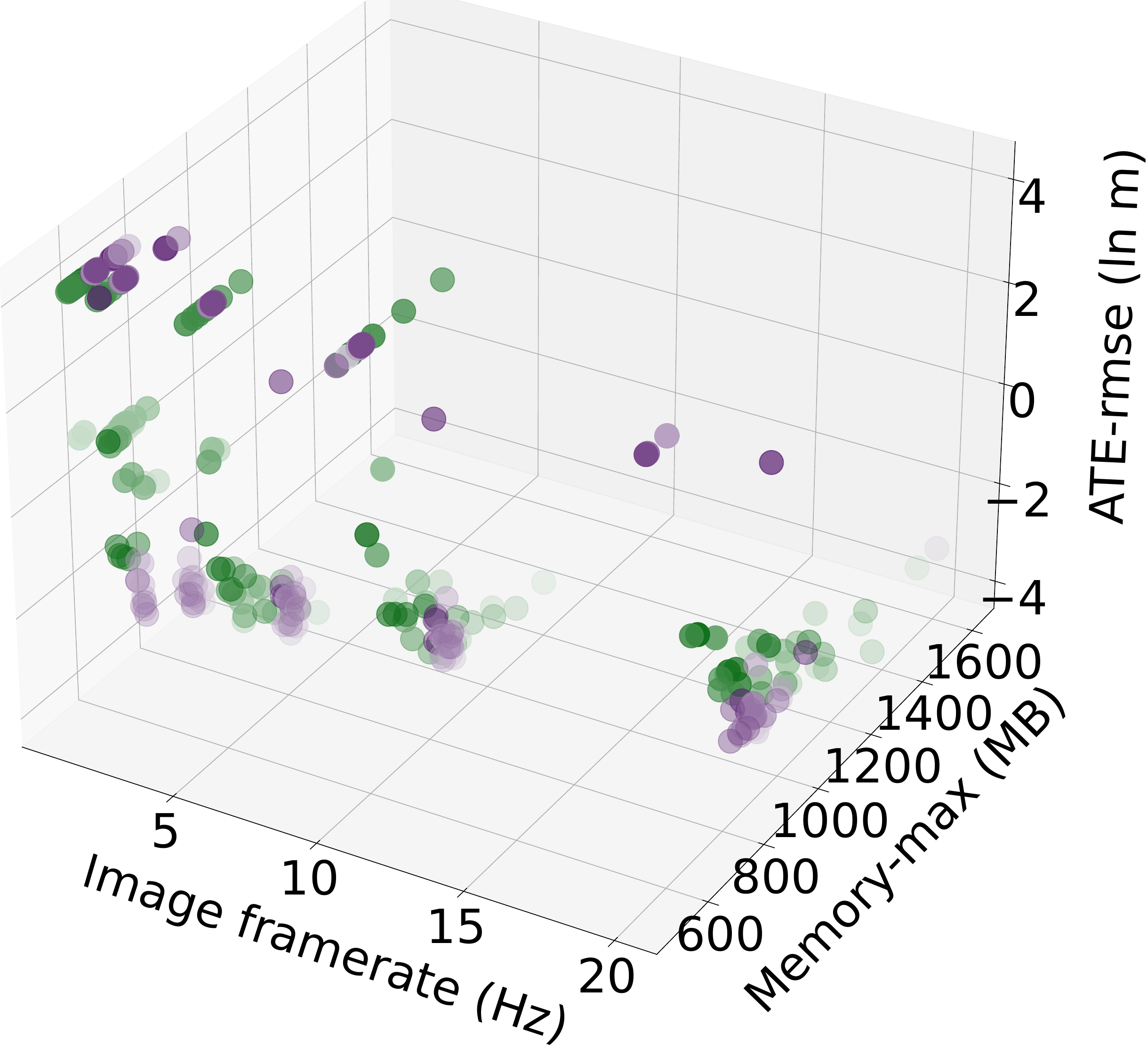}}
		\newline
		
		\subfloat[Mono: Res vs CPU vs ATE]{
			\label{fig:subfig:onefunction331} 
			\includegraphics[width=1.6in]{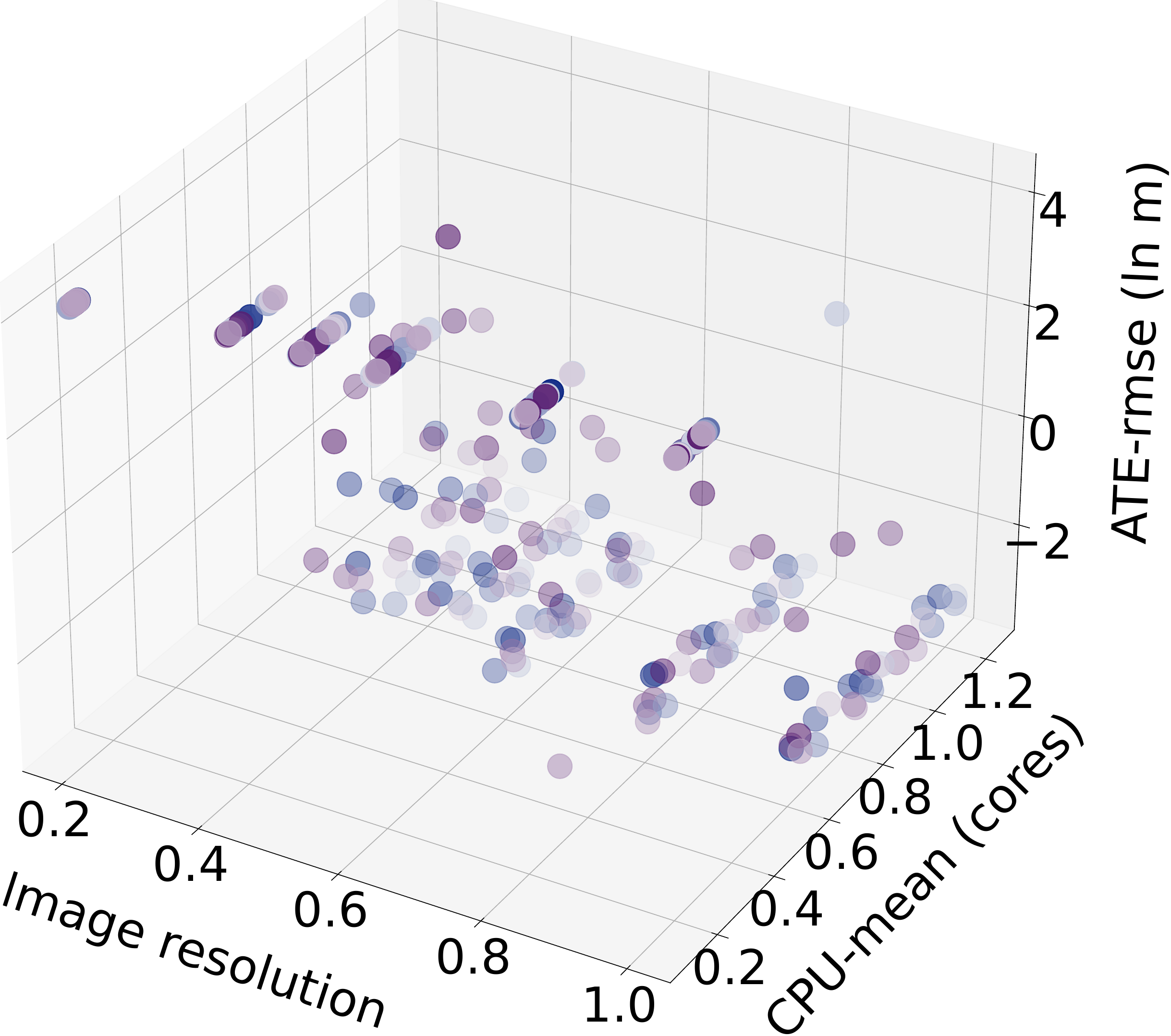}}
		\subfloat[Mono-IMU: Res vs CPU vs ATE]{
			\label{fig:subfig:onefunction332} 
			\includegraphics[width=1.6in]{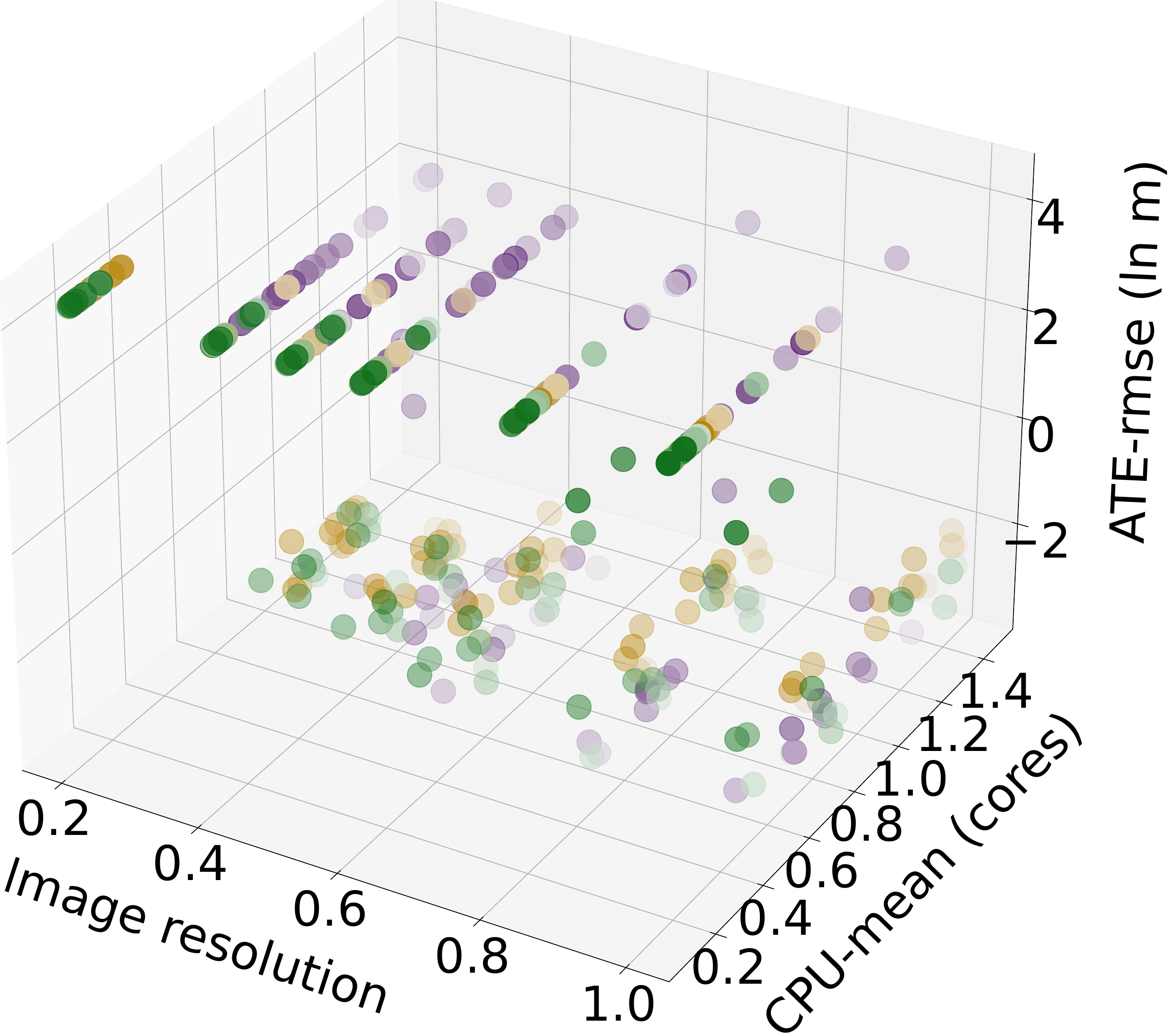}}
		\subfloat[Stereo: Res vs CPU vs ATE]{
			\label{fig:subfig:onefunction333} 
			\includegraphics[width=1.6in]{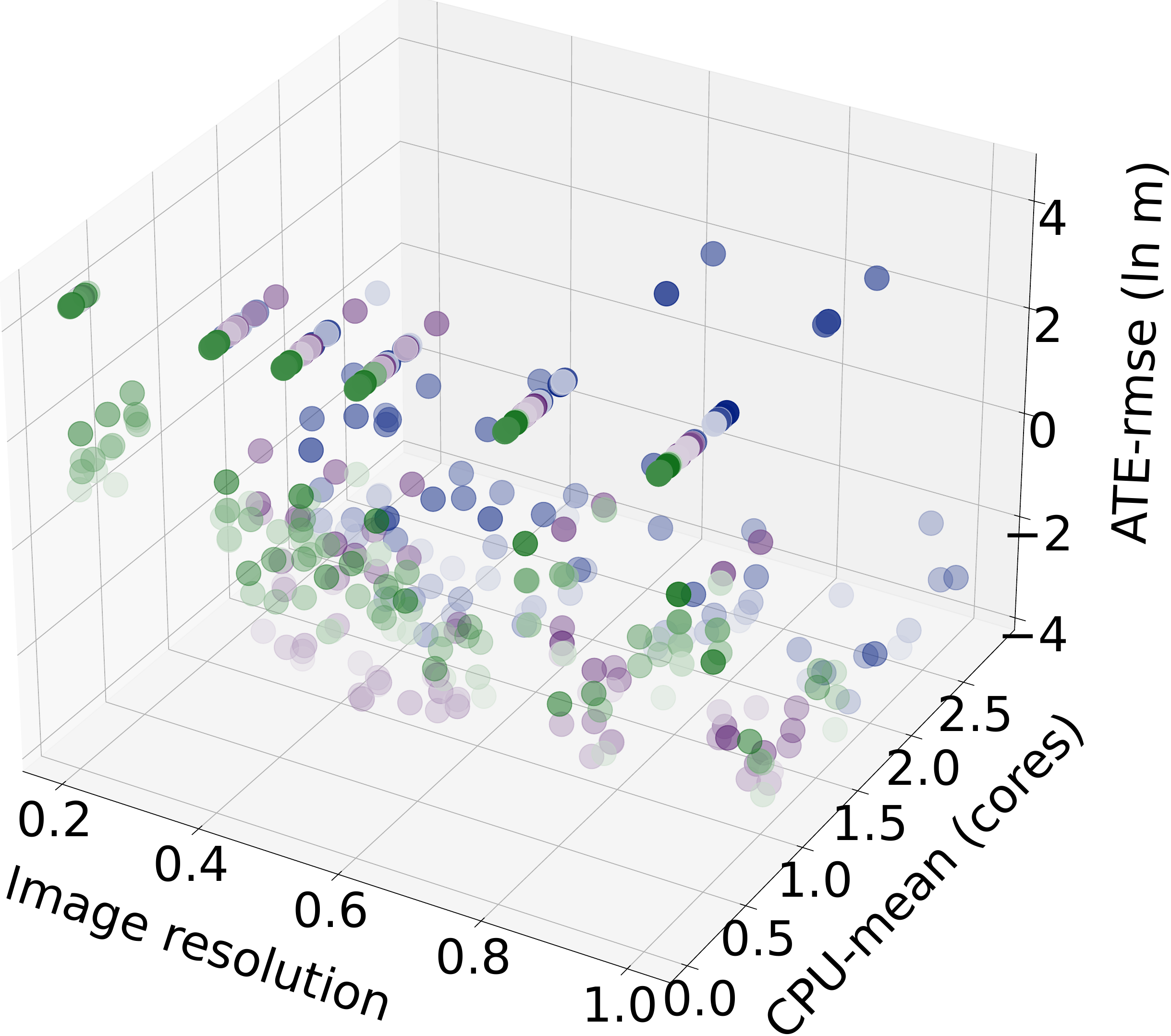}}
		\subfloat[Stereo-IMU: Res vs CPU vs ATE]{
			\label{fig:subfig:onefunction334} 
			\includegraphics[width=1.6in]{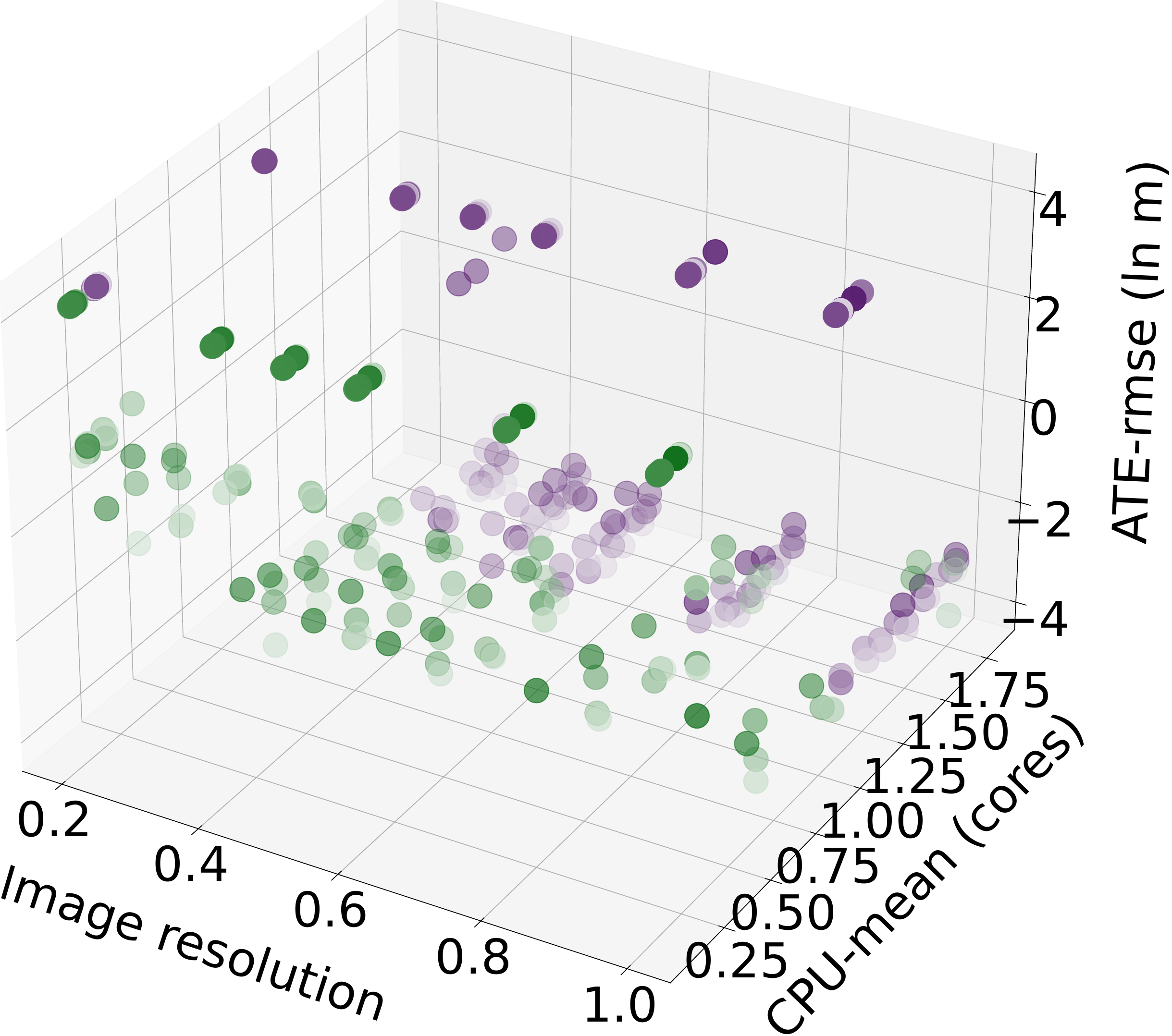}}
		\newline
		
		\subfloat[Mono: Res vs Mem vs ATE]{
			\label{fig:subfig:twofunction341} 
			\includegraphics[width=1.6in]{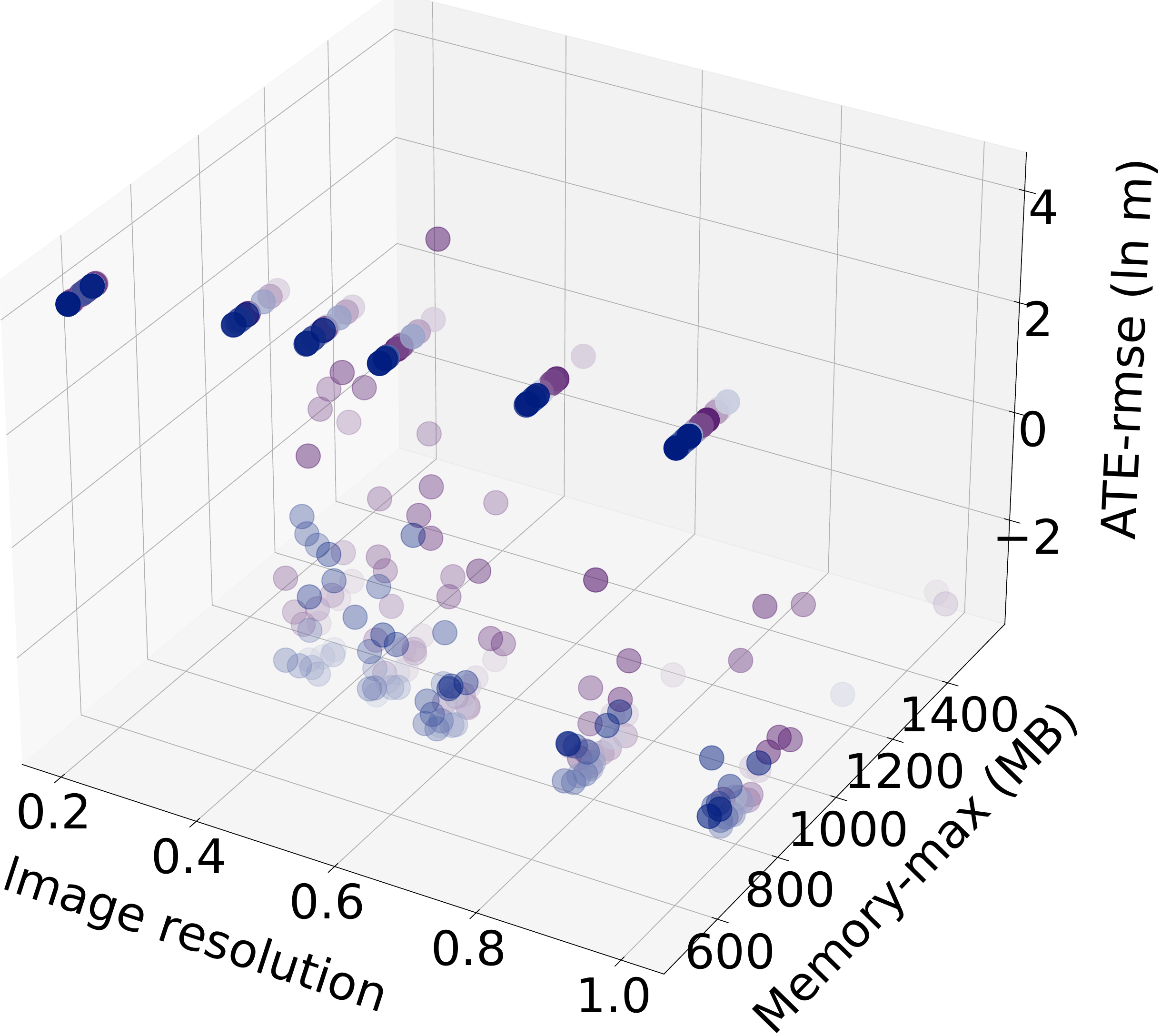}}
		\subfloat[Mono-IMU: Res vs Mem vs ATE]{
			\label{fig:subfig:twofunction342} 
			\includegraphics[width=1.6in]{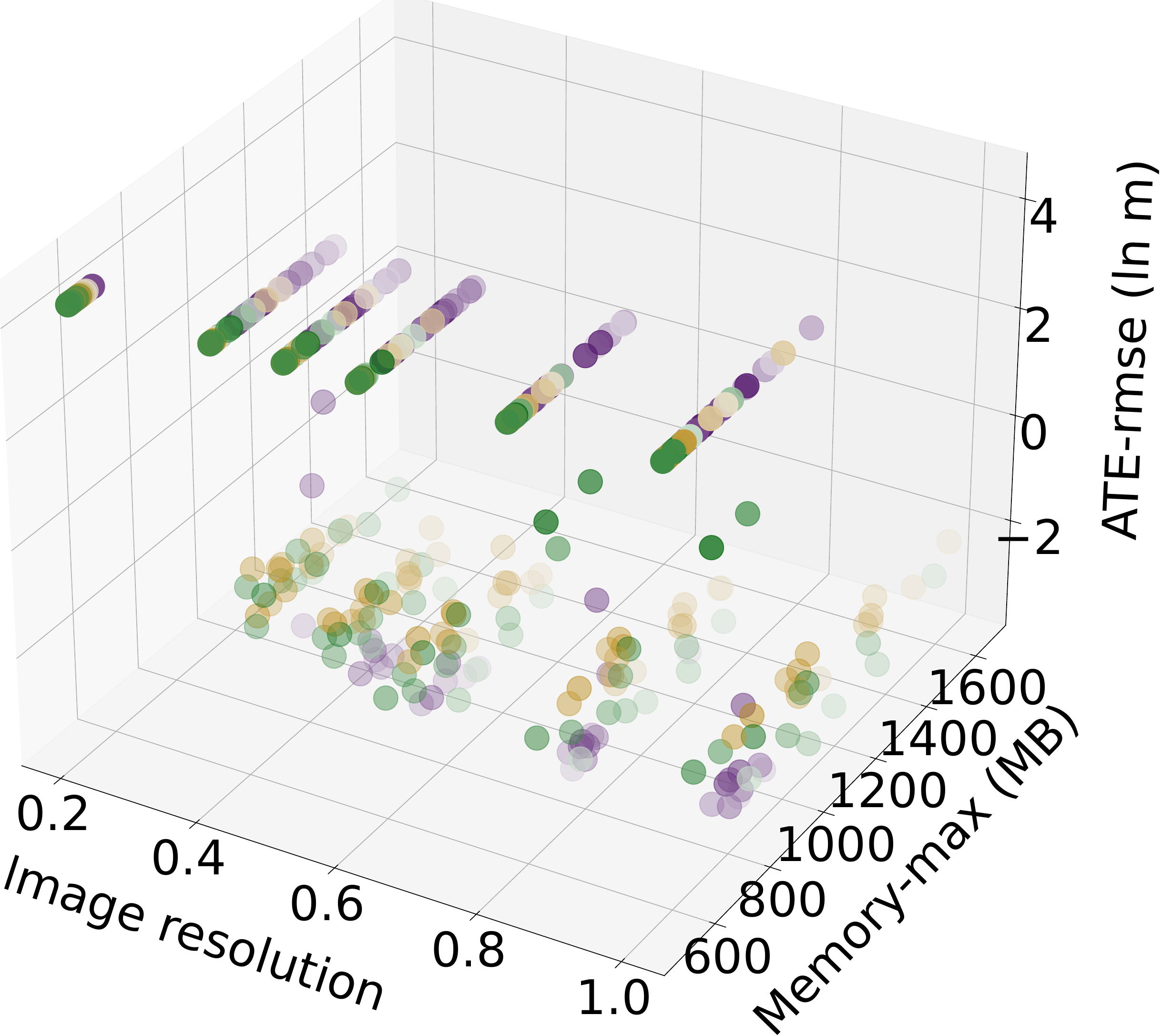}}
		\subfloat[Stereo: Res vs Mem vs ATE]{
			\label{fig:subfig:twofunction343} 
			\includegraphics[width=1.6in]{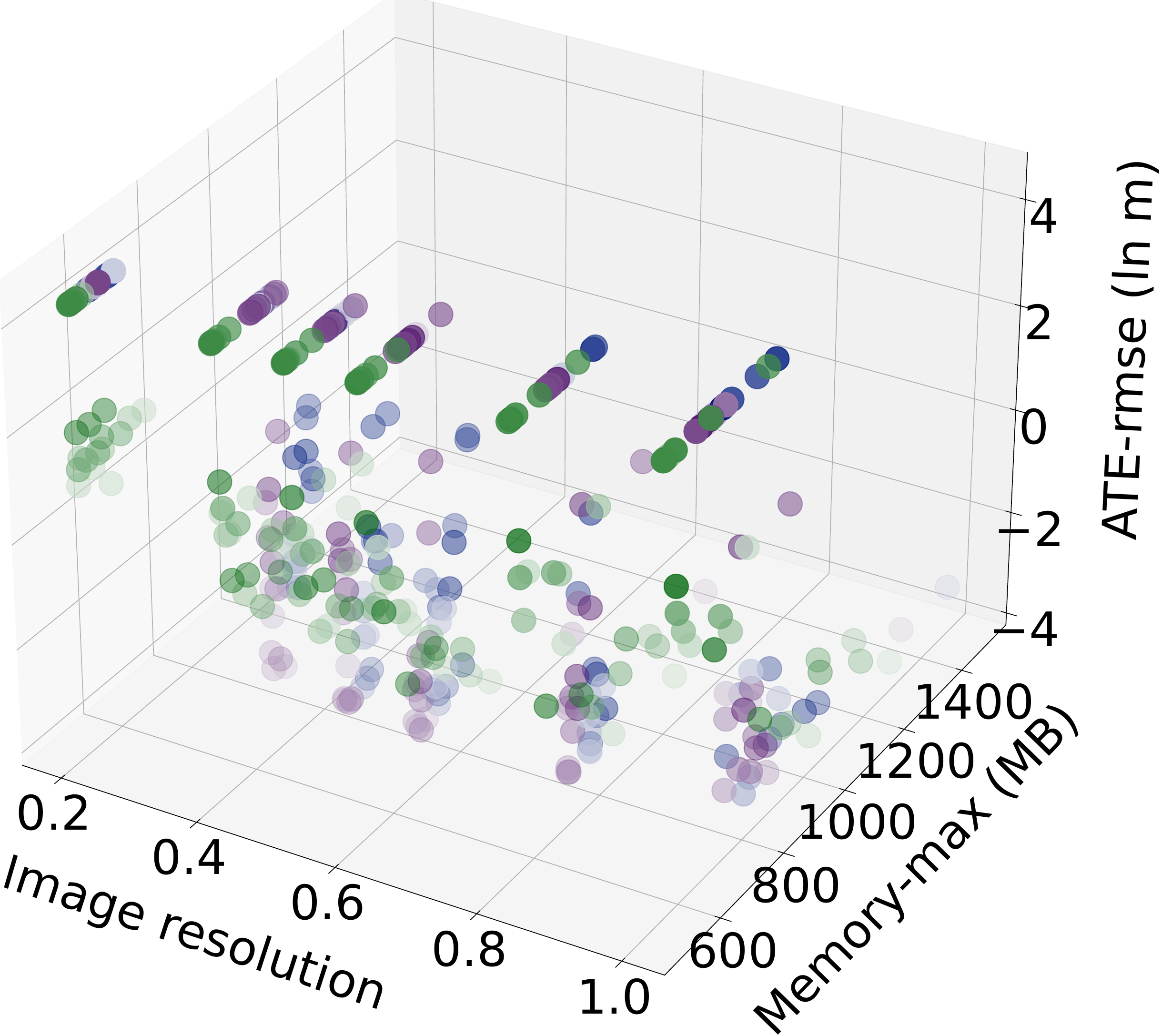}}
		\subfloat[Stereo-IMU: Res vs Mem vs ATE]{
			\label{fig:subfig:twofunction344} 
			\includegraphics[width=1.6in]{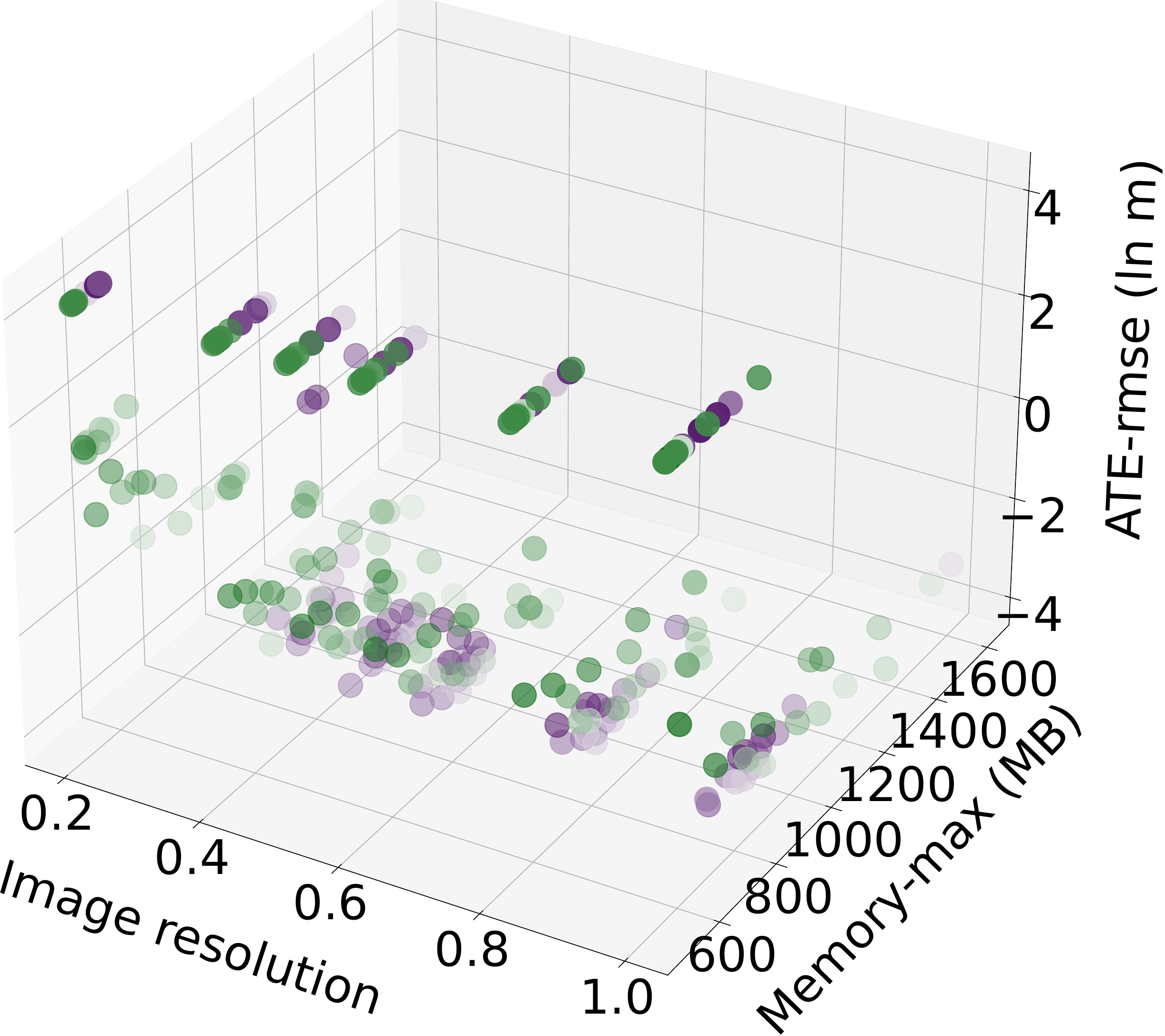}}
		\newline
		
		\subfloat[Mono: Res vs FPS vs ATE]{
			\label{fig:subfig:twofunction351} 
			\includegraphics[width=1.6in]{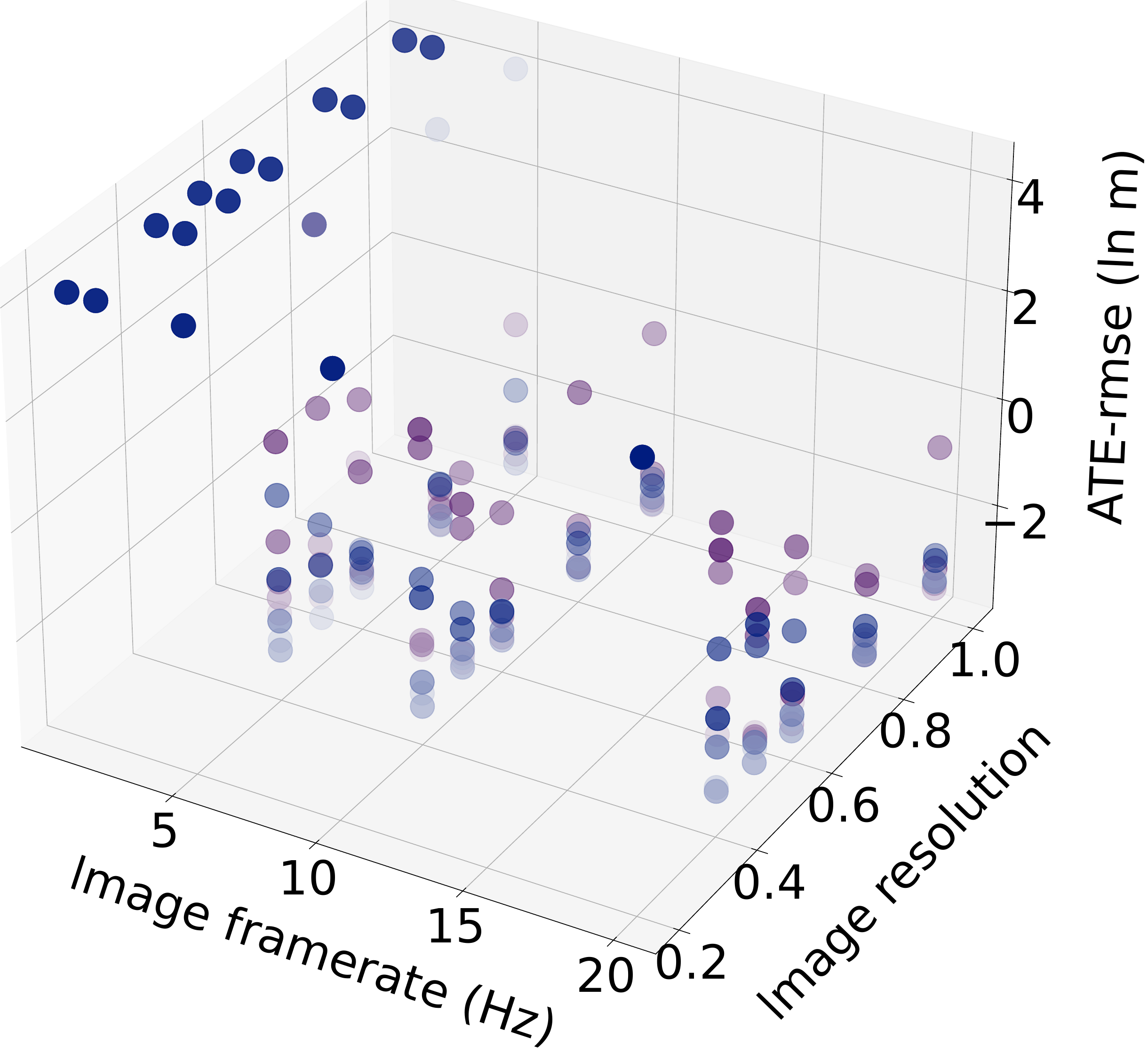}}
		\subfloat[Mono-IMU: Res vs FPS vs ATE]{
			\label{fig:subfig:twofunction352} 
			\includegraphics[width=1.6in]{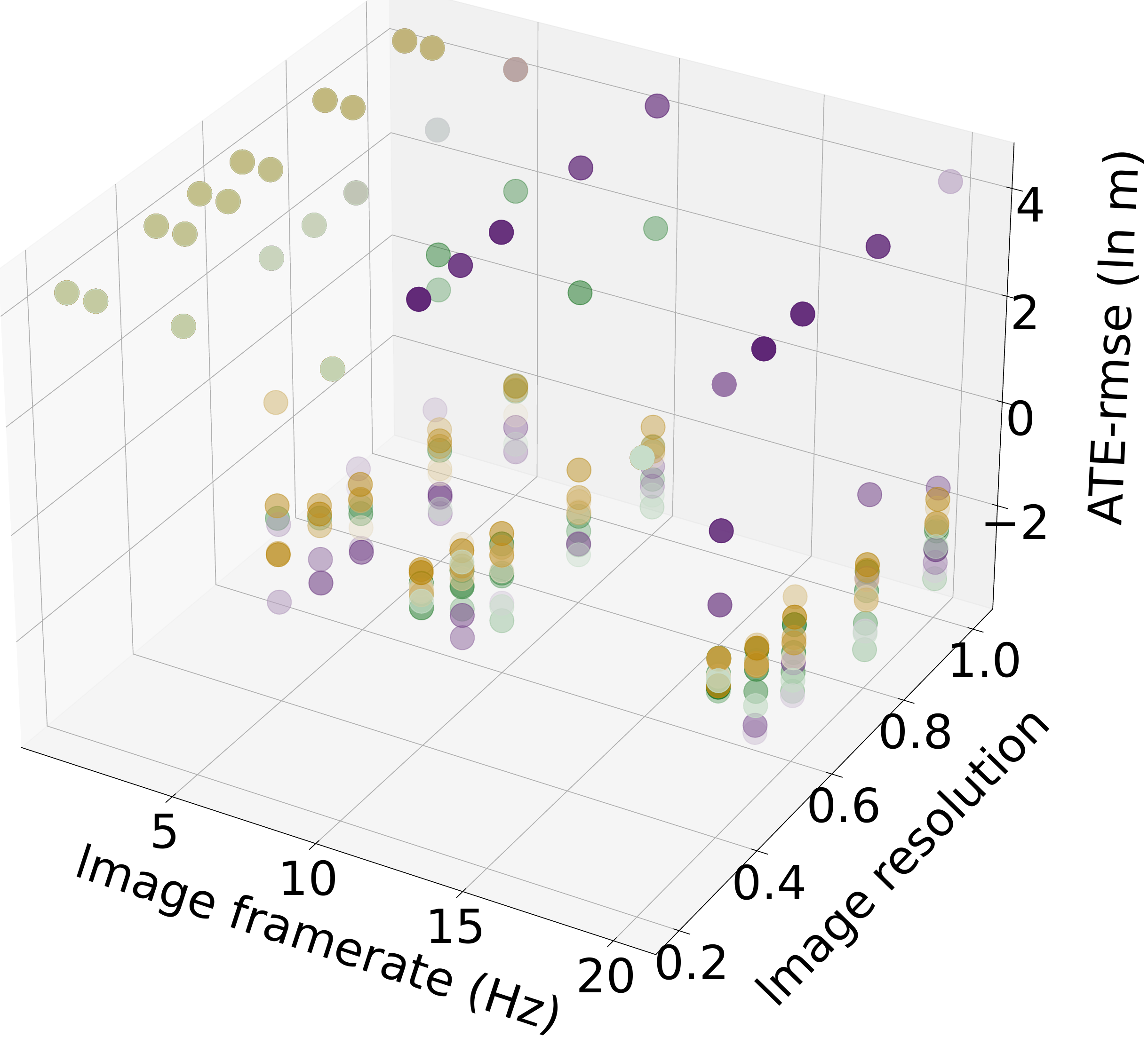}}
		\subfloat[Stereo: Res vs FPS vs ATE]{
			\label{fig:subfig:twofunction353} 
			\includegraphics[width=1.6in]{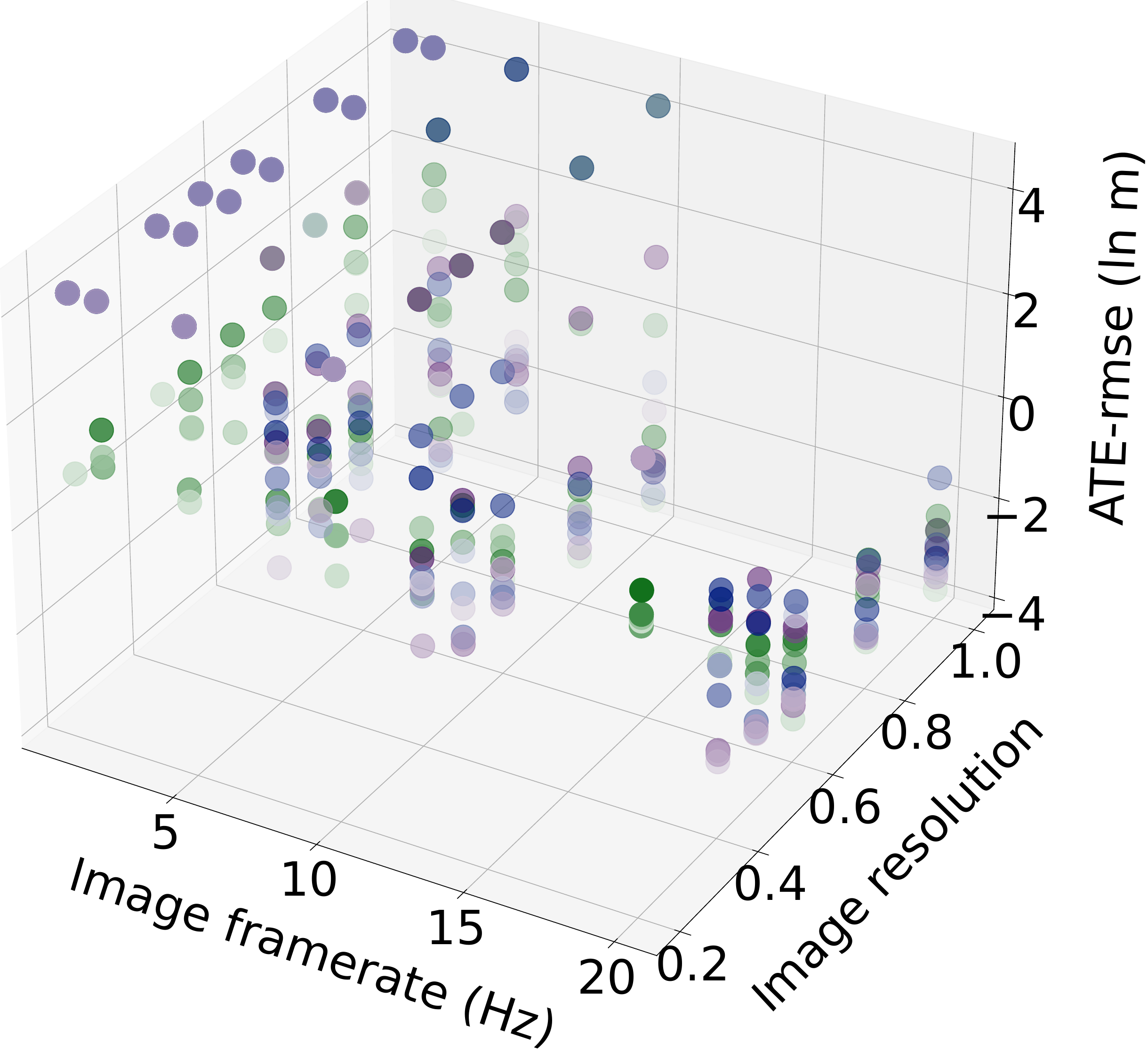}}
		\subfloat[Stereo-IMU: Res vs FPS vs ATE]{
			\label{fig:subfig:twofunction354} 
			\includegraphics[width=1.6in]{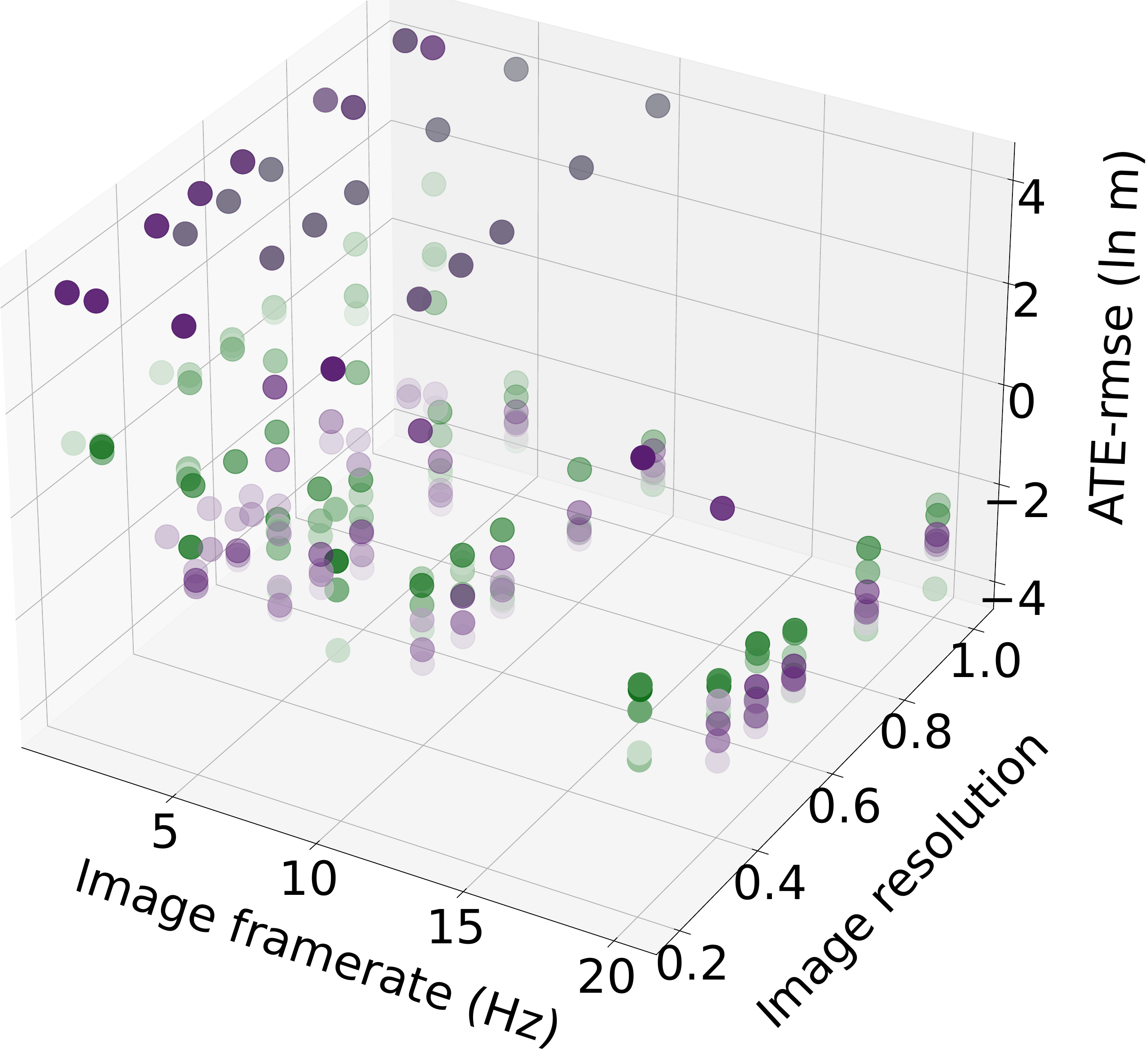}}

		\caption{Four visual SLAM algorithms comparison of image framerate, image resolution (original resolution is: $752$ $\times$ $480$), CPU and Memory usage and ATE RMSE on EuRoC dataset's 5 sequences. Each column represents one sensor mode. Blue is ORB-SLAM2, purple is ORB-SLAM3, yellow is VINS-Mono and green is VINS-Fusion. The colors from light to dark are the 5 sequences from 01 to 05 for the EuRoC dataset.}
		\label{fig: exploration3_1} 
	\end{figure*}

	Fig. \ref{fig: exploration3_1} shows 20 3D-diagrams: Each sensor combination (Mono, Mono-IMU, Stereo, Stereo-IMU) framerate (FPS) and resolution w.r.t. CPU and memory-max against ATE-RMSE. Additionally, the last row shows framerate vs. resolution vs. ATE-RMSE. Again, failed runs (traj\_length $< 0.75$) are shown with ATE of $1.2$ of the maximum. The diagrams are created from the data of the analysis pages of the 4 sensor combinations
	\footnote{
	 Mono: \hfill \url{https://archive.slam-hive.net/analysis/show/1730983871713760}\\
	 Mono-IMU: \hfill \url{https://archive.slam-hive.net/analysis/show/1730998946166233}\\
	 Stereo: \hfill \url{https://archive.slam-hive.net/analysis/show/1731048546700526}\\
	 Stereo-IMU: \hfill \url{https://archive.slam-hive.net/analysis/show/1731049902105419}
	}. 
	SLAM Hive shows that data without the failed runs.

	\begin{figure*}[t]
		\centering
		
		\captionsetup[subfloat]{font=scriptsize}
		\subfloat[Filter for traj\_length $>$ 75\%.]{
			\label{fig:subfig:newexp41} 
			\includegraphics[width=0.48\linewidth]{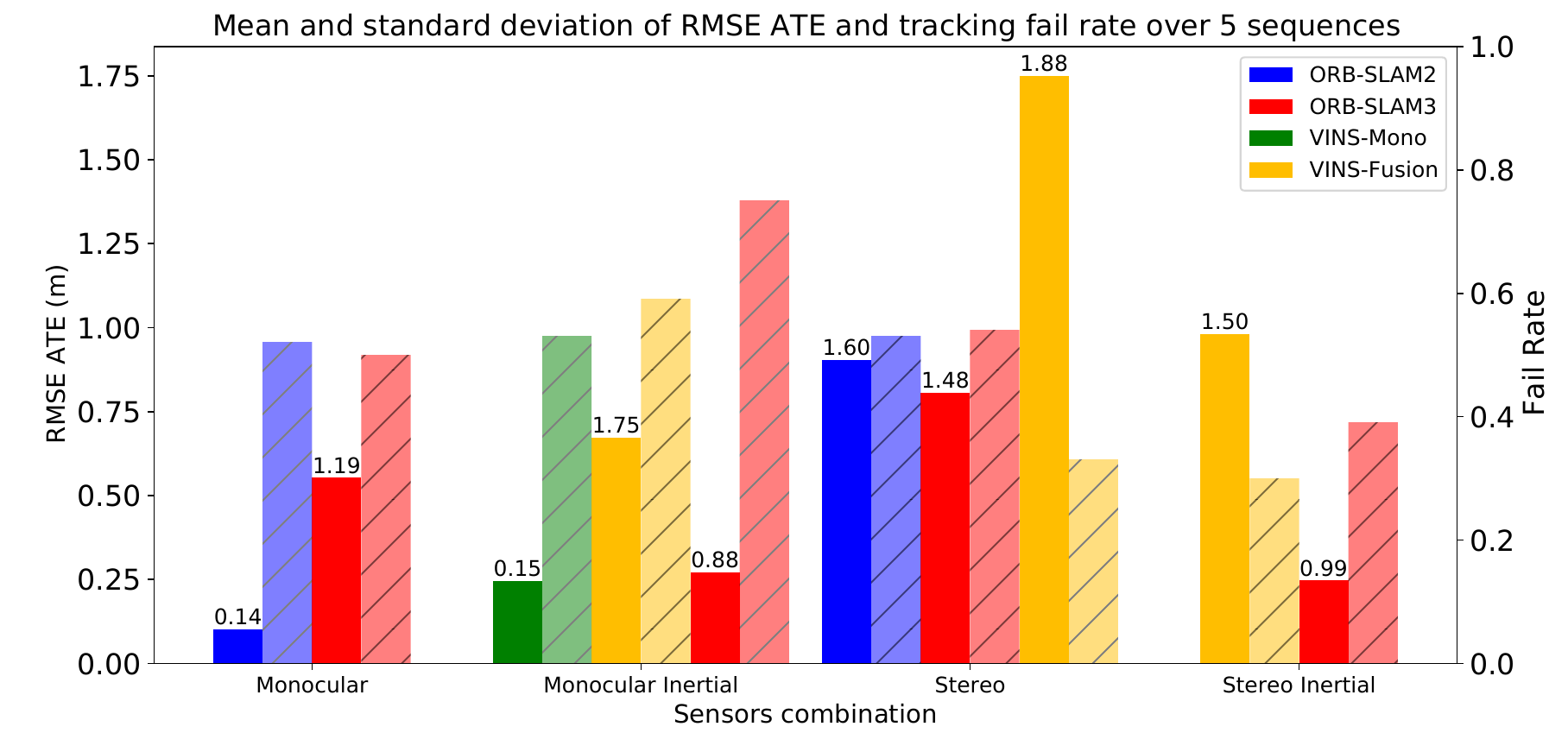}}
		\subfloat[Filter for traj\_length $>$ 75\% and ATE $<$ 1.0m.]{
			\label{fig:subfig:newexp42} 
			\includegraphics[width=0.48\linewidth]{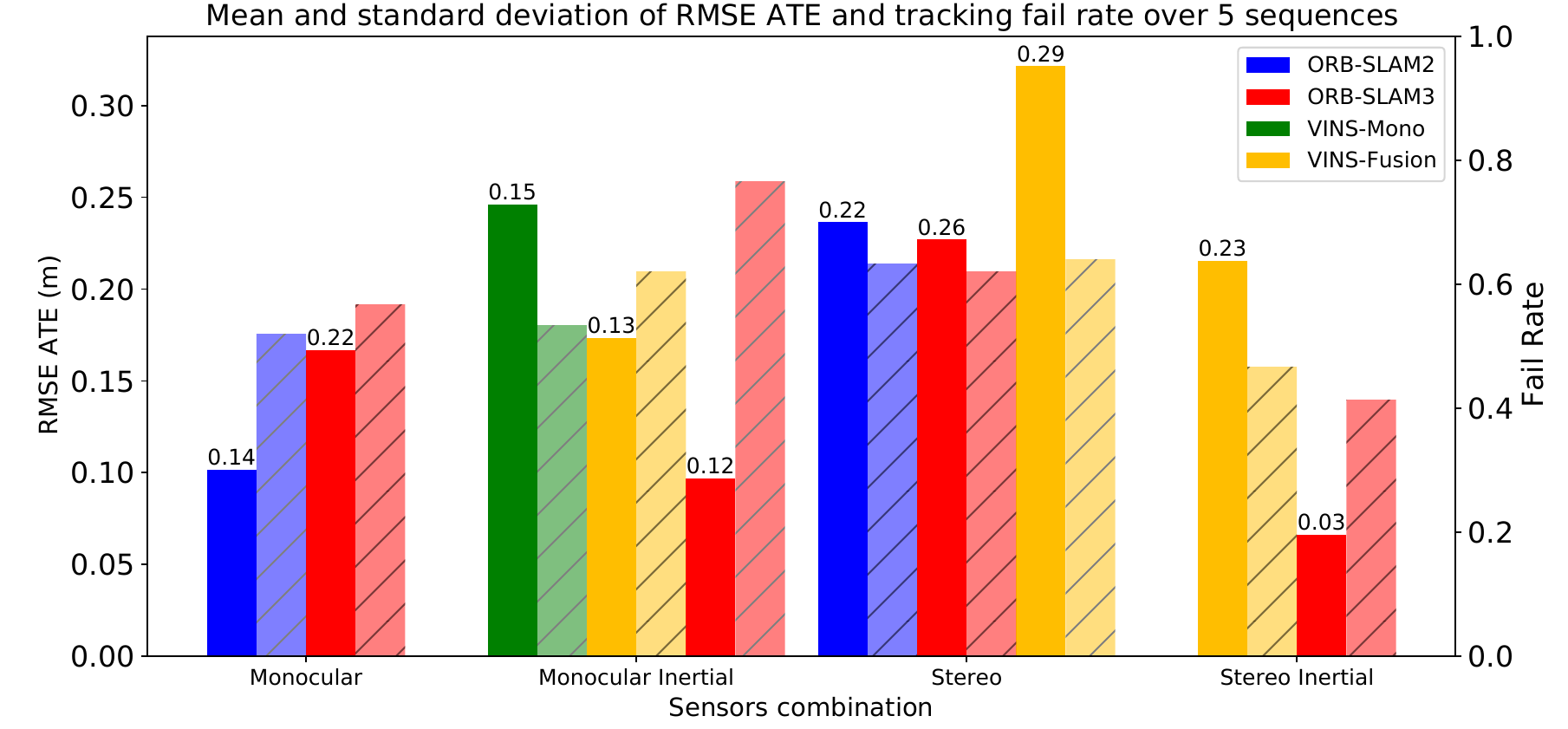}}
		\caption{Comparison of vSLAM algorithms, filtered by (a) traj\_length $>$ 75\%  and (b) additionally by ATE $<$ 1.0m. ATE is shown as solid bars, its std. deviation as a number above and the failure rate as a striped bar. }
		\label{fig: exploration4} 
				\vspace{-0.4cm}

	\end{figure*}

	\begin{figure*}[t]
		\centering
		
		\captionsetup[subfloat]{font=scriptsize}
		\subfloat[CPU: 1.5 cores and Memory: 1000 MB]{
			\label{fig:subfig:newexp31} 
			\includegraphics[width=1.8in]{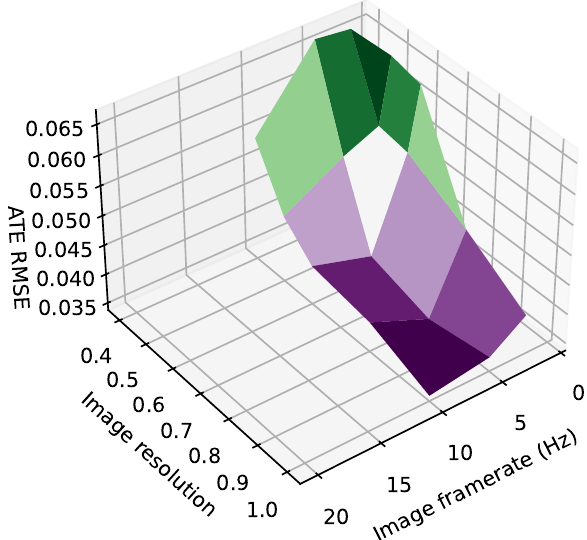}}
		\subfloat[CPU: 2.0 cores and Memory: 1500 MB]{
			\label{fig:subfig:newexp32} 
			\includegraphics[width=1.8in]{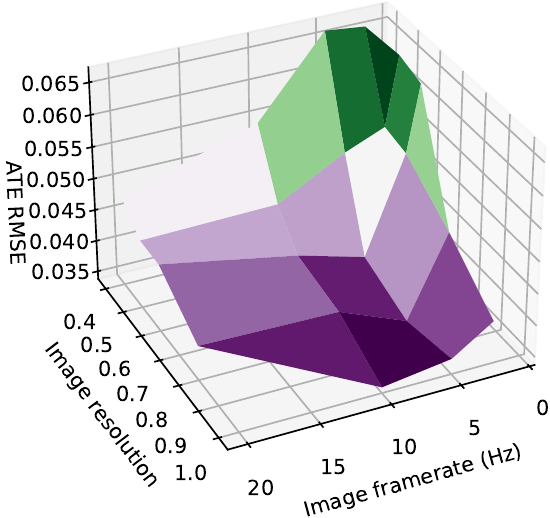}}
		\subfloat[CPU: 2.5 cores and Memory: 2000 MB]{
			\label{fig:subfig:newexp33} 
			\includegraphics[width=2.3in]{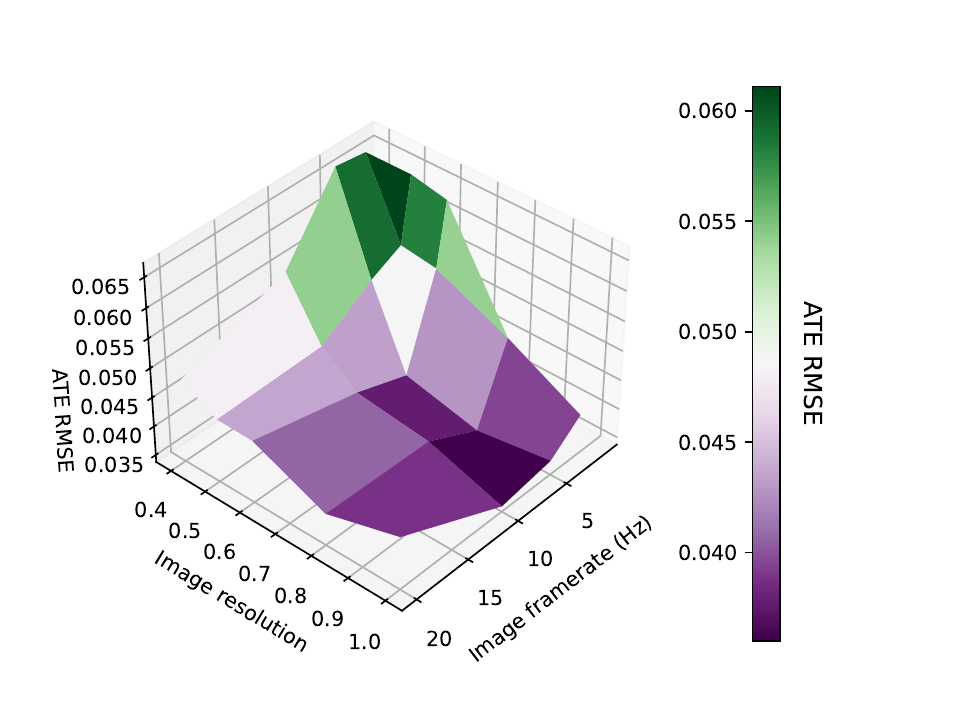}}
		\caption{ORB-SLAM3 algorithm stereo-inertial mode's ATE RMSE variation trend with image framerate and resolution on EuRoC dataset MH\_01\_easy sequence. The original image resolution is: $752$ $\times$ $480$.}
		\label{fig: new_exploration3_1} 
		
				\vspace{-0.4cm}
		
	\end{figure*}

	% 使用条件：[ 0.02 <= ATE.Mean <= 0.1]来对这1500个configurations进行过滤，得到了278个符合的结果，然后对这些configurations进行分析。
	
	Through the analysis result shown in Fig. \ref{fig: exploration1_1} and Fig. \ref{fig: exploration3_1}, we draw some conclusions for all vSLAM algorithms investigated:\\
	\begin{enumerate}
		\item Failed runs often use fewer CPU and Memory, because once they lost tracking they stop working.
		\item Looking at the last row, runs fail more often with lower frame rate, resolution seems to have a low effect on failure rate. 
		\item In contrast, especially for the IMU-based approaches, as long as they don't fail, the frame rate does not influence the ATE a lot. 
		\item The resolution does have an influence on the ATE.
		\item Resolutions of 0.2 (150 x 96) have quite bad ATE - the best one is VINS-Fusion-Stereo-IMU with 0.22m on MH\_02\_easy, most of them failed. In contrast, many 0.4 (300 x 192) resolution runs performed well. 
		\item Again looking at the last row, but for the successful runs, the IMU diagrams show a generally lower ATE. 
		\item The CPU consumption is mainly influenced by the frame-rate, while the resolution does not have a strong effect on CPU usage, presumably, because most of the work is done with the features, as compared to extracting the features from the images. 
		\item Memory consumption seems to be only little influenced by frame-rate and image resolution, presumably because the number of key-frames and number of features do not vary. 
	\end{enumerate}

	 \subsubsection{Algorithm Comparison} 
	 In Section \ref{subsec:vSLAM_experiments} we analyzed vSLAM algorithms using full resolution and 20Hz frame rate datasets. Using the big experiment with 1500 mapping runs we re-visit this algorithm comparison with lower-quality sensor data.  Fig. \ref{fig: exploration4} shows two diagrams of the different algorithms with their mean ATE-RMSE and their failure rates, where Fig. \ref{fig:subfig:newexp41} is using the default filter used in the big experiment of traj\_length $>$ 75\%, while Fig.  \ref{fig:subfig:newexp42} is additionally classifying mapping runs with an ATE $>$ 1.0m as failed, reasoning that such high ATE should also not count as successful mappings. 
	 
	 Parts of the mapping runs have to work with very difficult sensor data (low resolution and frame rate), that's why there are quite high failure rates and bad ATE, compared to Fig. \ref{fig: ate}. We observe that ORB-SLAM2 performs better (Monocular) or equal (Stereo) to ORB-SLAM3, while ORB-SLAM3 is better when using IMU data, which ORB-SLAM2 does not support. VINS-Mono is, in terms of failure rate, better than ORB-SLAM3, while the latter seems to have better accuracy with good data. Comparing VINS-Fusion with ORB-SLAM3, we can say that VINS-Fusion is more robust (lower failure rate in Fig.  \ref{fig:subfig:newexp41}), but ORB-SLAM3 is more accurate if it does not fail (so more accurate with good data).

    \subsubsection{Exploration on Limited Computation Resources} 
     Often engineers need to implement vSLAM on a robot with limited computation resources. Given the ORB-SLAM3 stereo-inertial algorithm, we explore what influence the choice of camera and its configuration, w.r.t. image frame rate and resolution, has on ATE-RMSE, average CPU and memory. So in fact we explore a 5-dimensional space. Fig. \ref{fig: new_exploration3_1} shows ATE-RMSE vs. resolution vs. framerate for three cases, where the CPU and memory are maxed out at 1.5 cores and 1GB, 2.0 cores and 1.5GB and 2.5cores and 2GB, respectively. 
     
     It can be seen that resolutions of 0.2 all fail (they are filtered out by traj\_length). And in general, as already observed above, the resolution has a significant influence on the ATE-RMSE. Frame rate seems of less importance. Consequently, for all three limitation scenarios on CPU and memory, the diagram indicates that a higher-resolution camera should be used and frame-rate can be sacrificed - as least for scenarios (motion and environment) similar to the dataset used.

	\section{Conclusion}

   This paper presented the SLAM Hive Benchmarking Suite, a containerized
toolkit to systematically test and evaluate SLAM algorithms
under various configurations using datasets. The system is
very salable and flexible, supporting software using different standards, operating systems and also closed source solutions, as long as scripts can be written to interface
with them. Our software is providing a benchmarking system that, for the first time, is able to
explore and analyze the vast space of possible permutations
of configurations, datasets and algorithms, and can be deployed on a workstation, 
in a cluster or the cloud. It is also flexible towards the evaluation by
implementing these themselves as programs running in their
Docker container. The software is, together with a demo, available on \url{https://slam-hive.net}.
We have shown the capability of our system by evaluating 1716 mapping runs and analyzing them in detail against various aspects.

In the future we will use SLAM Hive in conjunction with the heterogeneous and massive datasets collected by the ShanghaiTech Mapping Robot  \cite{yang2022cluster, xu2024shanghaitechmappingrobotneed, lin2024collecting}   to, for the first time, systematically evaluate all kinds of different SLAM approaches against each other, thus driving the research on SLAM as a whole forward.

We will also extend SLAM Hive to support algorithms that need GPUs and also add evaluation modules that analyze the quality of the map, not just the trajectory. Furthermore, we are planning to extend the scope of SLAM Hive beyond just SLAM algorithms, to also evaluate algorithms like segmentation or object recognition. We then also plan on hosting a competition on all of these algorithms by running them in their Dockers on our hardware on secret datasets, thus offering fair and unbiased results that can be widely compared across very different modalities (e.g. RGB vs RGBD vs LiDAR vs event camera vs radar based localization, SLAM and segmentation). 
	%先用slamhive的？
	
%		Limited by the quality of raw data in the dataset, we are temporarily unable to perform large-scale parameter space exploration.
%	But in the future, ShanghaiTech MappingRobot \cite{yang2022cluster} \cite{xu2024shanghaitechmappingrobotneed} \cite{lin2024collecting} will collect an all-around dataset, which can be used for more comprehensive evaluation.

	\section*{Acknowledgments}
	This work was supported by Science and Technology Commission of Shanghai Municipality (STCSM), project 22JC1410700 \textquotedblleft Evaluation of real-time localization and mapping algorithms for intelligent robots\textquotedblright .
	%This work was supported by the Science and Technology Commission of Shanghai Municipality (STCSM), project 22JC1410700 "Evaluation of real-time localization and mapping algorithms for intelligent robots". 
	This work has also been partially funded by the Shanghai Frontiers Science Center of Human-centered Artificial Intelligence.
	The experiments of this work were supported by the core facility Platform of Computer Science and Communication, SIST, ShanghaiTech University.

	% TODO position of appendix and reference

	% \newpage
	%\twocolumn
	\bibliographystyle{IEEEtran}
	\bibliography{ref}

	%\begin{thebibliography}{1}
	%\bibliographystyle{IEEEtran}
	%
	%\bibitem{ref1}
	%{\it{Mathematics Into Type}}. American Mathematical Society. [Online]. Available: https://www.ams.org/arc/styleguide/mit-2.pdf
	%
	%\end{thebibliography}

	% \newpage
	\begin{IEEEbiography}[{\includegraphics[width=1in,height=1.25in,clip,keepaspectratio]{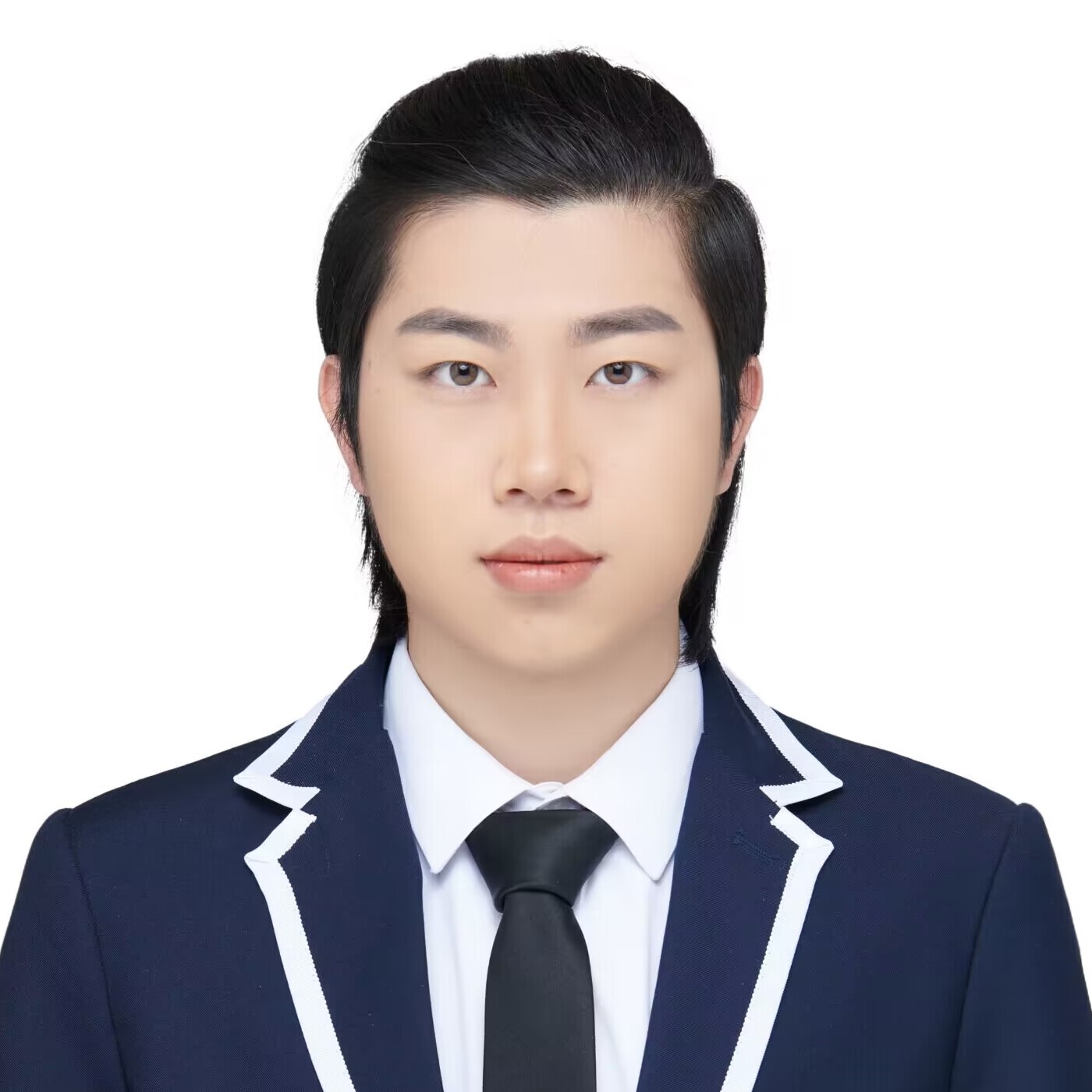}}]{Xinzhe Liu}
		received his B.S. at Shandong University in China, and now he is a second-year graduate student at ShanghaiTech University. His research focuses on SLAM evaluation. He is also interesting in Embodied intelligence and how to apply Large Language Models (LLMs) in Robotics and Autonomous Driving.
	\end{IEEEbiography}

	\begin{IEEEbiography}[{\includegraphics[width=1in,height=1.25in,clip,keepaspectratio]{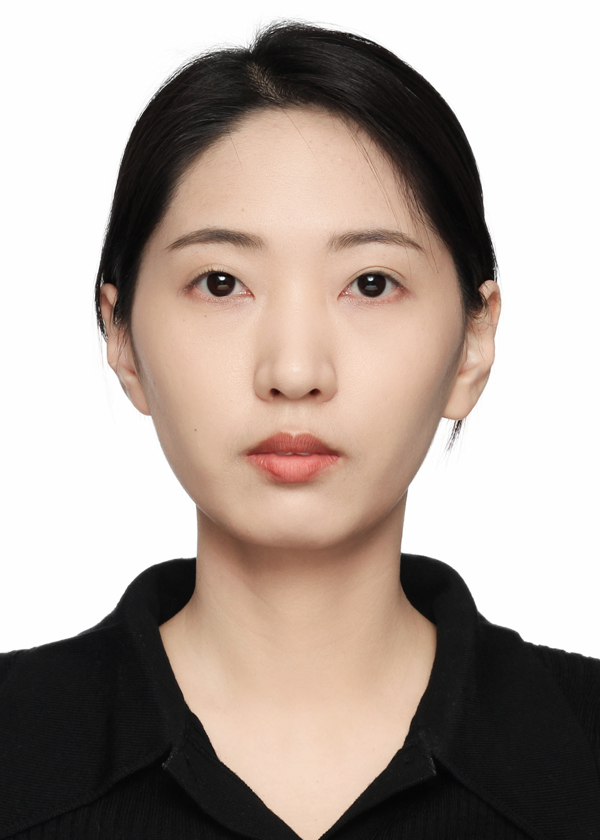}}]{Yuanyuan Yang}
		received Master degree in Computer Science and Technology from ShanghaiTech University, Shanghai, China, in 2023. Her research interests include SLAM evaluation, sensor fusion and mapping for moblie robots.  She is currently working in industry on autonomous  driving of intelligent electric vehicles.
	\end{IEEEbiography}
	
	\begin{IEEEbiography}[{\includegraphics[width=1in,height=1.25in,clip,keepaspectratio]{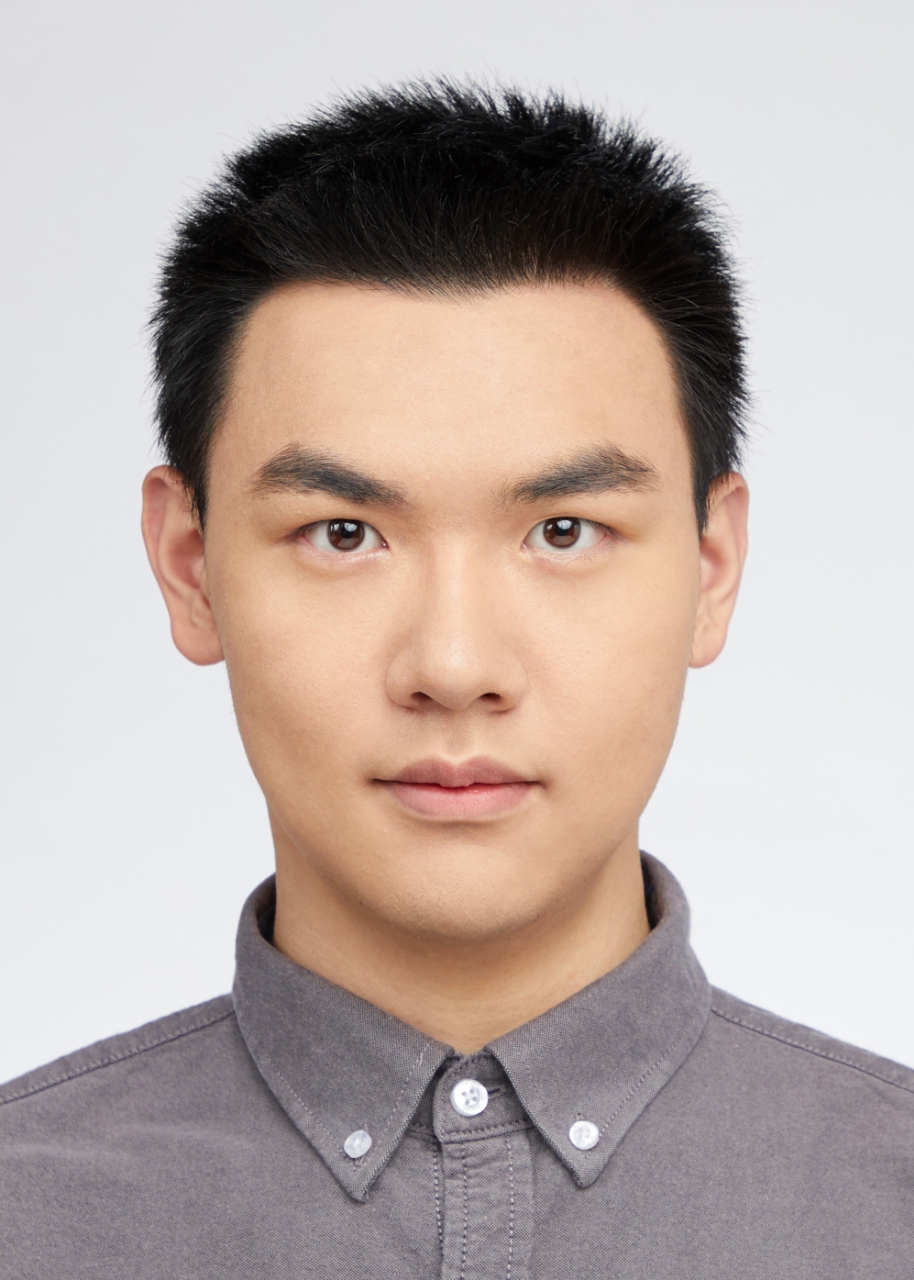}}]{Bowen Xu}
		Bowen Xu received his B.S. at ShanghaiTech University in China, and now he is a third-year graduate student at ShanghaiTech University. His research focuses on Robot Mapping and SLAM. 
	\end{IEEEbiography}

	\begin{IEEEbiography}[{\includegraphics[width=1in,height=1.25in,clip,keepaspectratio]{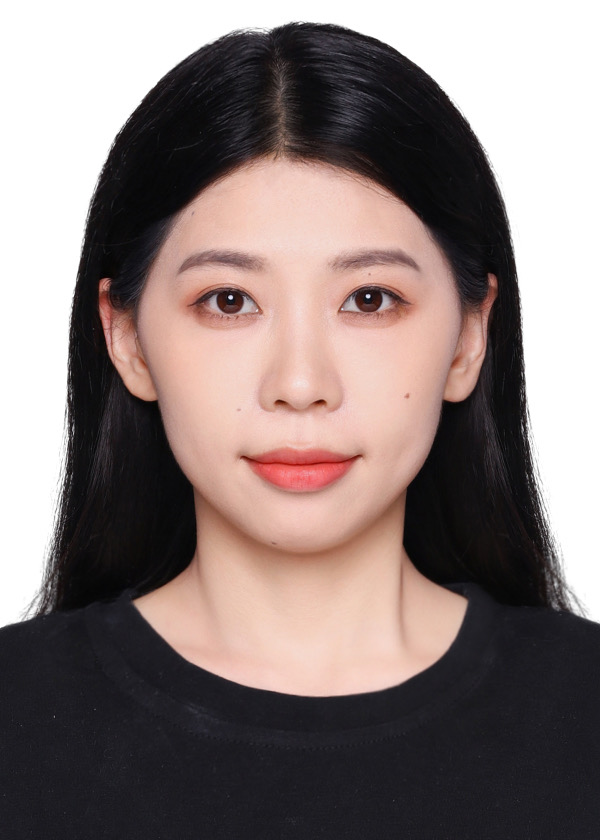}}]{Delin Feng}
	received Master degree in Computer Science and Technology from ShanghaiTech University, Shanghai, China, in 2023. Her research interests include autonomous driving.  She is currently working in industry on autonomous  driving.
\end{IEEEbiography}
	
	\begin{IEEEbiography}[{\includegraphics[width=1in,height=1.25in,clip,keepaspectratio]{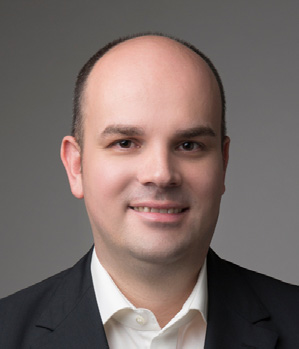}}]{Prof. Dr. S\"oren Schwertfeger}
		is an Associate Professor
		at ShanghaiTech University, where he joined in 2014. He is the head of the Mobile Autonomous Robotics Systems Lab and a director of the ShanghaiTech Automation and Robotics Center. 
		In 2012 he
		received his Ph.D. in Computer Science from the Jacobs
		University Bremen, researching at the robotics group of Prof.
		Andreas Birk. In 2010
		Dr. Schwertfeger was a guest researcher at the National Institute of Standards
		and Technology (NIST) in Gaithersburg, Maryland, USA. His
		research interest is in robotics, especially intelligent functions for mobile robots.
		Besides his work on mapping, map representation and SLAM, Dr. Schwertfeger is working on mobile
		manipulation, robot autonomy an AI for robotics. He was the general chair of the 2017 IEEE International Symposium
		on Safety, Security, and Rescue Robotics (SSRR), a guest editor of the Journal of Field Robotics and an associate editor of the IEEE Robotics and Automation Magazine (2018-2021). He is also a member of the board of Trustees of the RoboCup Federation.
	\end{IEEEbiography}

	\vfill
	
\end{document}